%% file: main.tex
\def\year{2018}\relax
\documentclass[letterpaper]{article} %DO NOT CHANGE THIS
\usepackage{aaai18}  %Required
\usepackage{times}  %Required
\usepackage{helvet}  %Required
\usepackage{courier}  %Required
\usepackage{url}  %Required
\usepackage{graphicx}  %Required
\usepackage{booktabs}       % professional-quality tables
\usepackage{amsfonts}       % blackboard math symbols
\usepackage{nicefrac}       % compact symbols for 1/2, etc.
\usepackage{microtype}      % microtypography
\usepackage{multirow}
\usepackage{makecell}
\usepackage{footnote}
\usepackage{amsmath}
\usepackage{amssymb}
\usepackage{float}

\newcommand*\samethanks[1][\value{footnote}]{\footnotemark[#1]}

\begin{document}
% The file aaai.sty is the style file for AAAI Press
% proceedings, working notes, and technical reports.

\title{Deep Reinforcement Learning that Matters}
\author{
Peter Henderson\textsuperscript{1}\thanks{These two authors contributed equally},
Riashat Islam\textsuperscript{1,2}\samethanks,
Philip Bachman\textsuperscript{2}\\
\Large{\bf{Joelle Pineau\textsuperscript{1},
Doina Precup\textsuperscript{1},
David Meger\textsuperscript{1}}}\\
\textsuperscript{1} McGill University, Montreal, Canada\\
\textsuperscript{2} Microsoft Maluuba, Montreal, Canada\\
\url{{peter.henderson,riashat.islam}@mail.mcgill.ca}, \url{phbachma@microsoft.com}\\
\url{{jpineau,dprecup}@cs.mcgill.ca}, \url{dmeger@cim.mcgill.ca}
}

\maketitle

% ABSTRACT
\begin{abstract}
% \todo{abstract to go straight into punchline}
In recent years, significant progress has been made in solving challenging problems across various domains using deep reinforcement learning (RL). Reproducing existing work and accurately judging the improvements offered by novel methods is vital to sustaining this progress. Unfortunately, reproducing results for state-of-the-art deep RL methods is seldom straightforward. In particular, non-determinism in standard benchmark environments, combined with variance intrinsic to the methods, can make reported results tough to interpret. Without significance metrics and tighter standardization of experimental reporting, it is difficult to determine whether improvements over the prior state-of-the-art are meaningful. In this paper, we investigate challenges posed by reproducibility, proper experimental techniques, and reporting procedures. We illustrate the variability in reported metrics and results when comparing against common baselines and suggest guidelines to make future results in deep RL more reproducible. We aim to spur discussion about how to ensure continued progress in the field by minimizing wasted effort stemming from results that are non-reproducible and easily misinterpreted.
\end{abstract}

% INTRODUCTION
\section{Introduction}
\label{sec:introduction}
\begin{figure}
\centering\includegraphics[width=0.41\textwidth]{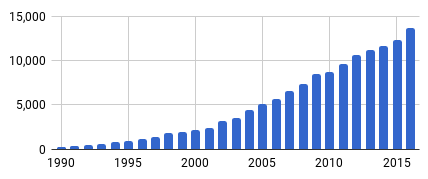}
\caption{Growth of published reinforcement learning papers. Shown are the number of RL-related publications (y-axis) per year (x-axis) scraped from Google Scholar searches.}
\label{fig:rlpubs}
\vspace{-1.5mm}
\end{figure}

Reinforcement learning (RL) is the study of how an agent can interact with its environment to learn a policy which maximizes expected cumulative rewards for a task. Recently, RL has experienced dramatic growth in attention and interest due to promising results in areas like: controlling continuous systems in robotics~\cite{lillicrap2015continuous}, playing Go~\cite{silver2016mastering}, Atari~\cite{mnih2013playing}, and competitive video games~\cite{vinyals2017starcraft,silva2017moba}. Figure~\ref{fig:rlpubs} illustrates growth of the field through the number of publications per year. To maintain rapid progress in RL research, it is important that existing works can be easily reproduced and compared to accurately judge improvements offered by novel methods.

However, reproducing deep RL results is seldom straightforward, and the literature reports a wide range of results for the same baseline algorithms \cite{repro_workshop}. Reproducibility can be affected by extrinsic factors (e.g.~hyperparameters or codebases) and intrinsic factors (e.g.~effects of random seeds or environment properties). We investigate these sources of variance in reported results through a representative set of experiments. For clarity, we focus our investigation on policy gradient (PG) methods in continuous control. Policy gradient methods with neural network function approximators have been particularly successful in continuous control \cite{TRPO,PPO,DDPG} and are competitive with value-based methods in discrete settings. We note that the diversity of metrics and lack of significance testing in the RL literature creates the potential for misleading reporting of results. We demonstrate possible benefits of significance testing using techniques common in machine learning and statistics.

% \vspace{-2mm}

Several works touch upon evaluating RL algorithms. \citeauthor{rllab} \shortcite{rllab} benchmark several RL algorithms and provide the community with baseline implementations. Generalizable RL evaluation metrics are proposed in~\cite{environment_overfitting}. \citeauthor{machado2017revisiting} \shortcite{machado2017revisiting} revisit the Arcade Learning Environment to propose better evaluation methods in these benchmarks. However, while the question of \emph{reproducibility and good experimental practice} has been examined in related fields~\cite{wagstaff2012machine,boulesteix2013plea,stodden2014implementing,bouckaert2004evaluating,bouckaert2004estimating,vaughan}, to the best of our knowledge this is the first work to address this important question in the context of deep RL.

In each section of our experimental analysis, we pose questions regarding key factors affecting reproducibility. We find that there are numerous sources of non-determinism when reproducing and comparing RL algorithms. To this end, we show that fine details of experimental procedure can be critical. Based on our experiments, we conclude with possible recommendations, lines of investigation, and points of discussion for future works to ensure that deep reinforcement learning is reproducible and continues to matter.

% BACKGROUND ON POLICY GRADIENTS
\section{Technical Background}
\label{sec:background}

This work focuses on several model-free policy gradient algorithms with publicly available implementations which appear frequently in the literature as baselines for comparison against novel methods.
We experiment with Trust Region Policy Optimization (TRPO) \cite{TRPO}, Deep Deterministic Policy Gradients (DDPG) \cite{DDPG}, Proximal Policy Optimization (PPO) \cite{PPO}, and Actor Critic using Kronecker-Factored Trust Region (ACKTR) \cite{ACKTR}. These methods have shown promising results in continuous control MuJoCo domain tasks \cite{MuJoCo} from OpenAI Gym \cite{gym}.
Generally, they optimize $\rho (\theta, s_0) = \mathbb{E}_{\pi_\theta}\left[ \sum_{t=0}^\infty \gamma^t r(s_t) | s_0 \right]$, using the policy gradient theorem: $\frac{\delta \rho (\theta, s_0)}{\delta \theta} = \sum_s \mu_{\pi_\theta} (s | s_0) \sum_a \frac{\delta \pi_\theta (a | s)}{\delta \theta} Q_{\pi_\theta} (s,a)$.
Here, $\mu_{\pi_\theta} (s | s_0) = \sum_{t=0}^\infty \gamma^t P(s_t = s | s_0)$ \cite{sutton2000policy}.
TRPO \cite{TRPO} and PPO \cite{PPO} use constraints and advantage estimation to perform this update, reformulating the optimization problem as: $\max_\theta \mathbb{E}_{t} \left[ \frac{\pi_\theta (a_t|s_t)}{\pi_{\theta_{old}}(a_t|s_t)} A_t(s_t, a_t) \right]$. Here, $A_t$ is the generalized advantage function \cite{schulman2015high}.
TRPO uses conjugate gradient descent as the optimization method with a KL constraint: $\mathbb{E}_t \left[\text{KL} \left[ \pi_{\theta_{old}} ( \cdot | s_t), \pi_\theta (\cdot | s_t) \right] \right] \le \delta$. PPO reformulates the constraint as a penalty (or clipping objective). DDPG and ACKTR use actor-critic methods which estimate $Q(s,a)$ and optimize a policy that maximizes the $Q$-function based on Monte-Carlo rollouts. DDPG does this using deterministic policies, while ACKTR uses Kronecketer-factored trust regions to ensure stability with stochastic policies.

% EXPERIMENTAL ANALYSIS, WHAT ARE WE INVESTIGATING
\section{Experimental Analysis}
\label{sec:experimental_analysis}
We pose several questions about the factors affecting reproducibility of state-of-the-art RL methods. We perform a set of experiments designed to provide insight into the questions posed. In particular, we investigate the effects of: specific hyperparameters on algorithm performance if not properly tuned; random seeds and the number of averaged experiment trials; specific environment characteristics; differences in algorithm performance due to stochastic environments; differences due to codebases with most other factors held constant. For most of our experiments\footnote{Specific details can be found in the supplemental and code can be found at: \url{https://git.io/vFHnf}}, except for those comparing codebases, we generally use the OpenAI Baselines\footnote{\url{https://www.github.com/openai/baselines}} implementations of the following algorithms: ACKTR~\cite{ACKTR}, PPO~\cite{PPO}, DDPG~\cite{plappert2017parameter}, TRPO~\cite{PPO}. We use the Hopper-v1 and HalfCheetah-v1 MuJoCo~\cite{MuJoCo} environments from OpenAI Gym~\cite{gym}. These two environments provide contrasting dynamics (the former being more unstable).

To ensure fairness we run five experiment trials for each evaluation, each with a different preset random seed (all experiments use the same set of random seeds). In all cases, we highlight important results here, with full descriptions of experimental setups and additional learning curves included in the supplemental material. Unless otherwise mentioned, we use default settings whenever possible, while modifying only the hyperparameters of interest. All results (including graphs) show mean and standard error across random seeds.

We use multilayer perceptron function approximators in all cases. We denote the hidden layer sizes and activations as $(N, M, \text{activation})$. For default settings, we vary the hyperparameters under investigation one at a time. For DDPG we use a network structure of $(64, 64, \text{ReLU})$ for both actor and critic. For TRPO and PPO, we use $(64, 64, \text{tanh})$ for the policy. For ACKTR, we use $(64, 64, \text{tanh})$ for the actor and $(64, 64, \text{ELU})$ for the critic.

\subsection{Hyperparameters}

\textit{What is the magnitude of the effect hyperparameter settings can have on baseline performance?}
\medskip

\noindent Tuned hyperparameters play a large role in eliciting the best results from many algorithms. However, the choice of optimal hyperparameter configuration is often not consistent in related literature, and the range of values considered is often not reported\footnote{A sampled literature review can be found in the supplemental.}. Furthermore, poor hyperparameter selection can be detrimental to a fair comparison against baseline algorithms. Here, we investigate several aspects of hyperparameter selection on performance.

\begin{figure*}[!htb]
    \centering
    \includegraphics[width=.31\textwidth]{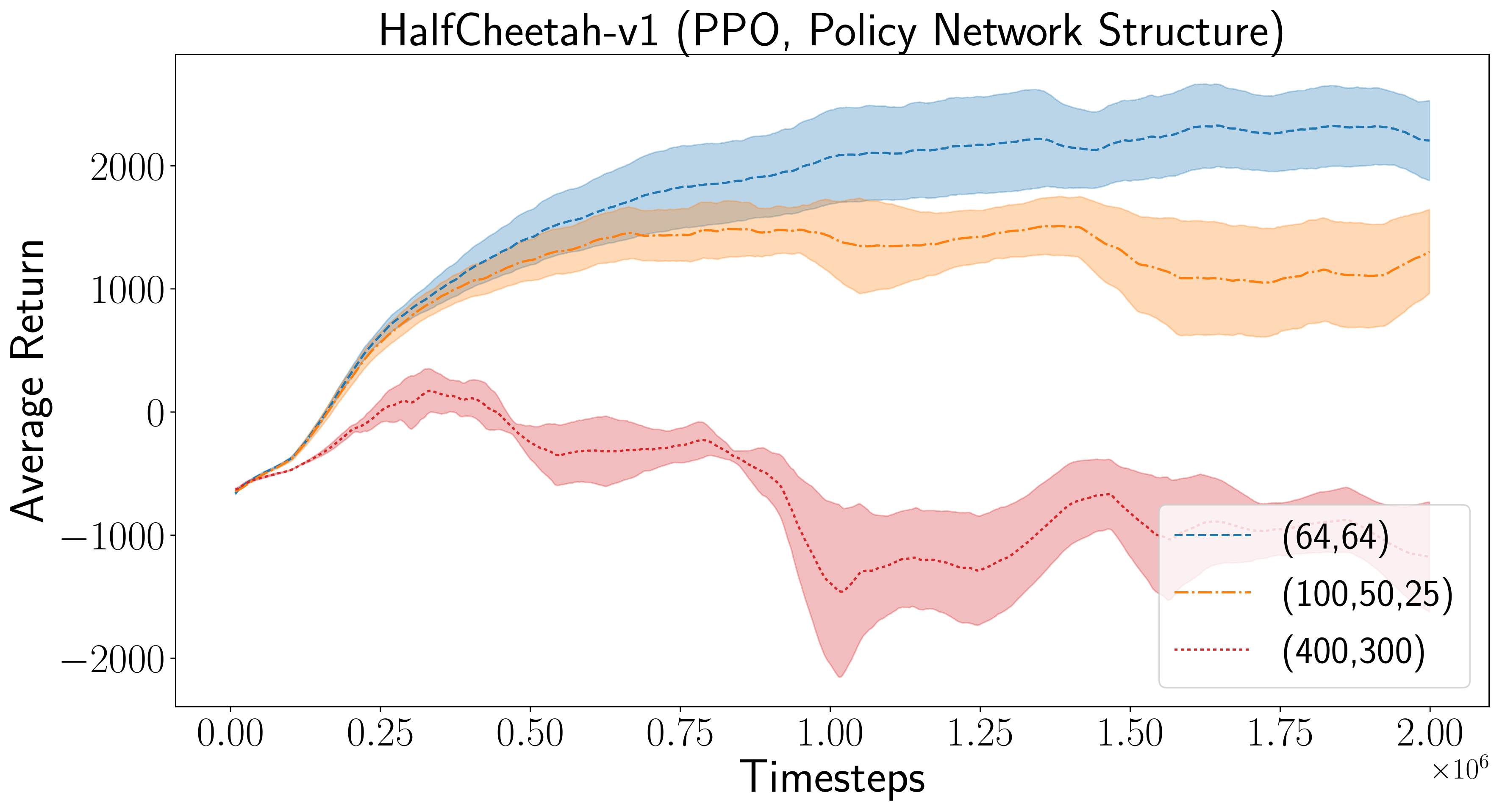}
    \includegraphics[width=.31\textwidth]{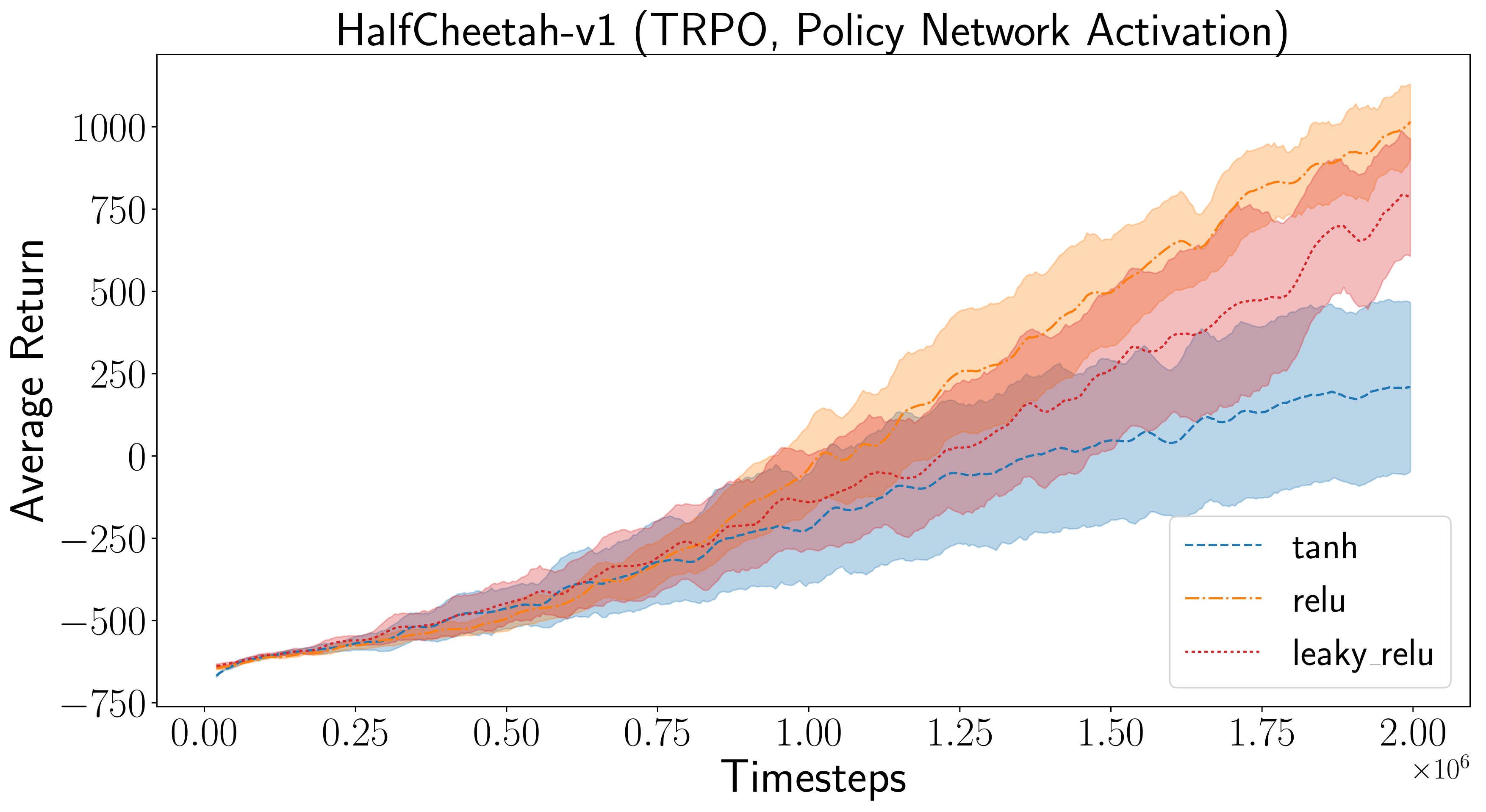}
    \includegraphics[width=.31\textwidth]{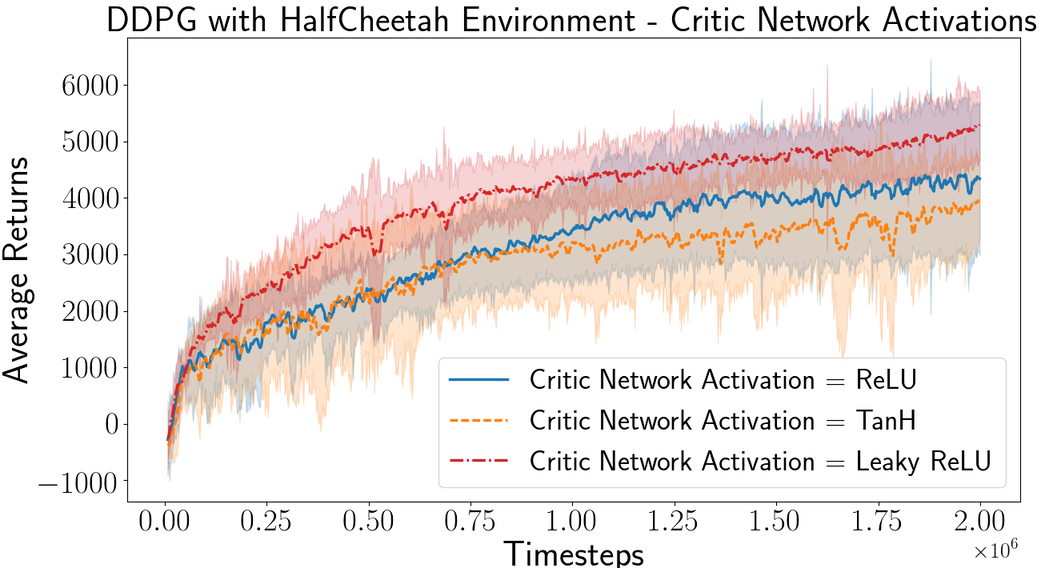}
    \caption{Significance of Policy Network Structure and Activation Functions PPO (left), TRPO (middle) and DDPG (right).}
    \label{fig:ppo_network_structure}
\end{figure*}

\subsection{Network Architecture}
\textit{How does the choice of network architecture for the policy and value function approximation affect performance?}
\medskip

\noindent In~\cite{repro_workshop}, it is shown that policy network architecture can significantly impact results in both TRPO and DDPG. Furthermore, certain activation functions such as Rectified Linear Unit (ReLU) have been shown to cause worsened learning performance due to the ``dying relu'' problem~\cite{xu2015empirical}. As such, we examine network architecture and activation functions for both policy and value function approximators. In the literature, similar lines of investigation have shown the differences in performance when comparing linear approximators, RBFs, and neural networks~\cite{rajeswaran2017towards}. Tables~\ref{tab:architecture_results} and~\ref{tab:architecture_results_vf} summarize the final evaluation performance of all architectural variations after training on 2M samples (i.e. 2M timesteps in the environment). All learning curves and details on setup can be found in the supplemental material. We vary hyperparameters one at a time, while using a default setting for all others. We investigate three multilayer perceptron (MLP) architectures commonly seen in the literature: $(64,64)$, $(100,50,25)$, and $(400,300)$. Furthermore, we vary the activation functions of both the value and policy networks across tanh, ReLU, and Leaky ReLU activations.

\begin{figure*}[!htb]
    \centering
    \includegraphics[width=.31\textwidth]{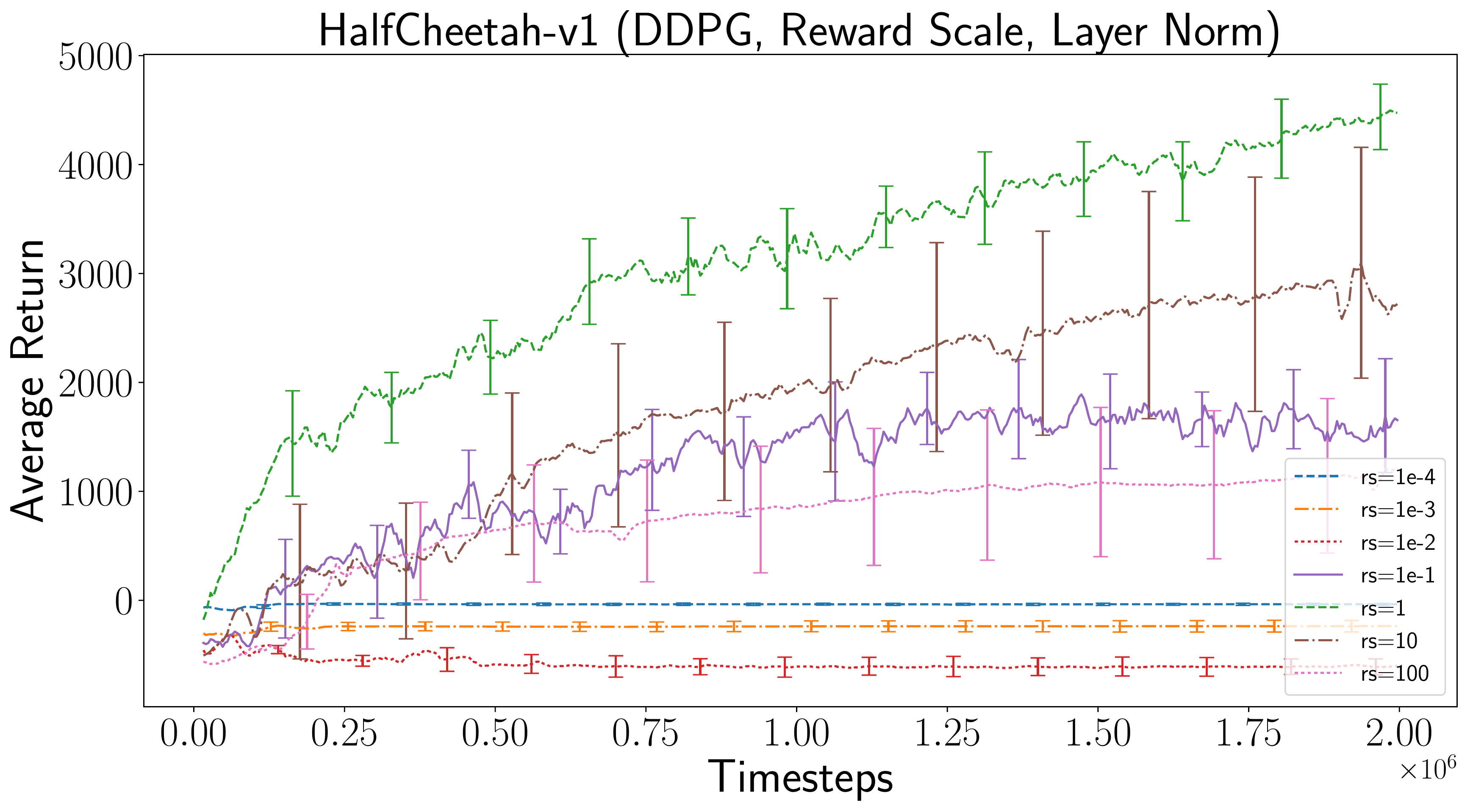}
    \includegraphics[width=.31\textwidth]{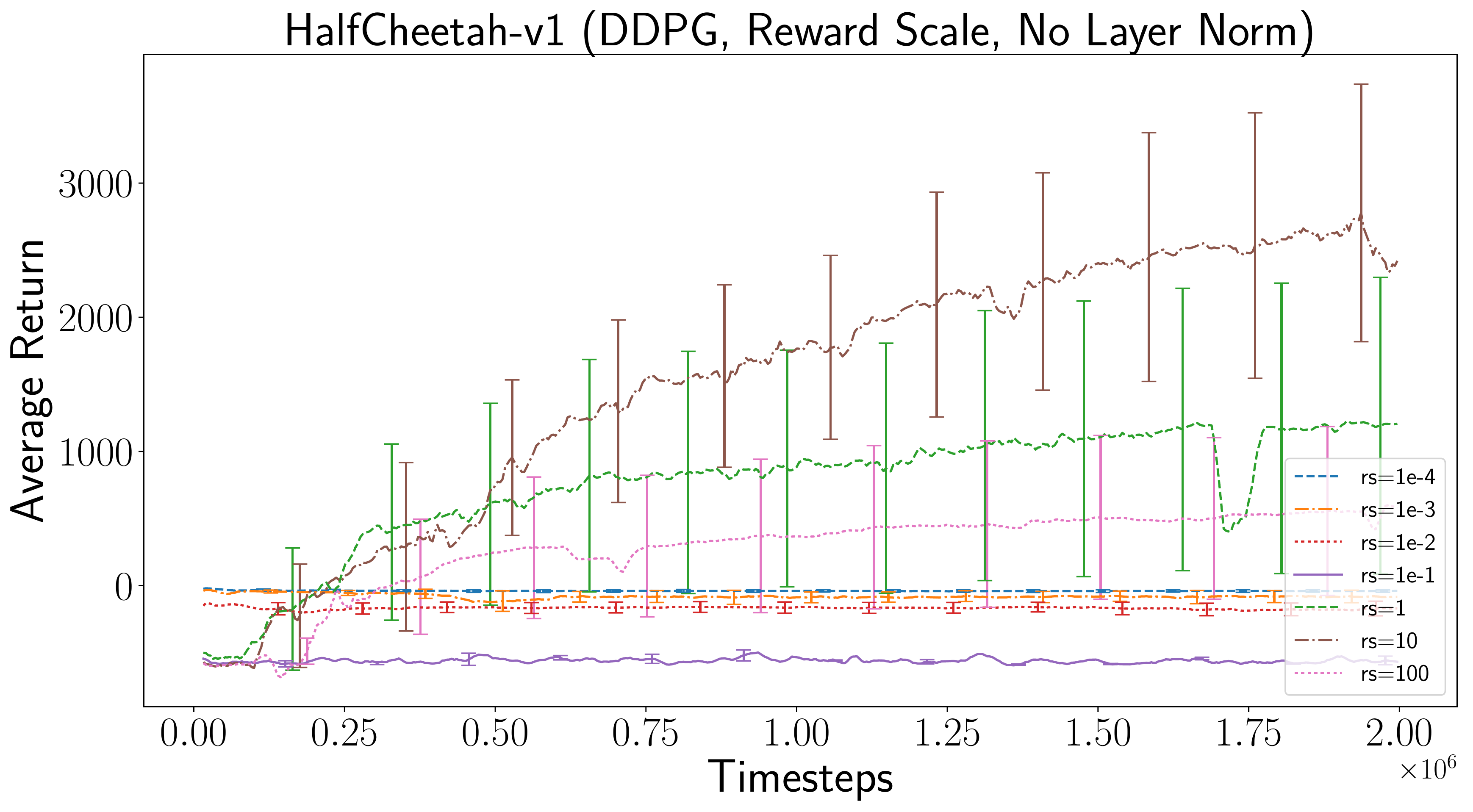}
    \caption{DDPG reward rescaling on HalfCheetah-v1, with and without layer norm.}
    \label{fig:rewardscale}
    \vspace{-1mm}
\end{figure*}

\textbf{Results} Figure~\ref{fig:ppo_network_structure} shows how significantly performance can be affected by simple changes to the policy or value network activations. We find that usually ReLU or Leaky ReLU activations perform the best across environments and algorithms. The effects are not consistent across algorithms or environments. This inconsistency demonstrates how interconnected network architecture is to algorithm methodology. For example, using a large network with PPO may require tweaking other hyperparameters such as the trust region clipping or learning rate to compensate for the architectural change\footnote{We find that the KL divergence of updates with the large network $(400,300)$ seen in Figure~\ref{fig:ppo_network_structure} is on average $33.52$ times higher than the KL divergence of updates with the $(64,64)$ network.}. This intricate interplay of hyperparameters is one of the reasons reproducing current policy gradient methods is so difficult. It is exceedingly important to choose an appropriate architecture for proper baseline results. This also suggests a possible need for hyperparameter agnostic algorithms---that is algorithms that incorporate hyperparameter adaptation as part of the design---such that fair comparisons can be made without concern about improper settings for the task at hand.

\begin{table*}
    \centering
    \scriptsize{\begin{tabular}{|c|c|c|c|c|c|c|c|}
    \hline
         \textbf{Algorithm} & \textbf{Environment} & \textbf{400,300} & \textbf{64,64} & \textbf{100,50,25} & \textbf{tanh} & \textbf{ReLU} & \textbf{LeakyReLU} \\
         \hline
         TRPO& Hopper-v1 & 2980 $\pm$ 35& 2674 $\pm$ 227& 3110 $\pm$ 78 &2674 $\pm$ 227 & 2772 $\pm$ 211& - \\
         \cite{TRPO}& HalfCheetah-v1 & 1791 $\pm$ 224 &1939 $\pm$ 140 &2151 $\pm$ 27 & 1939 $\pm$ 140 & 3041 $\pm$ 161 &- \\
         \hline
         TRPO& Hopper-v1 & 1243 $\pm$ 55& 1303 $\pm$ 89& 1243 $\pm$ 55 &1303 $\pm$ 89& 1131 $\pm$ 65 & 1341$\pm$ 127\\
         \cite{rllab}& HalfCheetah-v1 &738 $\pm$ 240&834 $\pm$ 317 &850$\pm$378 &834 $\pm$ 317 & 784 $\pm$ 352&1139 $\pm$364 \\
         \hline
         TRPO& Hopper-v1 & 2909 $\pm$ 87 & 2828 $\pm$ 70 & 2812 $\pm$ 88 &2828 $\pm$ 70& 2941 $\pm$ 91 & 2865 $\pm$ 189 \\
         \cite{PPO}& HalfCheetah-v1 & -155 $\pm$ 188 & 205 $\pm$ 256 & 306 $\pm$ 261 & 205 $\pm$ 256 & 1045 $\pm$ 114 & 778 $\pm$ 177 \\
         \hline
         PPO& Hopper-v1 & 61 $\pm$ 33 & 2790 $\pm$ 62 & 2592 $\pm$ 196 & 2790 $\pm$ 62 & 2695 $\pm$ 86& 2587 $\pm$ 53 \\
         \cite{PPO}& HalfCheetah-v1 & -1180 $\pm$ 444 & 2201 $\pm$ 323 &1314 $\pm$ 340 &  2201 $\pm$ 323 & 2971 $\pm$ 364 & 2895 $\pm$ 365 \\
         \hline
         DDPG& Hopper-v1 & 1419 $\pm$ 313  & 1632 $\pm$ 459  & 2142 $\pm$ 436 & 1491 $\pm$ 205  & 1632 $\pm$ 459 & 1384 $\pm$ 285 \\
         \cite{plappert2017parameter}& HalfCheetah-v1 & 5579 $\pm$ 354 & 4198 $\pm$ 606 & 5600 $\pm$ 601 & 5325 $\pm$ 281 & 4198 $\pm$ 606  &  4094 $\pm$ 233\\
        %  \cite{plappert2017parameter}& HalfCheetah-v1 & 5579.2 $\pm$ 354.2 & 4197.7 $\pm$ 606.0 & 5600.6 $\pm$ 601.4 & 5325 $\pm$ 281 & 4198 $\pm$ 606  &  4094 $\pm$ 233\\
         \hline
         DDPG& Hopper-v1 & 600 $\pm$ 126 &593 $\pm$ 155 &501 $\pm$ 129 &436 $\pm$ 48 & 593 $\pm$ 155& 319 $\pm$ 127 \\
         ~\cite{QPROP}& HalfCheetah-v1 & 2845 $\pm$ 589& 2771 $\pm$ 535& 1638 $\pm$ 624& 1638 $\pm$ 624& 2771 $\pm$ 535& 1405$\pm$ 511  \\
         \hline
         DDPG& Hopper-v1 & 506 $\pm$ 208 & 749 $\pm$ 271 &629 $\pm$ 138 & 354 $\pm$ 91 &  749 $\pm$ 271 & - \\
         ~\cite{rllab}& HalfCheetah-v1 & 850 $\pm$ 41 &1573 $\pm$ 385 &1224 $\pm$ 553 & 1311 $\pm$ 271 & 1573 $\pm$ 385 & - \\
         \hline
         ACKTR& Hopper-v1 &2577 $\pm$ 529 & 1608 $\pm$ 66 &2287 $\pm$ 946 &1608 $\pm$ 66 & 2835 $\pm$ 503 &  2718 $\pm$ 434 \\
         \cite{ACKTR}& HalfCheetah-v1 &2653 $\pm$ 408&2691 $\pm$ 231 &2498 $\pm$ 112 &2621 $\pm$ 381&2160 $\pm$ 151&2691 $\pm$ 231\\
         \hline
    \end{tabular}}
    \caption{Results for our policy architecture permutations across various implementations and algorithms. Final average $\pm$ standard error across 5 trials of returns across the last 100 trajectories after 2M training samples. For ACKTR, we use \textbf{ELU} activations instead of leaky ReLU.}
    \label{tab:architecture_results}
\end{table*}

\begin{table*}
    \centering
    \scriptsize{\begin{tabular}{|c|c|c|c|c|c|c|c|}
    \hline
         \textbf{Algorithm} & \textbf{Environment} & \textbf{400,300} & \textbf{64,64} & \textbf{100,50,25} & \textbf{tanh} & \textbf{ReLU} & \textbf{LeakyReLU} \\
         \hline
         TRPO& Hopper-v1 & 3011 $\pm$ 171 & 2674 $\pm$ 227 & 2782 $\pm$ 120 & 2674 $\pm$ 227 & 3104 $\pm$ 84 & - \\
         \cite{TRPO}& HalfCheetah-v1 &2355 $\pm$ 48 & 1939 $\pm$ 140 & 1673 $\pm$ 148 & 1939 $\pm$ 140 & 2281 $\pm$ 91 & - \\
         \hline
        %  Rllab uses a linear function approximator for it.
        %  TRPO& Hopper-v1 & & & & & & \\
        %  \cite{rllab}& HalfCheetah-v1 & & & & & & \\
        %  \hline
         TRPO& Hopper-v1 & 2909 $\pm$ 87 & 2828 $\pm$ 70 & 2812 $\pm$ 88 & 2828 $\pm$ 70 & 2829 $\pm$ 76 & 3047 $\pm$ 68 \\
         \cite{PPO}& HalfCheetah-v1 & 178 $\pm$ 242 &205 $\pm$ 256& 172 $\pm$ 257 & 205 $\pm$ 256& 235 $\pm$ 260 & 325 $\pm$ 208 \\
         \hline
          PPO& Hopper-v1 & 2704 $\pm$ 37 & 2790 $\pm$ 62 & 2969 $\pm$ 111 & 2790 $\pm$ 62 & 2687 $\pm$ 144& 2748 $\pm$ 77 \\
         \cite{PPO}& HalfCheetah-v1 & 1523 $\pm$ 297 & 2201 $\pm$ 323 & 1807 $\pm$ 309 & 2201 $\pm$ 323 & 1288 $\pm$ 12 & 1227 $\pm$ 462 \\
         \hline
        DDPG& Hopper-v1 & 1419 $\pm$ 312  & 1632 $\pm$ 458 &  1569 $\pm$ 453  & 971 $\pm$ 137  & 852 $\pm$ 143  & 843 $\pm$ 160 \\
         \cite{plappert2017parameter} & HalfCheetah-v1 & 5600 $\pm$ 601 & 4197 $\pm$ 606 & 4713 $\pm$ 374 & 3908 $\pm$ 293 & 4197 $\pm$ 606 &  5324 $\pm$ 280\\
        %  DDPG& Hopper-v1 & 1419.0 $\pm$ 312.7  & 1632.1 $\pm$ 458.9 &  1569.3 $\pm$ 453.2  & 971.7 $\pm$ 137.2  & 851.9 $\pm$ 142.8  & 843.3 $\pm$ 159.9  \\
        %  \cite{plappert2017parameter} & HalfCheetah-v1 & 5600.6 $\pm$ 601.4 & 4197.7 $\pm$ 606.0 & 4713.4 $\pm$ 374.4 & 3908.8 $\pm$ 293.5 & 4197.7 $\pm$ 606.0 &  5324.5 $\pm$ 280.6\\
         \hline
         DDPG& Hopper-v1 &523 $\pm$ 248& 343 $\pm$ 34 & 345 $\pm$ 44 & 436 $\pm$ 48 & 343 $\pm$ 34& - \\
         \cite{QPROP}& HalfCheetah-v1 & 1373 $\pm$ 678 & 1717 $\pm$ 508 & 1868 $\pm$ 620 & 1128 $\pm$ 511 & 1717 $\pm$ 508 & - \\
         \hline
         DDPG& Hopper-v1 & 1208 $\pm$ 423 & 394 $\pm$ 144 & 380 $\pm$ 65 & 354 $\pm$ 91 & 394 $\pm$ 144 & - \\
         \cite{rllab}& HalfCheetah-v1 & 789 $\pm$ 91 &1095 $\pm$ 139 & 988 $\pm$ 52 & 1311 $\pm$ 271 & 1095 $\pm$ 139& -  \\
         \hline
         ACKTR& Hopper-v1 & 152 $\pm$ 47 &  1930 $\pm$ 185&1589 $\pm$ 225 & 691 $\pm$ 55 & 500 $\pm$ 379 & 1930 $\pm$ 185\\
         \cite{ACKTR}& HalfCheetah-v1 & 518 $\pm$ 632 &3018 $\pm$ 386& 2554 $\pm$ 219 & 2547 $\pm$ 172 & 3362 $\pm$ 682 & 3018 $\pm$ 38\\
         \hline
    \end{tabular}}
    \caption{Results for our value function ($Q$ or $V$) architecture permutations across various implementations and algorithms. Final average $\pm$ standard error across 5 trials of returns across the last 100 trajectories after 2M training samples. For ACKTR, we use \textbf{ELU} activations instead of leaky ReLU.}
    \label{tab:architecture_results_vf}
\end{table*}

\begin{figure*}
    \centering
    \includegraphics[width=0.33\textwidth]{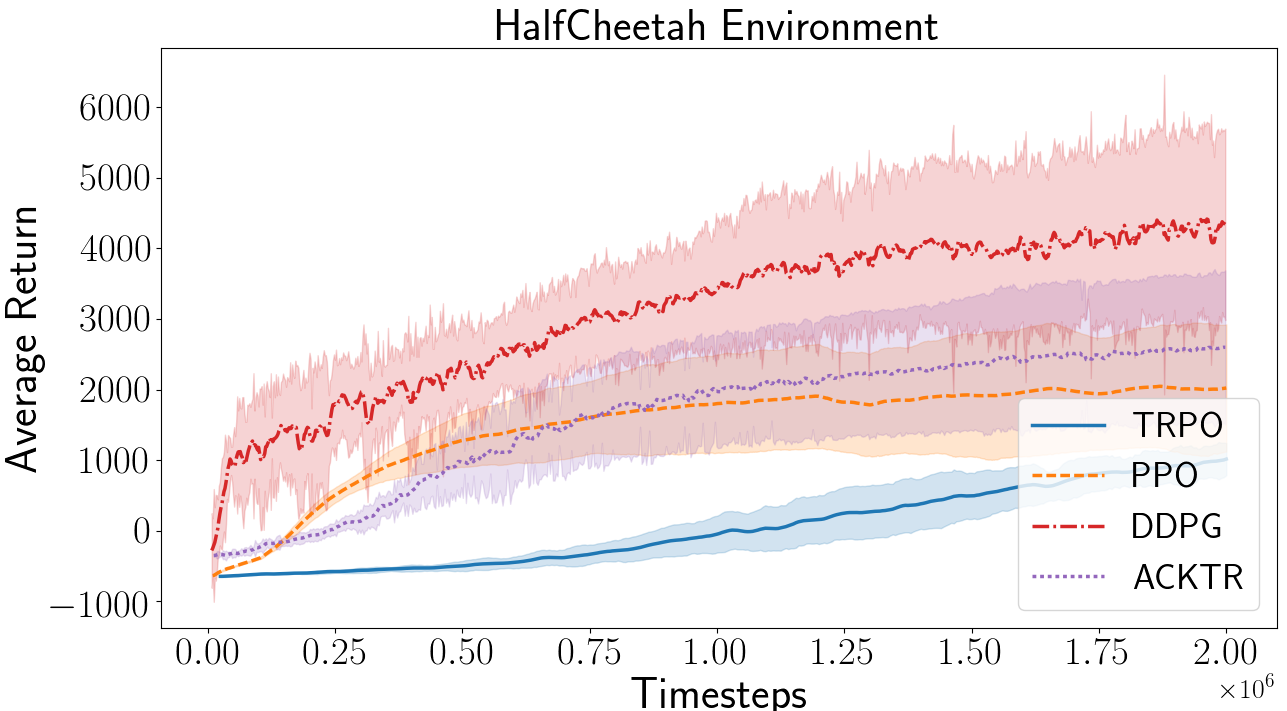}
    \includegraphics[width=0.32\textwidth]{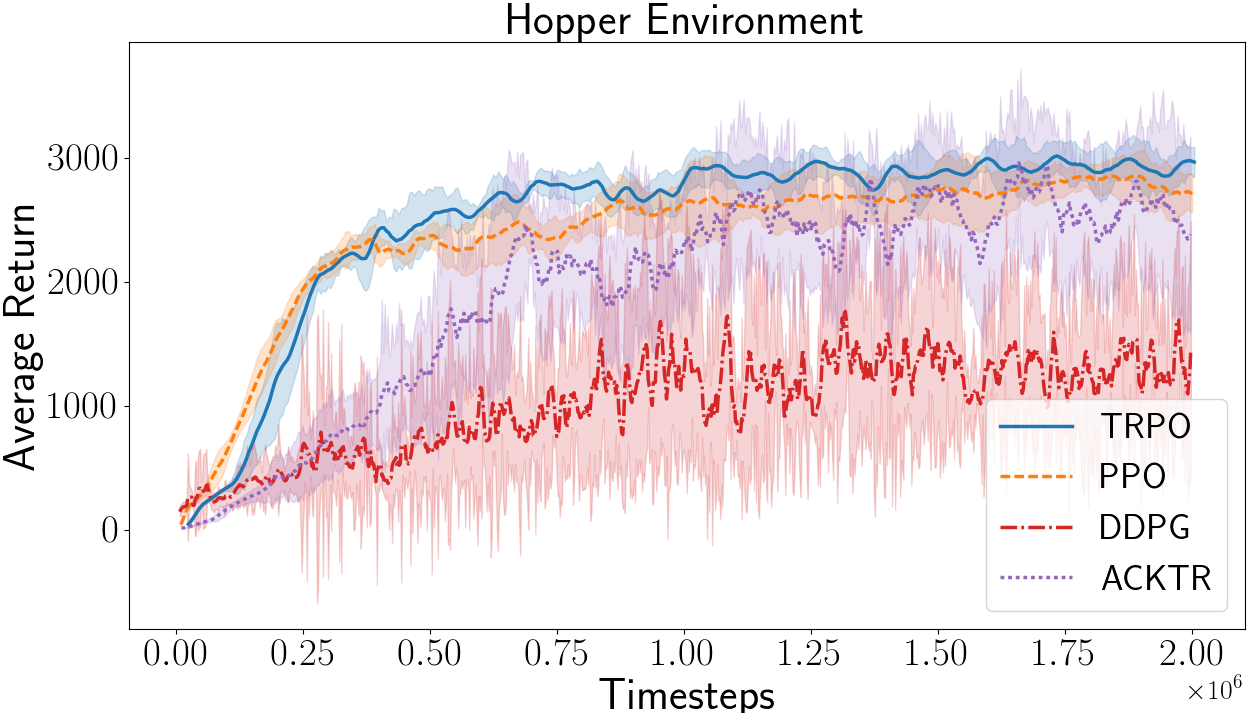}
    \includegraphics[width=0.325\textwidth]{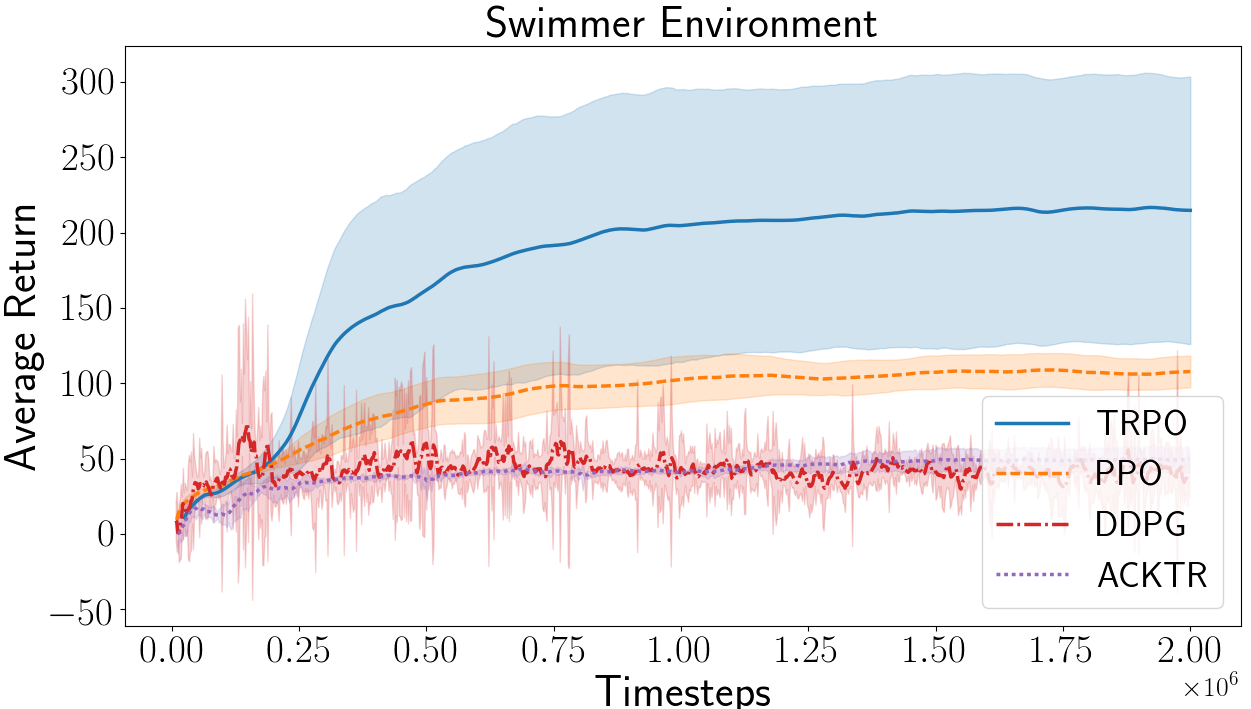}
    \caption{Performance of several policy gradient algorithms across benchmark MuJoCo environment suites}
    \label{fig:env_performance_algorithms}
\end{figure*}

\begin{table*}
    \centering
    \footnotesize{\begin{tabular}{|c|c|c|c|c|}
    \hline
         \textbf{Environment}&\textbf{DDPG}&\textbf{ACKTR}&\textbf{TRPO}&\textbf{PPO}  \\
         \hline
         HalfCheetah-v1&5037 (3664, 6574)&3888 (2288, 5131)&1254.5 (999, 1464)& 3043 (1920, 4165)\\
         \hline
         Hopper-v1&1632 (607, 2370)&2546 (1875, 3217)&2965 (2854, 3076)&2715 (2589, 2847)\\
         \hline
         Walker2d-v1&1582 (901, 2174)&2285 (1246, 3235)&3072 (2957, 3183)&2926 (2514, 3361)\\
         \hline
         Swimmer-v1&31 (21, 46)&50 (42, 55)&214 (141, 287)&107 (101, 118)\\
         \hline
    \end{tabular}}
    \caption{Bootstrap mean and 95\% confidence bounds for a subset of environment experiments. 10k bootstrap iterations and the pivotal method were used. }
    \label{tab:environments_results}
\end{table*}

\subsection{Reward Scale}

\textit{How can the reward scale affect results? Why is reward rescaling used?}
\medskip

\noindent Reward rescaling has been used in several recent works \cite{rllab,QPROP} to improve results for DDPG.
This involves simply multiplying the rewards generated from an environment by some scalar ($\hat{r} = r{\hat{\sigma}}$) for training.
Often, these works report using a reward scale of ${\hat{\sigma}}=0.1$. In Atari domains, this is akin to clipping the rewards to $[0,1]$.
By intuition, in gradient based methods (as used in most deep RL) a large and sparse output scale can result in problems regarding saturation and inefficiency in learning~\cite{lecun2012efficient,glorot2010understanding,vincent2015efficient}. Therefore clipping or rescaling rewards compresses the space of estimated expected returns in action value function based methods such as DDPG. We run a set of experiments using reward rescaling in DDPG (with and without layer normalization) for insights into how this aspect affects performance.

\textbf{Results} Our analysis shows that reward rescaling can have a large effect (full experiment results can be found in the supplemental material), but results were inconsistent across environments and scaling values. Figure~\ref{fig:rewardscale} shows one such example where reward rescaling affects results, causing a failure to learn in small settings below ${\hat{\sigma}}=0.01$. In particular, layer normalization changes how the rescaling factor affects results, suggesting that these impacts are due to the use of deep networks and gradient-based methods. With the value function approximator tracking a moving target distribution, this can potentially affect learning in unstable environments where a deep $Q$-value function approximator is used. Furthermore, some environments may have untuned reward scales (e.g. the HumanoidStandup-v1 of OpenAI gym which can reach rewards in the scale of millions). Therefore, we suggest that this hyperparameter has the potential to have a large impact if considered properly. Rather than rescaling rewards in some environments, a more principled approach should be taken to address this. An initial foray into this problem is made in~\cite{van2016learning}, where the authors adaptively rescale reward targets with normalized stochastic gradient, but further research is needed.

\subsection{Random Seeds and Trials}
\textit{Can random seeds drastically alter performance? Can one distort results by averaging an improper number of trials?}
\medskip

\noindent A major concern with deep RL is the variance in results due to environment stochasticity or stochasticity in the learning process (e.g. random weight initialization). As such, even averaging several learning results together across totally different random seeds can lead to the reporting of misleading results. We highlight this in the form of an experiment.

\begin{figure}
    \centering
    \includegraphics[width=0.43\textwidth]{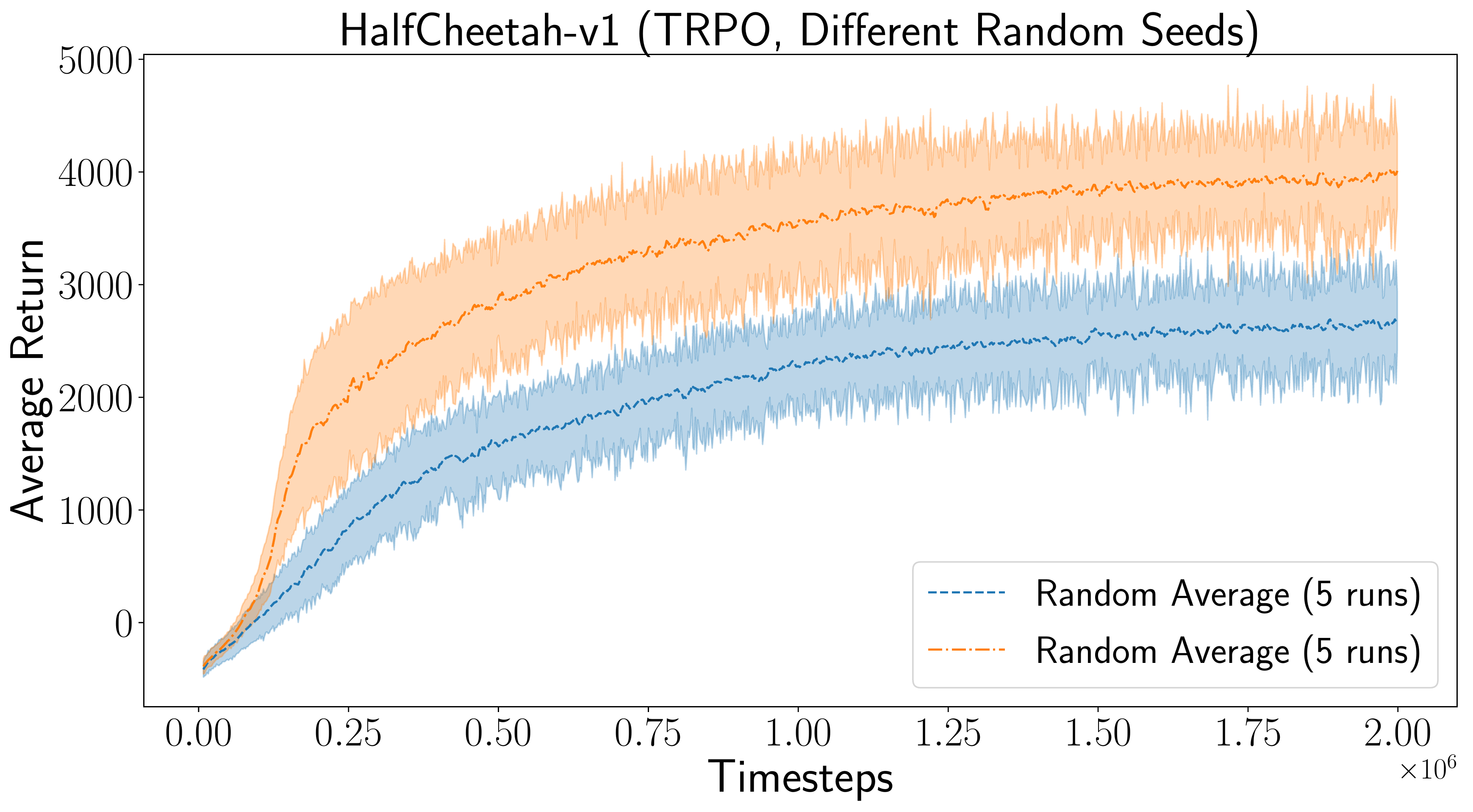}
    \caption{TRPO on HalfCheetah-v1 using the same hyperparameter configurations averaged over two sets of 5 different random seeds each. The average 2-sample $t$-test across entire training distribution resulted in $t = -9.0916$, $p = 0.0016$.}
    \label{fig:randomseeds_trpo}
    \vspace{-2mm}
\end{figure}

\textbf{Results} We perform 10 experiment trials, for the same hyperparameter configuration, only varying the random seed across all 10 trials. We then split the trials into two sets of 5 and average these two groupings together. As shown in Figure~\ref{fig:randomseeds_trpo}, we find that the performance of algorithms can be drastically different. We demonstrate that the variance between runs is enough to create statistically different distributions just from varying random seeds. Unfortunately, in recent reported results, it is not uncommon for the top-$N$ trials to be selected from among several trials~\cite{ACKTR,mnih2016asynchronous} or averaged over only small number of trials ($N<5$)~\cite{IPG,ACKTR}. Our experiment with random seeds shows that this can be potentially misleading. Particularly for HalfCheetah, it is possible to get learning curves that do not fall within the same distribution at all, just by averaging different runs with the same hyperparameters, but different random seeds. While there can be no specific number of trials specified as a recommendation, it is possible that power analysis methods can be used to give a general idea to this extent as we will discuss later. However, more investigation is needed to answer this open problem.

\subsection{Environments}
\textit{How do the environment properties affect variability in reported RL algorithm performance?}
\medskip

\noindent To assess how the choice of evaluation environment can affect the presented results, we use our aforementioned default set of hyperparameters across our chosen testbed of algorithms and investigate how well each algorithm performs across an extended suite of continuous control tasks. For these experiments, we use the following environments from OpenAI Gym: Hopper-v1, HalfCheetah-v1, Swimmer-v1 and Walker2d-v1. The choice of environment often plays an important role in demonstrating how well a new proposed algorithm performs against baselines. In continuous control tasks, often the environments have random stochasticity, shortened trajectories, or different dynamic properties. We demonstrate that, as a result of these differences, algorithm performance can vary across environments and the best performing algorithm across all environments is not always clear. Thus it is increasingly important to present results for a wide range of environments and not only pick those which show a novel work outperforming other methods.

\textbf{Results} As shown in Figure~\ref{fig:env_performance_algorithms}, in environments with stable dynamics (e.g. HalfCheetah-v1), DDPG outperforms all other algorithsm. However, as dynamics become more unstable (e.g. in Hopper-v1) performance gains rapidly diminish. As DDPG is an off-policy method, exploration noise can cause sudden failures in unstable environments. Therefore, learning a proper $Q$-value estimation of expected returns is difficult, particularly since many exploratory paths will result in failure. Since failures in such tasks are characterized by shortened trajectories, a local optimum in this case would be simply to survive until the maximum length of the trajectory (corresponding to one thousand timesteps and similar reward due to a survival bonus in the case of Hopper-v1). As can be seen in Figure~\ref{fig:env_performance_algorithms}, DDPG with Hopper does exactly this. This is a clear example where showing only the favourable and stable HalfCheetah when reporting DDPG-based experiments would be unfair.

Furthermore, let us consider the Swimmer-v1 environment shown in Figure~\ref{fig:env_performance_algorithms}. Here, TRPO significantly outperforms all other algorithms. Due to the dynamics of the water-like environment, a local optimum for the system is to curl up and flail without proper swimming. However, this corresponds to a return of ${\sim} 130$. By reaching a local optimum, learning curves can indicate successful optimization of the policy over time, when in reality the returns achieved are not qualitatively representative of learning the desired behaviour, as demonstrated in video replays of the learned policy\footnote{\url{https://youtu.be/lKpUQYjgm80}}. Therefore, it is important to show not only returns but demonstrations of the learned policy in action. Without understanding what the evaluation returns indicate, it is possible that misleading results can be reported which in reality only optimize local optima rather than reaching the desired behaviour.

\subsection{Codebases}
\textit{Are commonly used baseline implementations comparable?}
\medskip

\noindent In many cases, authors implement their own versions of baseline algorithms to compare against. We investigate the OpenAI baselines implementation of TRPO as used in~\cite{PPO}, the original TRPO code~\cite{TRPO}, and the rllab~\cite{rllab} Tensorflow implementation of TRPO. We also compare the rllab Theano~\cite{rllab}, rllabplusplus~\cite{QPROP}, and OpenAI baselines~\cite{plappert2017parameter} implementations of DDPG. Our goal is to draw attention to the variance due to implementation details across algorithms. We run a subset of our architecture experiments as with the OpenAI baselines implementations using the same hyperparameters as in those experiments\footnote{Differences are discussed in the supplemental (e.g. use of different optimizers for the value function baseline). Leaky ReLU activations are left out to narrow the experiment scope.}.

\begin{figure}
    \centering
    \includegraphics[width=.36\textwidth]{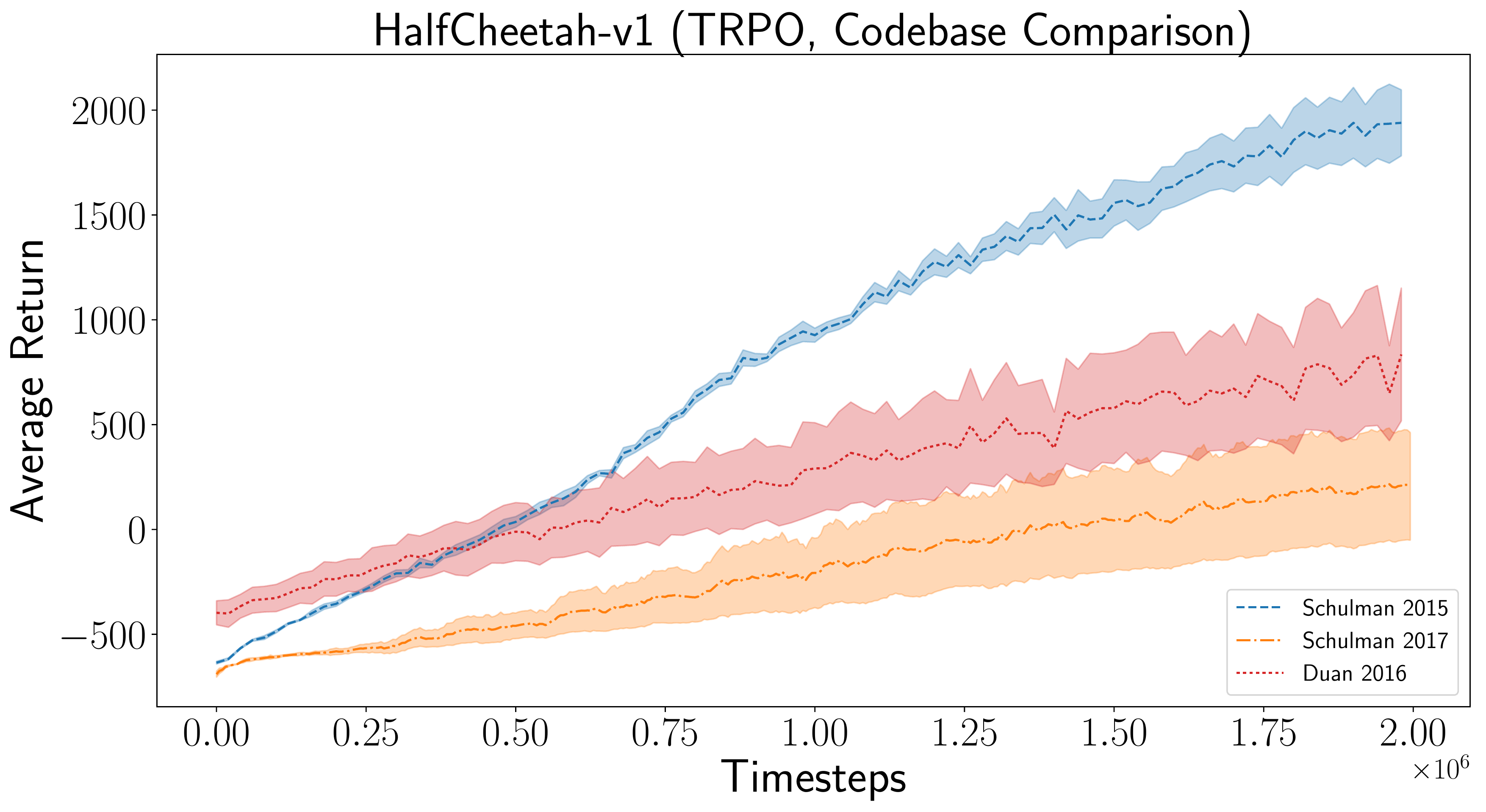}
    \includegraphics[width=.36\textwidth]{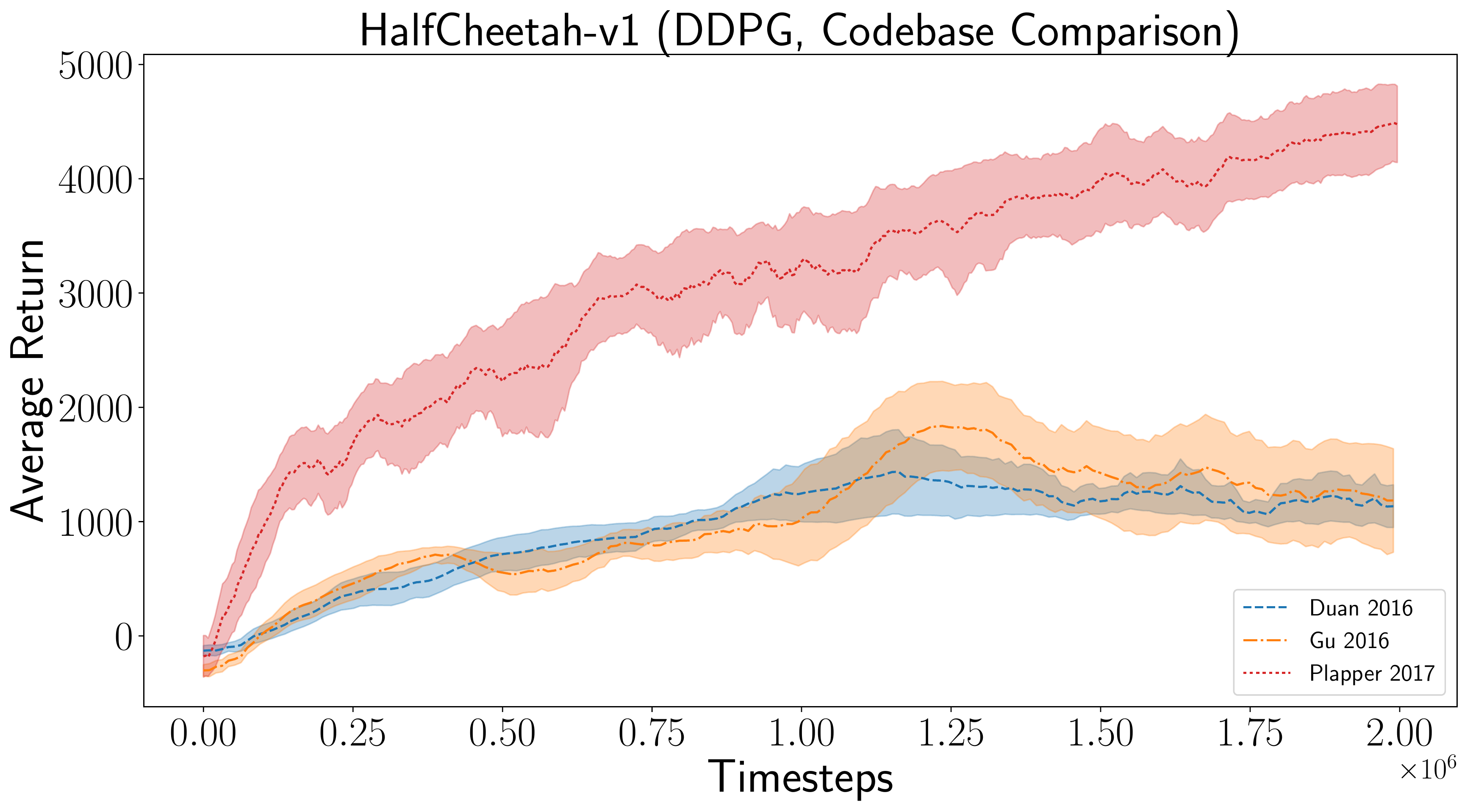}
    \caption{TRPO codebase comparison using our default set of hyperparameters (as used in other experiments).}
    \label{fig:trpo_codebase_diff}
    \vspace{-.5mm}
\end{figure}

\textbf{Results} We find that implementation differences which are often not reflected in publications can have dramatic impacts on performance. This can be seen for our final evaluation performance after training on 2M samples in Tables~\ref{tab:architecture_results} and~\ref{tab:architecture_results_vf}, as well as a sample comparison in Figure~\ref{fig:trpo_codebase_diff}. This demonstrates the necessity that implementation details be enumerated, codebases packaged with publications, and that performance of baseline experiments in novel works matches the original baseline publication code.

% SUGGESTIONS ON REPORTING OF RESULTS
\section{Reporting Evaluation Metrics}
\label{sec:result_presentation}
In this section we analyze some of the evaluation metrics commonly used in the reinforcement learning literature. In practice, RL algorithms are often evaluated by simply presenting plots or tables of average cumulative reward (average returns) and, more recently, of maximum reward achieved over a fixed number of timesteps. Due to the unstable nature of many of these algorithms, simply reporting the maximum returns is typically inadequate for fair comparison; even reporting average returns can be misleading as the range of performance across seeds and trials is unknown. Alone, these may not provide a clear picture of an algorithm's range of performance. However, when combined with confidence intervals, this may be adequate to make an informed decision given a large enough number of trials. As such, we investigate using the bootstrap and significance testing as in ML~\cite{kohavi1995study,bouckaert2004evaluating,nadeau2000inference} to evaluate algorithm performance.

\textbf{Online View vs. Policy Optimization} An important distinction when reporting results is the online learning view versus the policy optimization view of RL. In the online view, an agent will optimize the returns across the entire learning process and there is not necessarily an end to the agent's trajectory. In this view, evaluations can use the average cumulative rewards across the entire learning process (balancing exploration and exploitation) as in~\cite{hofer2016online}, or can possibly use offline evaluation as in~\cite{mandel2016offline}. The alternate view corresponds to policy optimization, where evaluation is performed using a target policy in an offline manner. In the policy optimization view it is important to run evaluations across the entire length of the task trajectory with a single target policy to determine the average returns that the target can obtain. We focus on evaluation methods for the policy optimization view (with offline evaluation), but the same principles can be applied to the online view.

\textbf{Confidence Bounds} The sample bootstrap has been a popular method to gain insight into a population distribution from a smaller sample~\cite{efron1994introduction}. Bootstrap methods are particularly popular for A/B testing, and we can borrow some ideas from this field. Generally a bootstrap estimator is obtained by resampling with replacement many times to generate a  statistically relevant mean and confidence bound. Using this technique, we can gain insight into what is the 95\% confidence interval of the results from our section on environments. Table~\ref{tab:environments_results} shows the bootstrap mean and 95\% confidence bounds on our environment experiments. Confidence intervals can vary wildly between algorithms and environments. We find that TRPO and PPO are the most stable with small confidence bounds from the bootstrap. In cases where confidence bounds are exceedingly large, it may be necessary to run more trials (i.e.~increase the sample size).

\textbf{Power Analysis} Another method to determine if the sample size must be increased is bootstrap power analysis~\cite{tuffery2011data,yuan2003bootstrap}. If we use our sample and give it some uniform lift (for example, scaling uniformly by 1.25), we can run many bootstrap simulations and determine what percentage of the simulations result in statistically significant values with the lift. If there is a small percentage of significant values, a larger sample size is needed (more trials must be run). We do this across all environment experiment trial runs and indeed find that, in more unstable settings, the bootstrap power percentage leans towards insignificant results in the lift experiment. Conversely, in stable trials (e.g.~TRPO on Hopper-v1) with a small sample size, the lift experiment shows that no more trials are needed to generate significant comparisons. These results are provided in the supplemental material.

\textbf{Significance} An important factor when deciding on an RL algorithm to use is the significance of the reported gains based on a given metric. Several works have investigated the use of significance metrics to assess the reliability of reported evaluation metrics in ML. However, few works in reinforcement learning assess the significance of reported metrics. Based on our experimental results which indicate that algorithm performance can vary wildly based simply on perturbations of random seeds, it is clear that some metric is necessary for assessing the significance of algorithm performance gains and the confidence of reported metrics. While more research and investigation is needed to determine the best metrics for assessing RL algorithms, we investigate an initial set of metrics based on results from ML.

In supervised learning, $k$-fold $t$-test, corrected resampled $t$-test, and other significance metrics have been discussed when comparing machine learning results~\cite{bouckaert2004evaluating,nadeau2000inference}. However, the assumptions pertaining to the underlying data with corrected metrics do not necessarily apply in RL. Further work is needed to investigate proper corrected significance tests for RL. Nonetheless, we explore several significance measures which give insight into whether a novel algorithm is truly performing as the state-of-the-art. We consider the simple 2-sample $t$-test (sorting all final evaluation returns across $N$ random trials with different random seeds); the Kolmogorov-Smirnov test~\cite{wilcox2005kolmogorov}; and bootstrap percent differences with 95\% confidence intervals. All calculated metrics can be found in the supplemental. Generally, we find that the significance values match up to what is to be expected. Take, for example, comparing Walker2d-v1 performance of ACKTR vs. DDPG. ACKTR performs slightly better, but this performance is not significant due to the overlapping confidence intervals of the two: $t=1.03,p=0.334$, $KS=0.40,p=0.697$, bootstrapped percent difference 44.47\% (-80.62\%, 111.72\%).

% DETAILED DISCUSSION
\section{Discussion and Conclusion}
\label{sec:discussion}
Through experimental methods focusing on PG methods for continuous control, we investigate problems with reproducibility in deep RL. We find that both intrinsic (e.g. random seeds, environment properties) and extrinsic sources (e.g. hyperparameters, codebases) of non-determinism can contribute to difficulties in reproducing baseline algorithms. Moreover, we find that highly varied results due to intrinsic sources bolster the need for using proper significance analysis. We propose several such methods and show their value on a subset of our experiments.
\medskip

\noindent \textit{What recommendations can we draw from our experiments?}
\medskip

\noindent Based on our experimental results and investigations, we can provide some general recommendations. Hyperparameters can have significantly different effects across algorithms and environments. Thus it is important to find the working set which at least matches the original reported performance of baseline algorithms through standard hyperparameter searches. Similarly, new baseline algorithm implementations used for comparison should match the original codebase results if available. Overall, due to the high variance across trials and random seeds of reinforcement learning algorithms, many trials must be run with different random seeds when comparing performance. Unless random seed selection is explicitly part of the algorithm, averaging multiple runs over different random seeds gives insight into the population distribution of the algorithm performance on an environment. Similarly, due to these effects, it is important to perform proper significance testing to determine if the higher average returns are in fact representative of better performance.

We highlight several forms of significance testing and find that they give generally expected results when taking confidence intervals into consideration. Furthermore, we demonstrate that bootstrapping and power analysis are possible ways to gain insight into the number of trial runs necessary to make an informed decision about the significance of algorithm performance gains. In general, however, the most important step to reproducibility is to report \textit{all} hyperparameters, implementation details, experimental setup, and evaluation methods for both baseline comparison methods and novel work. Without the publication of implementations and related details, wasted effort on reproducing state-of-the-art works will plague the community and slow down progress.
\medskip

\noindent \textit{What are possible future lines of investigation?}
\medskip

\noindent Due to the significant effects of hyperparameters (particularly reward scaling), another possibly important line of future investigation is in building hyperparameter agnostic algorithms. Such an approach would ensure that there is no unfairness introduced from external sources when comparing algorithms agnostic to parameters such as reward scale, batch size, or network structure. Furthermore, while we investigate an initial set of significance metrics here, they may not be the best fit for comparing RL algorithms. Several works have begun investigating policy evaluation methods for the purposes of safe RL~\cite{thomas2016data,thomas2015high}, but further work is needed in significance testing and statistical analysis. Similar lines of investigation to~\cite{nadeau2000inference,bouckaert2004evaluating} would be helpful to determine the best methods for evaluating performance gain significance.
\medskip

\noindent \textit{How can we ensure that deep RL matters?}
\medskip

\noindent We discuss many different factors affecting reproducibility of RL algorithms. The sensitivity of these algorithms to changes in reward scale, environment dynamics, and random seeds can be considerable and varies between algorithms and settings. Since benchmark environments are proxies for real-world applications to gauge generalized algorithm performance, perhaps more emphasis should be placed on the applicability of RL algorithms to real-world tasks. That is, as there is often no clear winner among all benchmark environments, perhaps recommended areas of application should be demonstrated along with benchmark environment results when presenting a new algorithm. Maybe new methods should be answering the question: in what setting would this work be useful? This is something that is addressed for machine learning in~\cite{wagstaff2012machine} and may warrant more discussion for RL. As a community, we must not only ensure reproducible results with fair comparisons, but we must also consider what are the best ways to demonstrate that RL continues to matter.

\section{Acknowledgements}
We thank NSERC, CIFAR, the Open Philanthropy Project, and the AWS Cloud Credits for Research Program. %for their generous contributions.

% REFERENCES
\fontsize{9.0pt}{10.0pt} \selectfont
\bibliographystyle{aaai}
\bibliography{bibliography}

\onecolumn
\section{Supplemental Material}
\label{sec:appendix}
\input{appendix}

\end{document}

%% file: appendix.tex
\pdfoutput=1

In this supplemental material, we include a detailed review of experiment configurations of related work with policy gradient methods in continuous control MuJoCo~\cite{MuJoCo} environment tasks from OpenAI Gym~\cite{gym}. We include a detailed list of the hyperparameters and reported metrics typically used in policy gradient literature in deep RL. We also include all our experimental results, with baseline algorithms DDPG \cite{DDPG}, TRPO \cite{TRPO}, PPO \cite{PPO} and ACKTR \cite{ACKTR}) as discussed in the paper. Our experimental results include figures with different hyperparameters (network architectures, activation functions) to highlight the differences this can have across algorithms and environments.  Finally, as discussed in the paper, we include discussion of significance metrics and show how these metrics can be useful for evaluating deep RL algorithms.

\section{Literature Reviews}

\subsection{Hyperparameters}

In this section, we include a list of hyperparameters that are reported in related literature, as shown in figure \ref{related_work_hyperparams}.  Our analysis shows that often there is no consistency in the type of network architectures and activation functions that are used in related literature. As shown in the paper and from our experimental results in later sections, we find, however, that these hyperparameters can have a significant effect in the performance of algorithms across benchmark environments typically used.

\begin{table}[H]
\centering
\caption{Evaluation Hyperparameters of baseline algorithms reported in related literature}
\label{related_work_hyperparams}
\begin{tabular}{|l|l|l|l|l|l|l|}
\hline
\textbf{\begin{tabular}[c]{@{}l@{}}Related Work\\ (Algorithm)\end{tabular}} & \textbf{\begin{tabular}[c]{@{}l@{}}Policy \\ Network\end{tabular}} & \textbf{\begin{tabular}[c]{@{}l@{}}Policy \\ Network \\ Activation\end{tabular}} & \textbf{\begin{tabular}[c]{@{}l@{}}Value \\ Network\end{tabular}} & \textbf{\begin{tabular}[c]{@{}l@{}}Value\\ Network\\ Activation\end{tabular}} & \textbf{\begin{tabular}[c]{@{}l@{}}Reward\\ Scaling\end{tabular}} & \textbf{\begin{tabular}[c]{@{}l@{}}Batch \\ Size\end{tabular}} \\ \hline
DDPG & 64x64 & ReLU & 64x64 & ReLU & 1.0 & 128 \\ \hline
TRPO & 64x64 & TanH & 64x64 & TanH & - & 5k \\ \hline
PPO & 64x64 & TanH & 64x64 & TanH & - & 2048 \\ \hline
ACKTR & 64x64 & TanH & 64x64 & ELU & - & 2500 \\ \hline
\begin{tabular}[c]{@{}l@{}}Q-Prop\\ (DDPG)\end{tabular} & 100x50x25 & TanH & 100x100 & ReLU & 0.1 & 64 \\ \hline
\begin{tabular}[c]{@{}l@{}}Q-Prop\\ (TRPO)\end{tabular} & 100x50x25 & TanH & 100x100 & ReLU & - & 5k \\ \hline
\begin{tabular}[c]{@{}l@{}}IPG\\ (TRPO)\end{tabular} & 100x50x25 & TanH & 100x100 & ReLU & - & 10k \\ \hline
\begin{tabular}[c]{@{}l@{}}Param Noise\\ (DDPG)\end{tabular} & 64x64 & ReLU & 64x64 & ReLU & - & 128 \\ \hline
\begin{tabular}[c]{@{}l@{}}Param Noise\\ (TRPO)\end{tabular} & 64x64 & TanH & 64x64 & TanH & - & 5k \\ \hline
\begin{tabular}[c]{@{}l@{}}Benchmarking\\ (DDPG)\end{tabular} & 400x300 & ReLU & 400x300 & ReLU & 0.1 & 64 \\ \hline
\begin{tabular}[c]{@{}l@{}}Benchmarking\\ (TRPO)\end{tabular} & 100x50x25 & TanH & 100x50x25 & TanH & - & 25k \\ \hline
\end{tabular}
\end{table}

\subsection{Reported Results on Benchmarked Environments}

We then demonstrate how experimental reported results, on two different environments (HalfCheetah-v1 and Hopper-v1) can vary across different related work that uses these algorithms for baseline comparison. We further show the results we get, using the same hyperparameter configuration, but using two different codebase implementations (note that these implementations are often used as baseline codebase to develop algorithms). We highlight that, depending on the codebase used, experimental results can vary significantly.

\begin{table}[H]
\centering
\caption{Comparison with Related Reported Results with Hopper Environment}
\label{tab:comparison_baseline_results2}
\small{\begin{tabular}{llllllll}
Environment & Metric & rllab  & QProp & IPG  & TRPO & \begin{tabular}[c]{@{}l@{}}Our Results\\ (rllab)\end{tabular} & \begin{tabular}[c]{@{}l@{}}Our Results \\ (Baselines)\end{tabular}\\
\multirow{3}{*}{\begin{tabular}[c]{@{}l@{}}TRPO on\\ Hopper\\ Environment\end{tabular}}
& Number of Iterations & 500    & 500   & 500  & 500  & 500 & 500 \\
& Average Return & 1183.3  &  -  &  -   &   -  & 2021.34  &  2965.3\\
& Max Average Return & - & 2486   &  & 3668.8  &  3229.1  &  3034.4\\
\end{tabular}}
\end{table}

\begin{table}[H]
\centering
\caption{Comparison with Related Reported Results with HalfCheetah Environment}
\label{tab:comparison_baseline_results}
\small{\begin{tabular}{llllllll}
Environment & Metric & rllab  & QProp & IPG  & TRPO & \begin{tabular}[c]{@{}l@{}}Our Results\\ (rllab)\end{tabular} & \begin{tabular}[c]{@{}l@{}}Our Results \\ (Baselines)\end{tabular}\\
\multirow{3}{*}{\begin{tabular}[c]{@{}l@{}}TRPO on\\ HalfCheetah\\ Environment\end{tabular}}
& Number of Iterations & 500    & 500   & 500  & 500  & 500 & 500 \\
& Average Return & 1914.0 &    & -    & -    & 3576.08 & 1045.6 \\
& Max Average Return   & -      & 4734  & 2889 & 4855 & 5197 & 1045.6\\
\end{tabular}}
\end{table}

\begin{table}[H]
    \centering
    \begin{tabular}{c|c}
         \hline
         Work & Number of Trials \\
         \hline
         \cite{mnih2016asynchronous}& top-5\\
         \cite{PPO} & 3-9 \\
         \cite{rllab} & 5 (5)\\
         \cite{IPG} & 3\\
         \cite{DDPG} & 5\\
         \cite{TRPO} & 5\\
         \cite{ACKTR} & top-2, top-3\\
         \hline
    \end{tabular}
    \caption{Number of trials reported during evaluation in various works.}
    \label{tab:trials}
\end{table}

\subsection{Reported Evaluation Metrics in Related Work}

In table \ref{reported_eval_metric} we show the evaluation metrics, and reported results in further details across related work.

\begin{table}[H]
\centering
\caption{Reported Evaluation Metrics of baseline algorithms in related literature}
\label{reported_eval_metric}
\begin{tabular}{|c|c|c|c|c|c|}
\hline
\textbf{\begin{tabular}[c]{@{}c@{}}Related Work\\ (Algorithm)\end{tabular}} & \textbf{Environments} & \textbf{\begin{tabular}[c]{@{}c@{}}Timesteps\\ or Episodes\\ or Iterations\end{tabular}} & \multicolumn{3}{c|}{\textbf{Evaluation Metrics}} \\ \hline
 &  &  & \textbf{\begin{tabular}[c]{@{}c@{}}Average \\ Return\end{tabular}} & \textbf{\begin{tabular}[c]{@{}c@{}}Max \\ Return\end{tabular}} & \textbf{\begin{tabular}[c]{@{}c@{}}Std \\ Error\end{tabular}} \\ \hline
PPO & \begin{tabular}[c]{@{}c@{}}HalfCheetah\\ Hopper\end{tabular} & 1M & \begin{tabular}[c]{@{}c@{}}$\sim$1800\\ $\sim$2200\end{tabular} & - & - \\ \hline
ACKTR & \begin{tabular}[c]{@{}c@{}}HalfCheetah\\ Hopper\end{tabular} & 1M & \begin{tabular}[c]{@{}c@{}}$\sim$2400\\ $\sim$3500\end{tabular} & - & - \\ \hline
\begin{tabular}[c]{@{}c@{}}Q-Prop\\ (DDPG)\end{tabular} & \begin{tabular}[c]{@{}c@{}}HalfCheetah\\ Hopper\end{tabular} & 6k (eps) & \begin{tabular}[c]{@{}c@{}}$\sim$6000\\ -\end{tabular} & \begin{tabular}[c]{@{}c@{}}7490\\ 2604\end{tabular} & \begin{tabular}[c]{@{}c@{}}-\\ -\end{tabular} \\ \hline
\begin{tabular}[c]{@{}c@{}}Q-Prop\\ (TRPO)\end{tabular} & \begin{tabular}[c]{@{}c@{}}HalfCheetah\\ Hopper\end{tabular} & 5k (timesteps) & \begin{tabular}[c]{@{}c@{}}$\sim$4000\\ -\end{tabular} & \begin{tabular}[c]{@{}c@{}}4734\\ 2486\end{tabular} & \begin{tabular}[c]{@{}c@{}}-\\ -\end{tabular} \\ \hline
\begin{tabular}[c]{@{}c@{}}IPG\\ (TRPO)\end{tabular} & \begin{tabular}[c]{@{}c@{}}HalfCheetah\\ Hopper\end{tabular} & 10k (eps) & \begin{tabular}[c]{@{}c@{}}$\sim$3000\\ -\end{tabular} & 2889 & \begin{tabular}[c]{@{}c@{}}-\\ -\end{tabular} \\ \hline
\begin{tabular}[c]{@{}c@{}}Param Noise\\ (DDPG)\end{tabular} & \begin{tabular}[c]{@{}c@{}}HalfCheetah\\ Hopper\end{tabular} & 1M & \begin{tabular}[c]{@{}c@{}}$\sim$1800\\ $\sim$500\end{tabular} & \begin{tabular}[c]{@{}c@{}}-\\ -\end{tabular} & \begin{tabular}[c]{@{}c@{}}-\\ -\end{tabular} \\ \hline
\begin{tabular}[c]{@{}c@{}}Param Noise\\ (TRPO)\end{tabular} & \begin{tabular}[c]{@{}c@{}}HalfCheetah\\ Hopper\end{tabular} & 1M & \begin{tabular}[c]{@{}c@{}}$\sim$3900\\ $\sim$2400\end{tabular} & \begin{tabular}[c]{@{}c@{}}-\\ -\end{tabular} & \begin{tabular}[c]{@{}c@{}}-\\ -\end{tabular} \\ \hline
\begin{tabular}[c]{@{}c@{}}Benchmarking\\ (DDPG)\end{tabular} & \begin{tabular}[c]{@{}c@{}}HalfCheetah\\ Hopper\end{tabular} & \begin{tabular}[c]{@{}c@{}}500 iters\\ (25k eps)\end{tabular} & \begin{tabular}[c]{@{}c@{}}$\sim$2148\\ $\sim$267\end{tabular} & \begin{tabular}[c]{@{}c@{}}-\\ -\end{tabular} & \begin{tabular}[c]{@{}c@{}}702\\ 43\end{tabular} \\ \hline
\begin{tabular}[c]{@{}c@{}}Benchmarking\\ (TRPO)\end{tabular} & \begin{tabular}[c]{@{}c@{}}HalfCheetah\\ Hopper\end{tabular} & \begin{tabular}[c]{@{}c@{}}500 iters\\ (925k eps)\end{tabular} & \begin{tabular}[c]{@{}c@{}}$\sim$1914\\ $\sim$1183\end{tabular} & \begin{tabular}[c]{@{}c@{}}-\\ -\end{tabular} & \begin{tabular}[c]{@{}c@{}}150\\ 120\end{tabular} \\ \hline
\end{tabular}
\end{table}

\section{Experimental Setup}

In this section, we show detailed analysis of our experimental results, using same hyperparameter configurations used in related work.  Experimental results are included for the OpenAI Gym~\cite{gym} Hopper-v1 and HalfCheetah-v1 environments, using the  policy gradient algorithms including DDPG, TRPO, PPO and ACKTR. Our experiments are done using the available codebase from OpenAI rllab \cite{rllab} and OpenAI Baselines. Each of our experiments are performed over 5 experimental trials with different random seeds, and results averaged over all trials. Unless explicitly specified as otherwise (such as in hyperparameter modifications where we alter a hyperparameter under investigation), hyperparameters were as follows. All results (including graphs) show mean and standard error across random seeds.

\begin{itemize}
    \item DDPG
    \begin{itemize}
        \item Policy Network: (64, relu, 64, relu, tanh); Q Network (64, relu, 64, relu, linear)
        \item Normalized observations with running mean filter
        \item Actor LR: $1e-4$; Critic LR: $1e-3$
        \item Reward Scale: 1.0
        \item Noise type: O-U 0.2
        \item Soft target update $\tau=.01$
        \item $\gamma=0.995$
        \item batch size = 128
        \item Critic L2 reg $1e-2$
    \end{itemize}
    \item PPO
    \begin{itemize}
        \item Policy Network: (64, tanh, 64, tanh, Linear) + Standard Deviation variable; Value Network (64, tanh, 64, tanh, linear)
        \item Normalized observations with running mean filter
        \item Timesteps per batch 2048
        \item clip param = 0.2
        \item entropy coeff = 0.0
        \item Optimizer epochs per iteration = 10
        \item Optimizer step size $3e-4$
        \item Optimizer batch size 64
        \item Discount $\gamma=0.995$, GAE $\lambda=0.97$
        \item learning rate schedule is constant
    \end{itemize}
    \item TRPO
    \begin{itemize}
        \item Policy Network: (64, tanh, 64, tanh, Linear) + Standard Deviation variable; Value Network (64, tanh, 64, tanh, linear)
        \item Normalized observations with running mean filter
        \item Timesteps per batch 5000
        \item max KL=0.01
        \item Conjugate gradient iterations = 20
        \item CG damping = 0.1
        \item VF Iterations = 5
        \item VF Batch Size = 64
        \item VF Step Size = $1e-3$
        \item entropy coeff = 0.0
        \item Discount $\gamma=0.995$, GAE $\lambda=0.97$
    \end{itemize}
    \item ACKTR
    \begin{itemize}
        \item Policy Network: (64, tanh, 64, tanh, Linear) + Standard Deviation variable; Value Network (64, elu, 64, elu, linear)
        \item Normalized observations with running mean filter
        \item Timesteps per batch 2500
        \item desired KL = .002
        \item Discount $\gamma=0.995$, GAE $\lambda=0.97$
    \end{itemize}
\end{itemize}

\subsection{Modifications to Baseline Implementations}
To ensure fairness of comparison, we make several modifications to the existing implementations. First, we change evaluation in DDPG~\cite{plappert2017parameter} such that during evaluation at the end of an epoch, 10 full trajectories are evaluated. In the current implementation, only a partial trajectory is evaluated immediately after training such that a full trajectory will be evaluated across several different policies, this corresponds more closely to the online view of evaluation, while we take a policy optimization view when evaluating algorithms.

\subsection{Hyperparameters : Network Structures and Activation Functions}
\noindent
Below, we examine the significance of the network configurations used for the non-linear function approximators in policy gradient methods. Several related work have used different sets of network configurations (network sizes and activation functions). We use the reported network configurations from other works, and demonstrate the significance of careful fine tuning that is required.  We demonstrate results using the network activation functions, ReLU, TanH and Leaky ReLU, where most papers use ReLU and TanH as activation functions without detailed reporting of the effect of these activation functions. We analyse the signifcance of using different activations in the policy and action value networks. Previously, we included a detailed table showing average reward with standard error obtained for each of the hyperparameter configurations. In the results below, we show detailed results of how each of these policy gradient algorithms are affected by the choice of the network configuration.

\subsection{Proximal Policy Optimization (PPO)}

\begin{figure}[H]
    \centering
    \includegraphics[width=.49\textwidth]{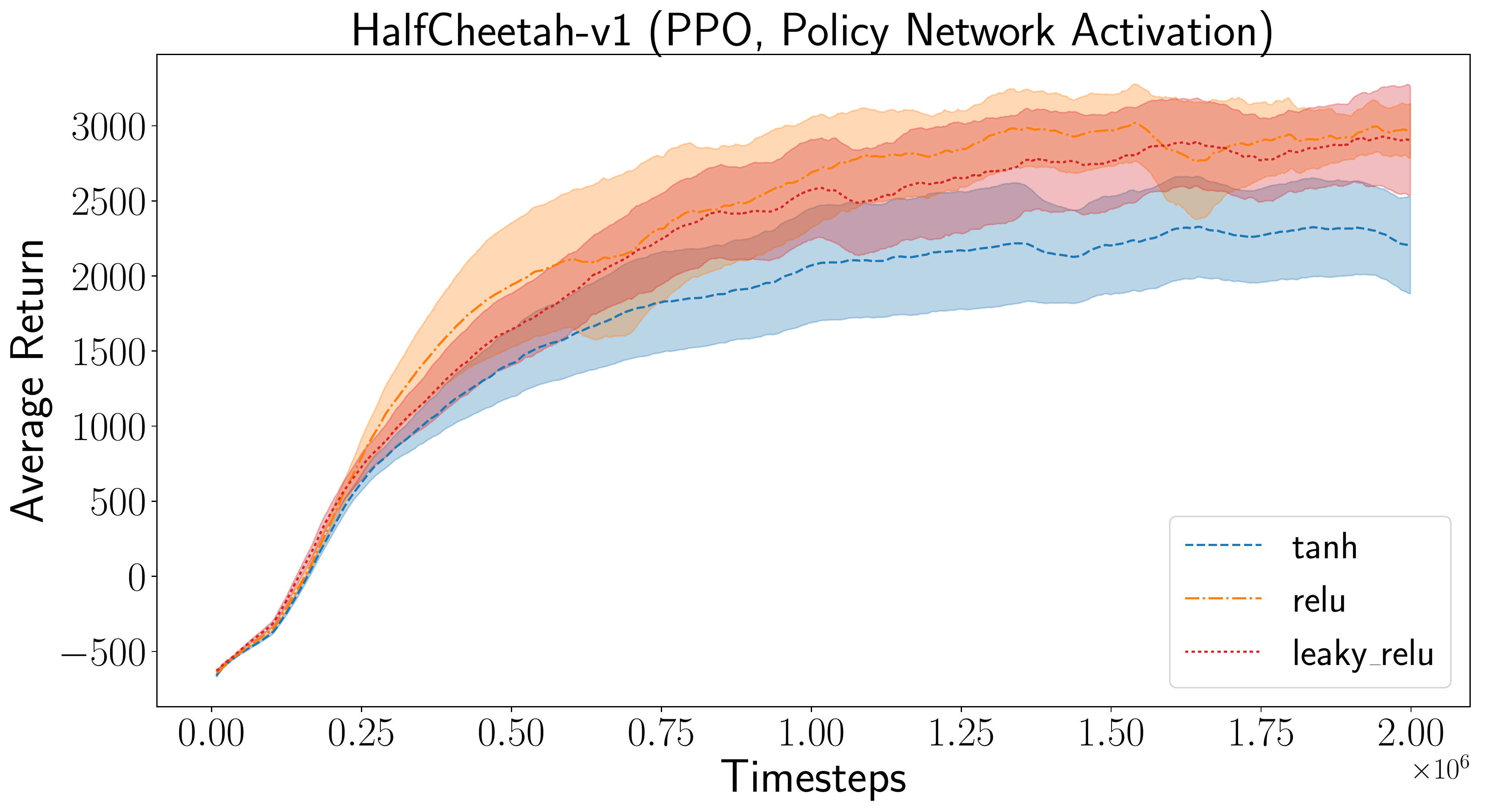}
    \includegraphics[width=.49\textwidth]{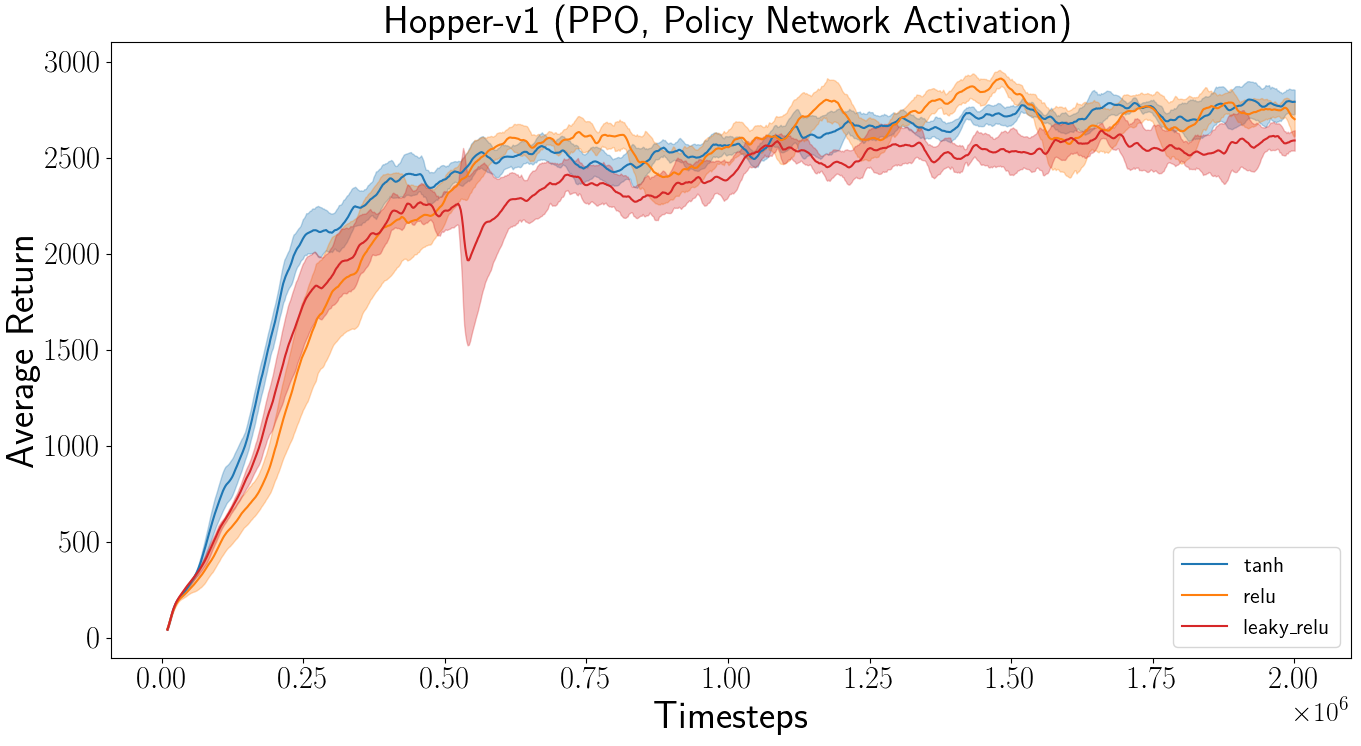}
    \includegraphics[width=.49\textwidth]{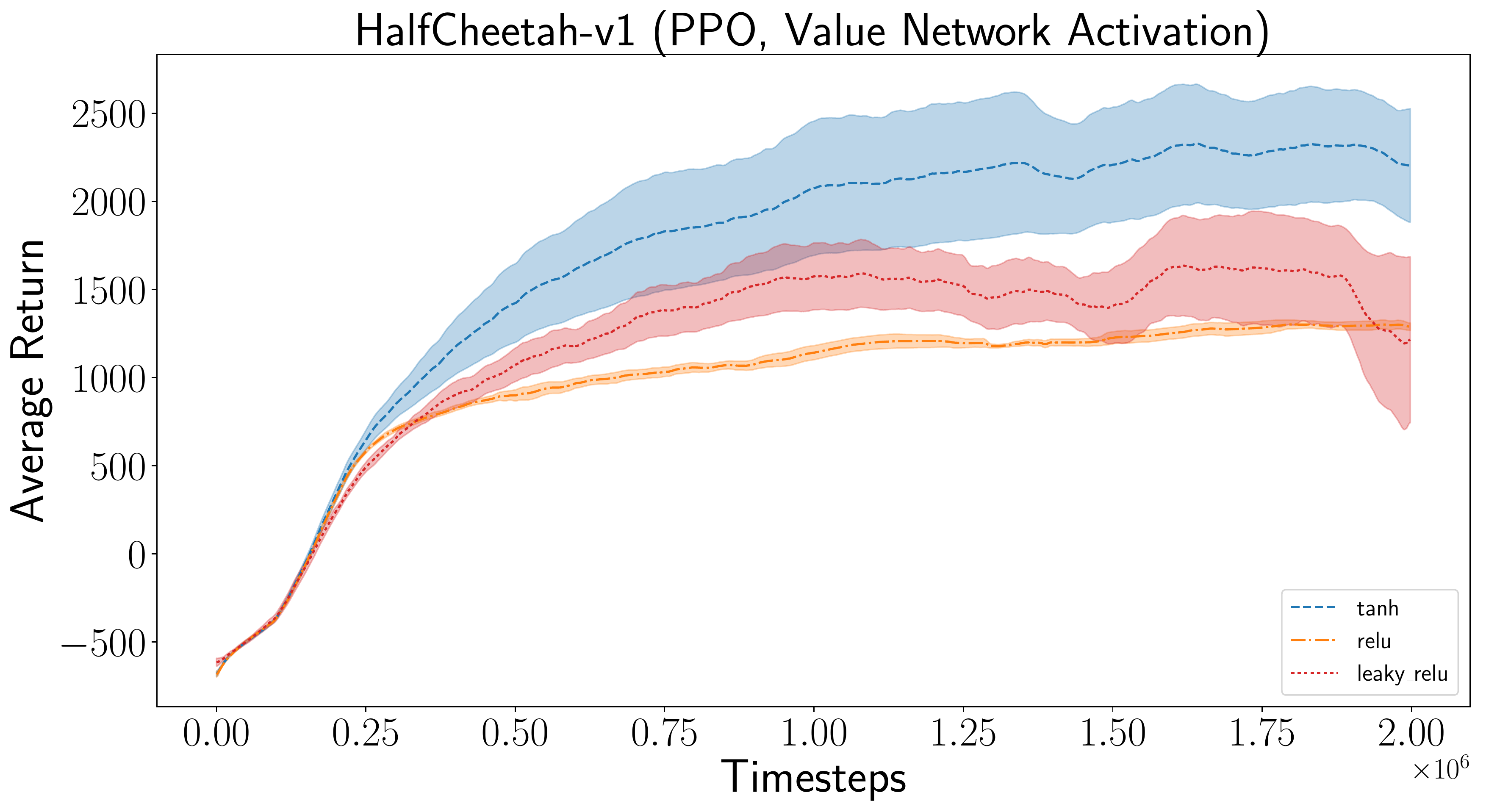}
    \includegraphics[width=.49\textwidth]{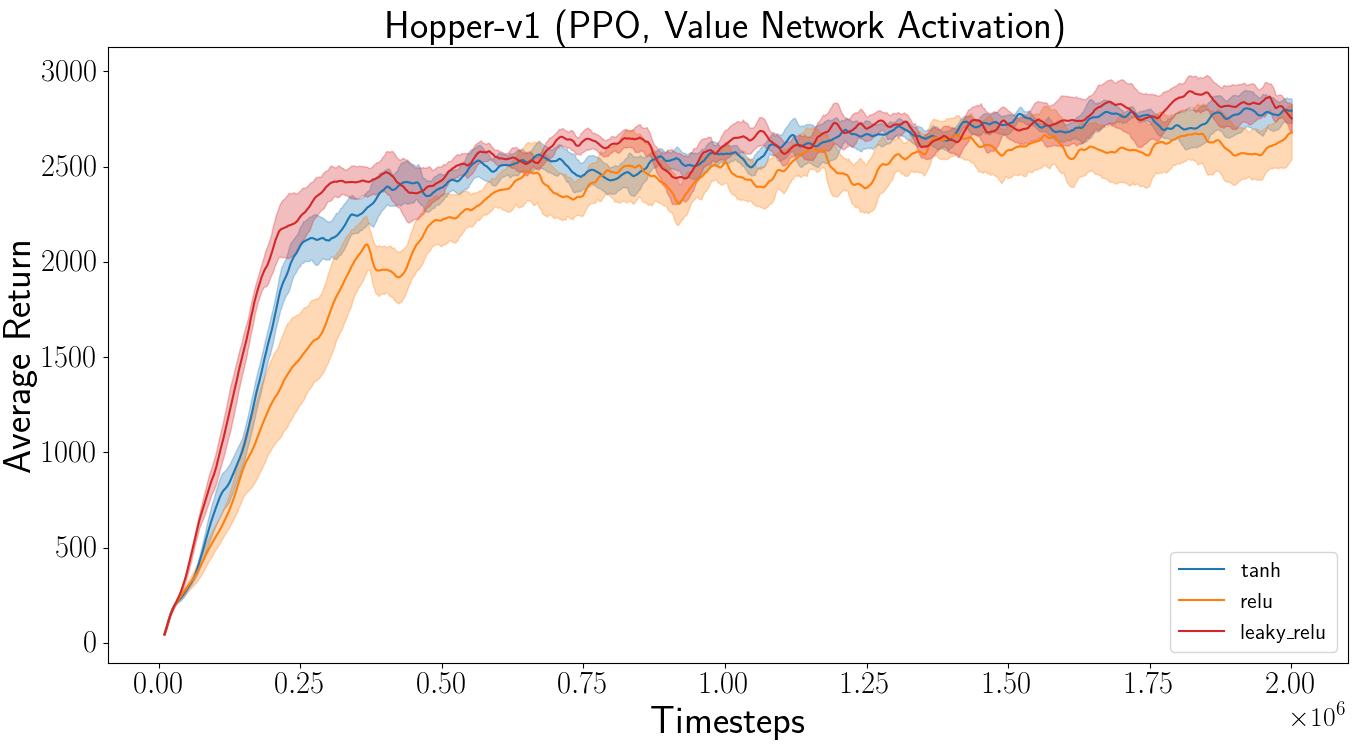}

    \caption{PPO Policy and Value Network activation}
    \label{fig:pponetworkact}
\end{figure}

\begin{figure}[H]
    \centering
    \includegraphics[width=.49\textwidth]{"images/HalfCheetah-v1__PPO,_Policy_Network_Structure_"}
    \includegraphics[width=.49\textwidth]{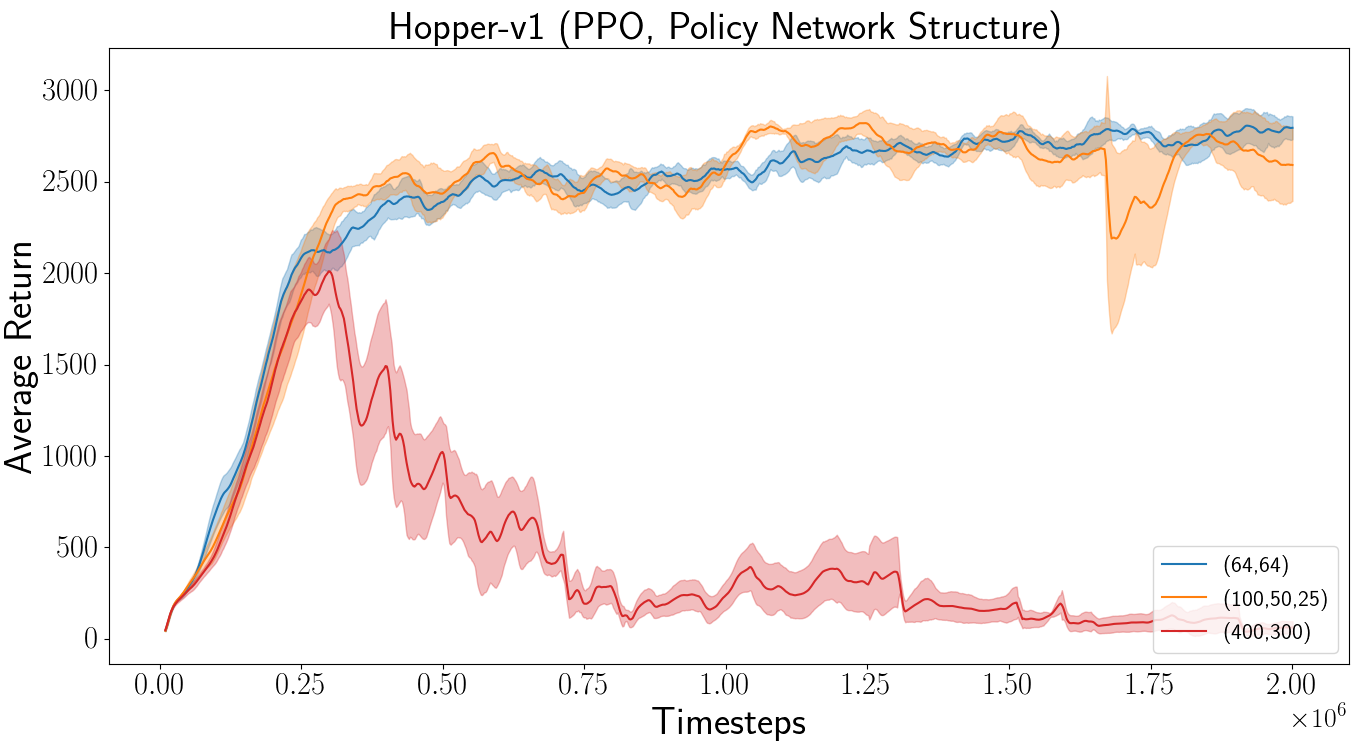}
    \caption{PPO Policy Network structure}
    \label{fig:pponetwork2}
\end{figure}

\begin{figure}[H]
    \includegraphics[width=.49\textwidth]{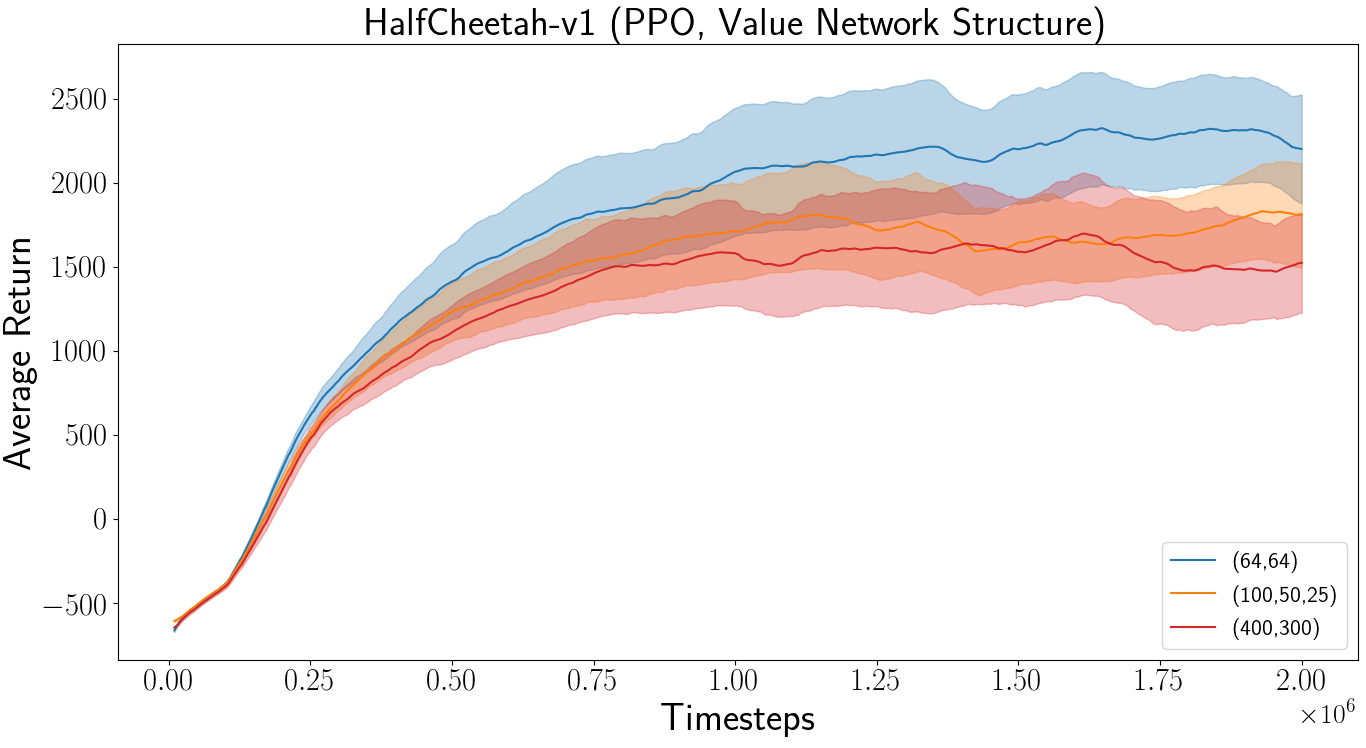}
    \includegraphics[width=.49\textwidth]{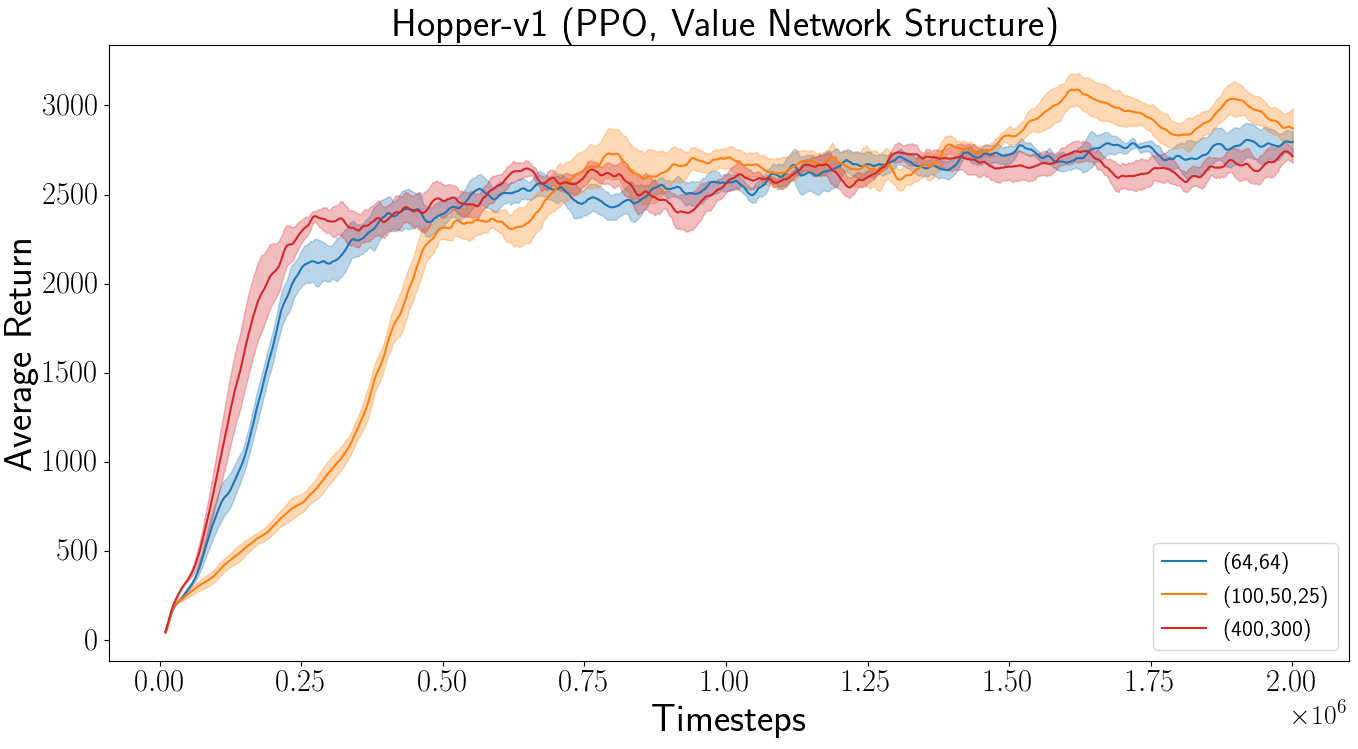}
    \caption{PPO Value Network structure}
    \label{fig:pponetwork}
\end{figure}

\noindent
Experiment results in Figure \ref{fig:pponetworkact}, \ref{fig:pponetwork2}, and \ref{fig:pponetwork} in this section show the effect of the policy network structures and activation functions in the Proximal Policy Optimization (PPO) algorithm.

\subsection{Actor Critic using
Kronecker-Factored Trust Region (ACKTR)}

\begin{figure}[H]
    \centering
    \includegraphics[width=.49\textwidth]{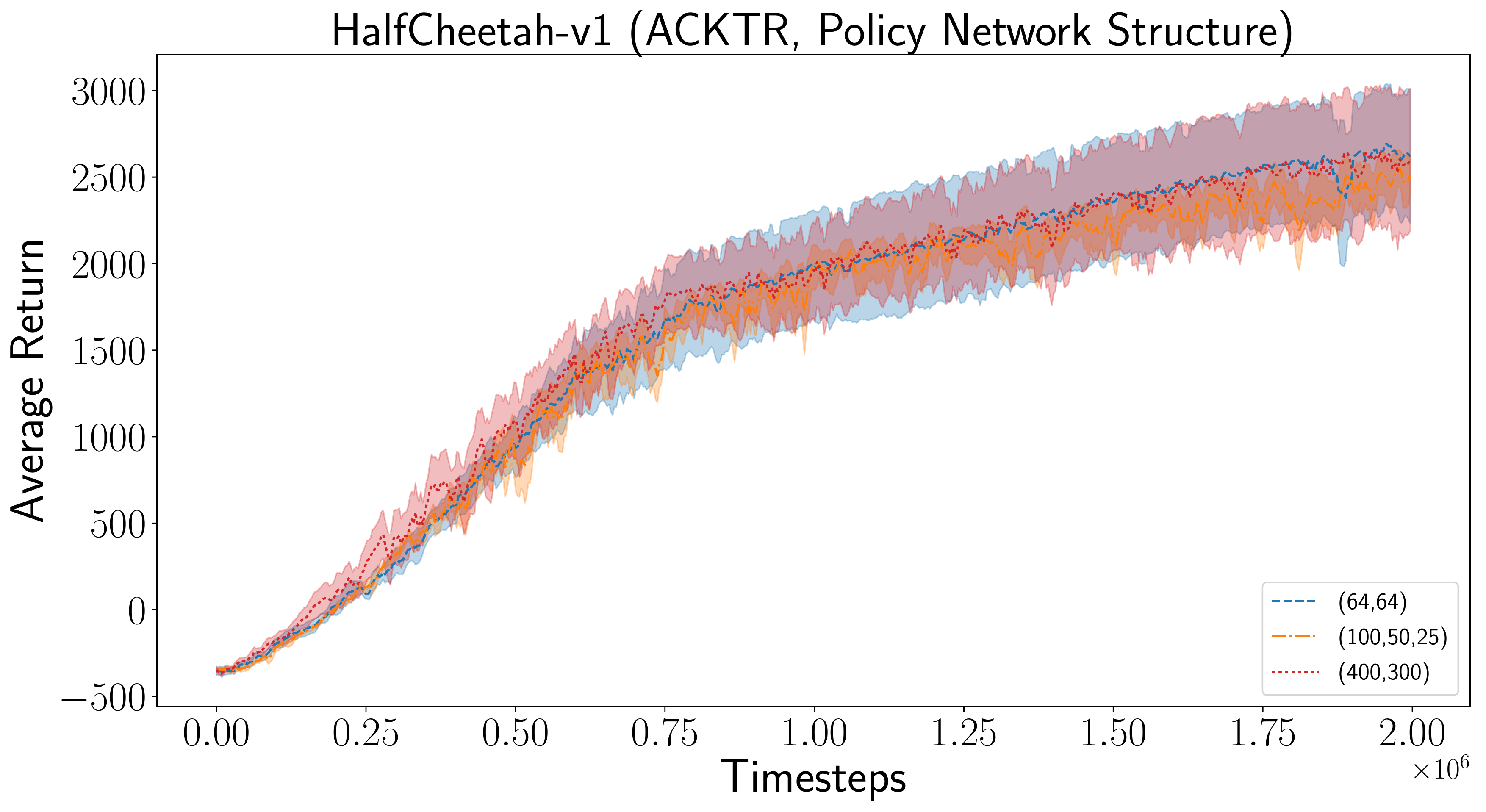}
    \includegraphics[width=.49\textwidth]{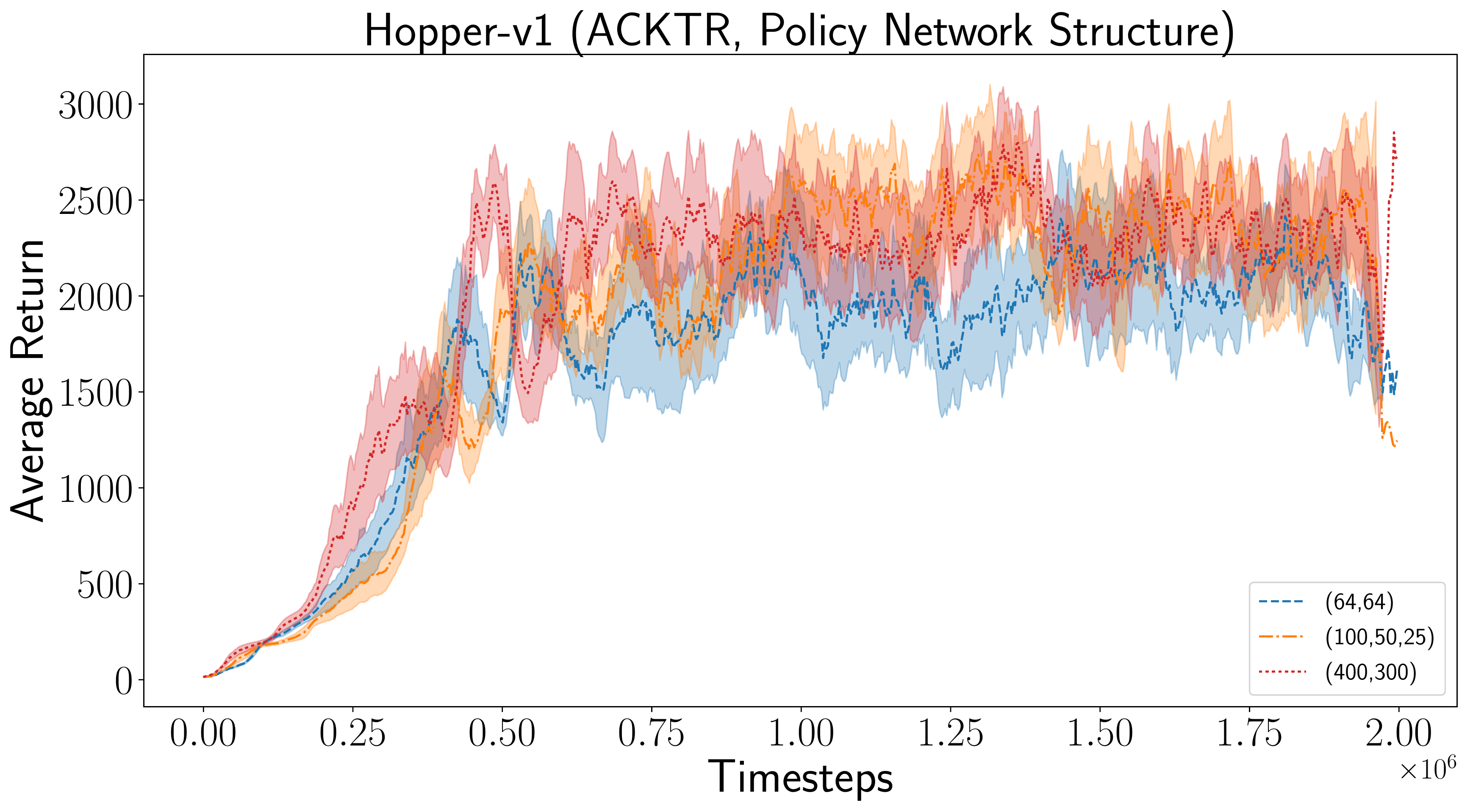}
    \caption{ACKTR Policy Network structure}
    \label{fig:acktrnetwork3}
\end{figure}

\begin{figure}[H]
    \centering
    \includegraphics[width=.49\textwidth]{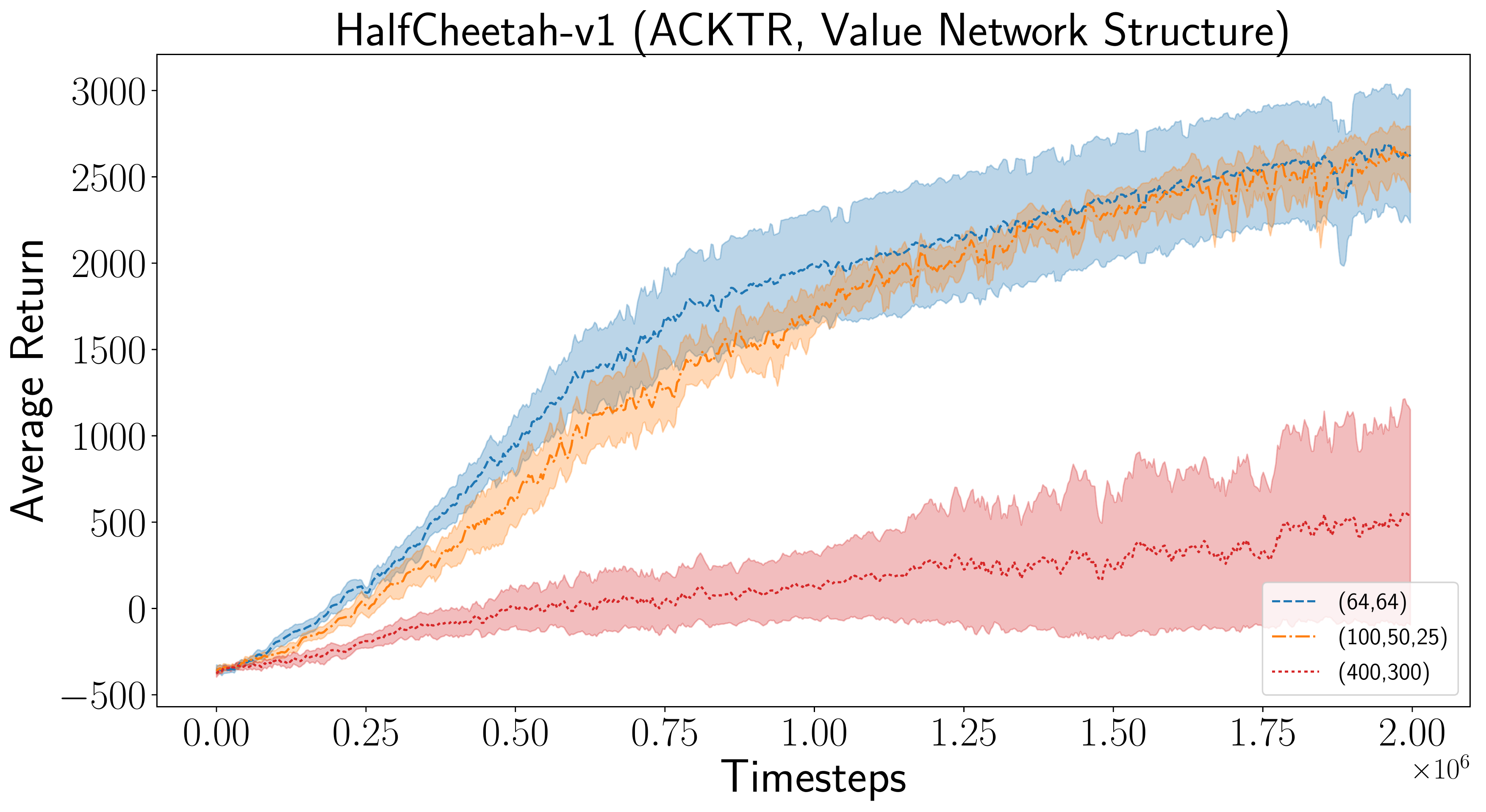}
    \includegraphics[width=.49\textwidth]{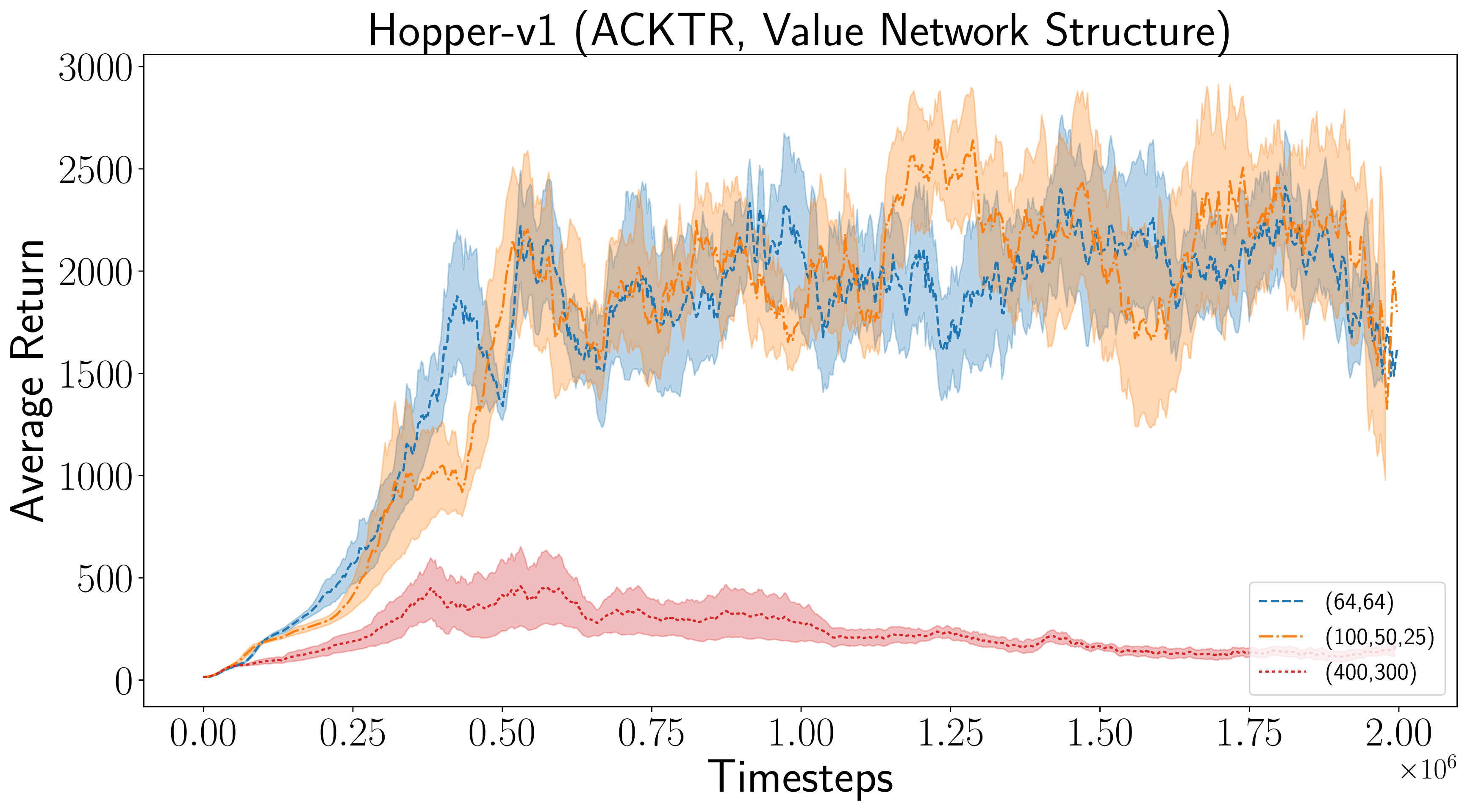}
    \caption{ACKTR Value Network structure}
    \label{fig:acktrnetwork2}
\end{figure}

\begin{figure}[H]
    \centering
    \includegraphics[width=.49\textwidth]{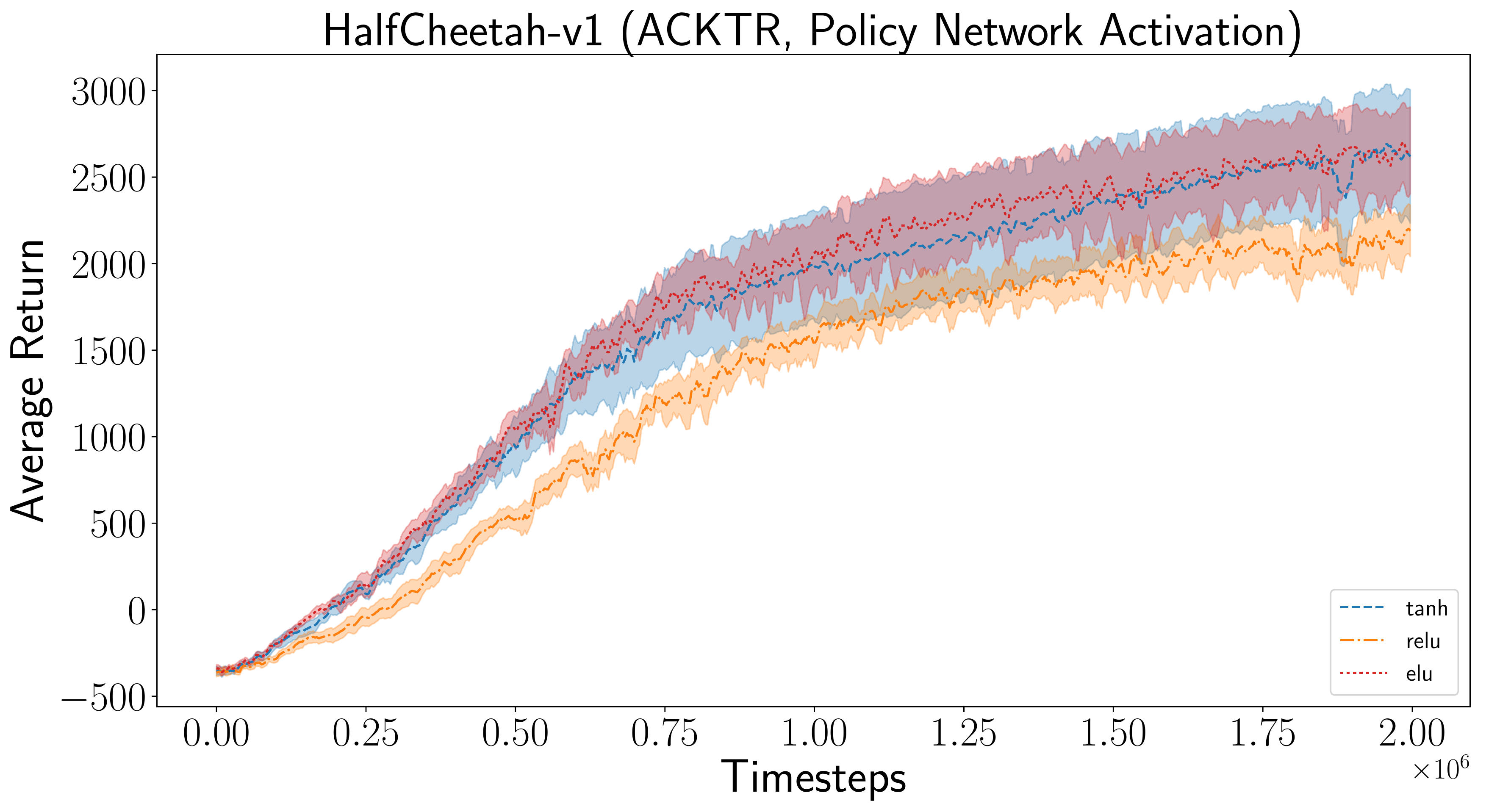}
    \includegraphics[width=.49\textwidth]{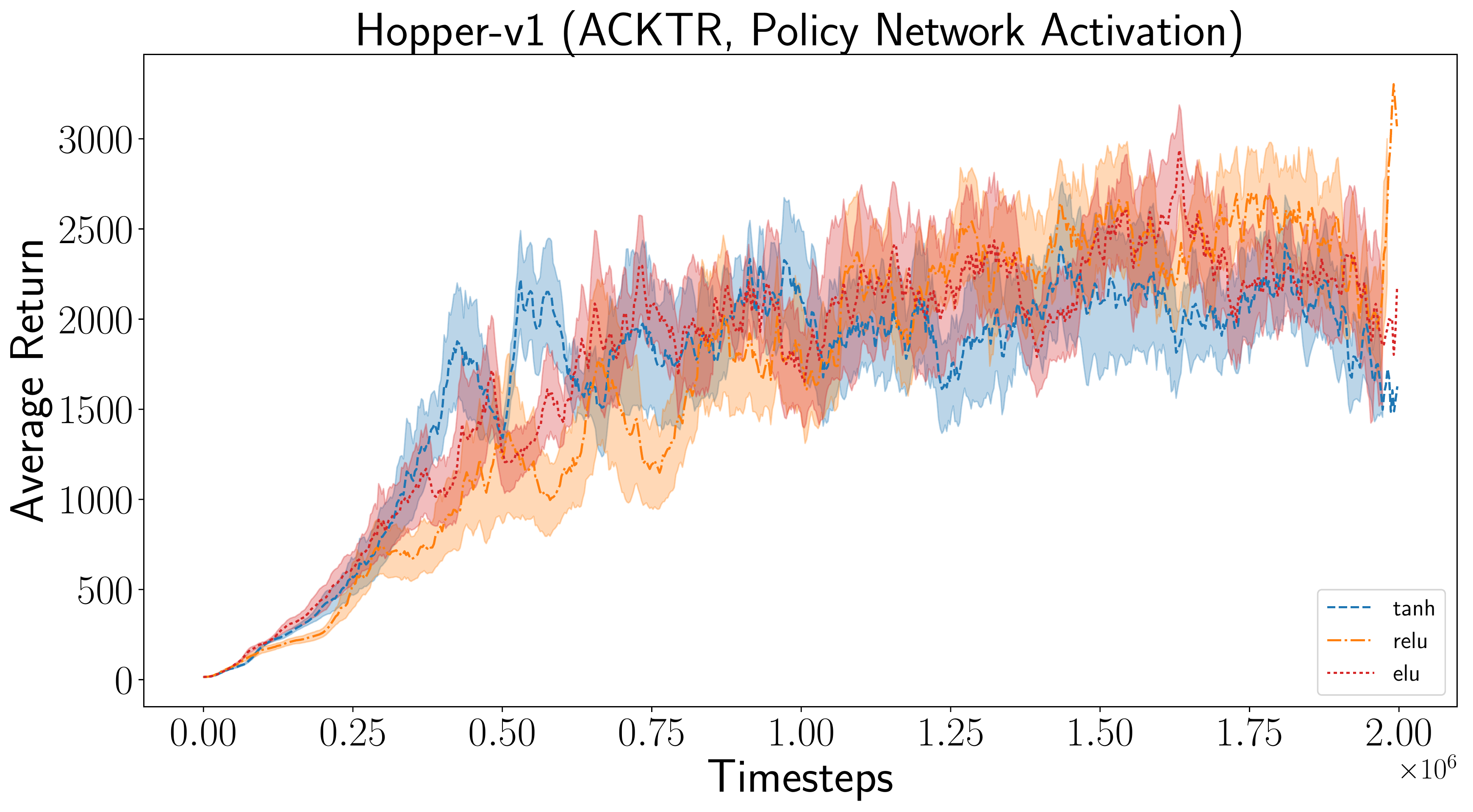}
    \caption{ACKTR Policy Network Activation}
    \label{fig:acktrnetwork4}
  \end{figure}

\begin{figure}[H]
    \centering
    \includegraphics[width=.49\textwidth]{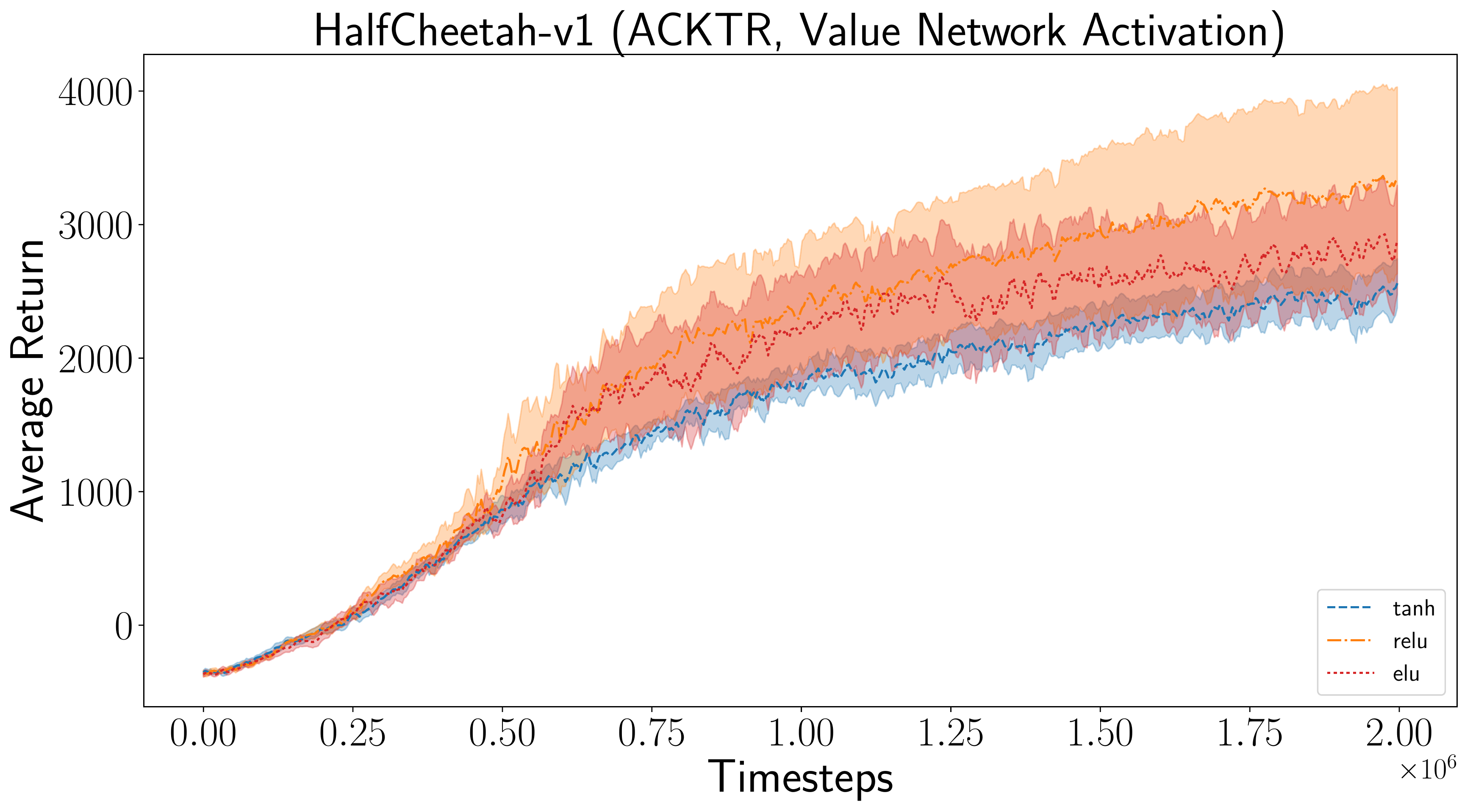}
    \includegraphics[width=.49\textwidth]{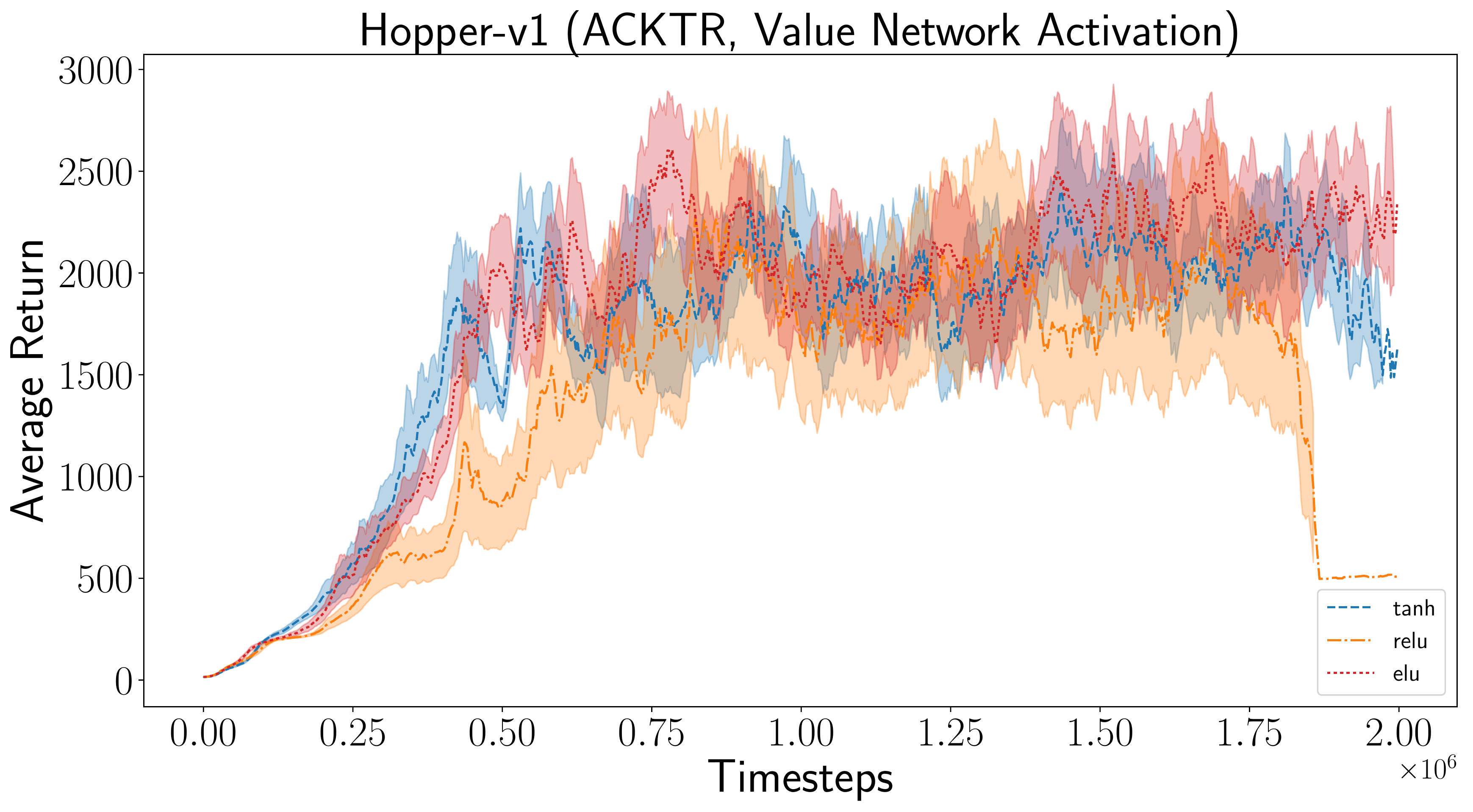}
    \caption{ACKTR Value Network Activation}
    \label{fig:acktrnetworkact}
\end{figure}

We then similarly, show the significance of these hyperparameters in the ACKTR algorithm. Our results show that the value network structure can have a significant effect on the performance of ACKTR algorithm.

\subsection{Trust Region Policy Optimization (TRPO)}
\noindent
\begin{figure}[H]
    \centering
    \includegraphics[width=.49\textwidth]{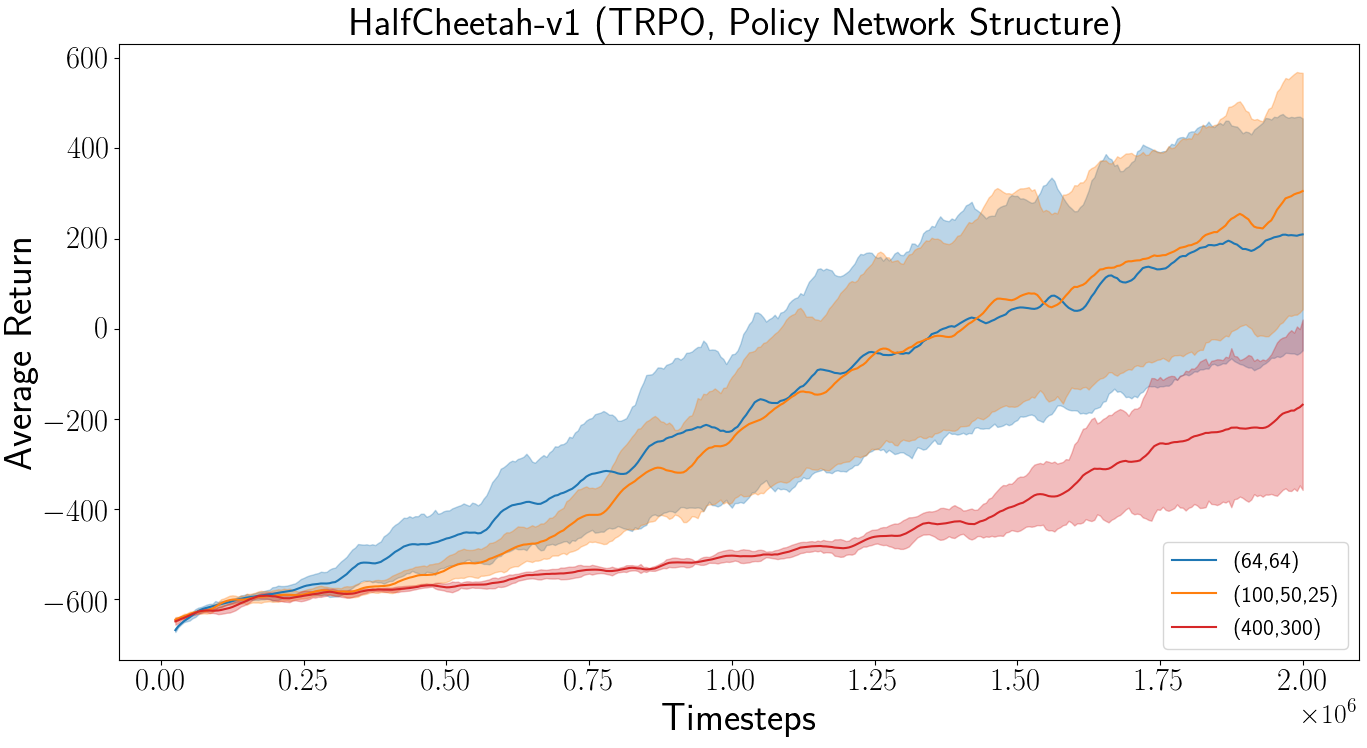}
    \includegraphics[width=.49\textwidth]{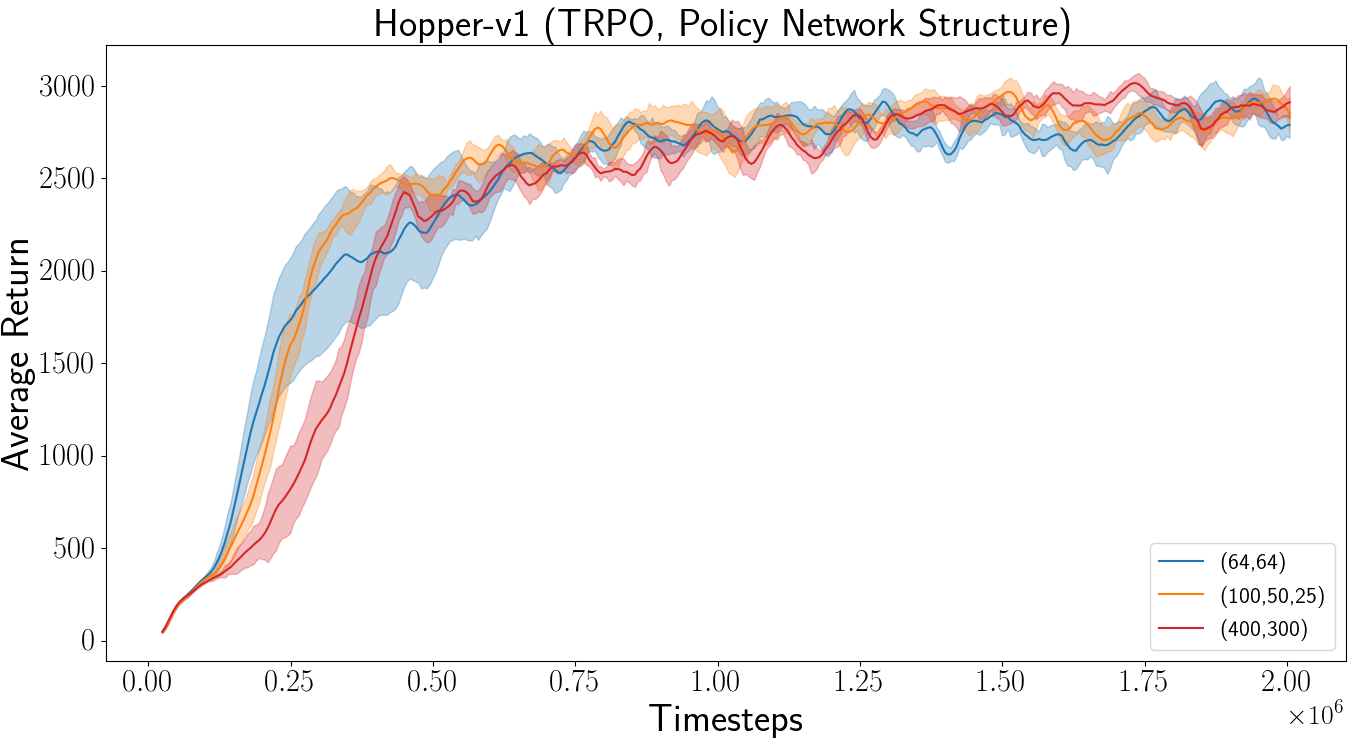}
    \caption{TRPO Policy Network structure}
    \label{fig:trponetwork2}
  \end{figure}

\begin{figure}[H]
    \centering
    \includegraphics[width=.49\textwidth]{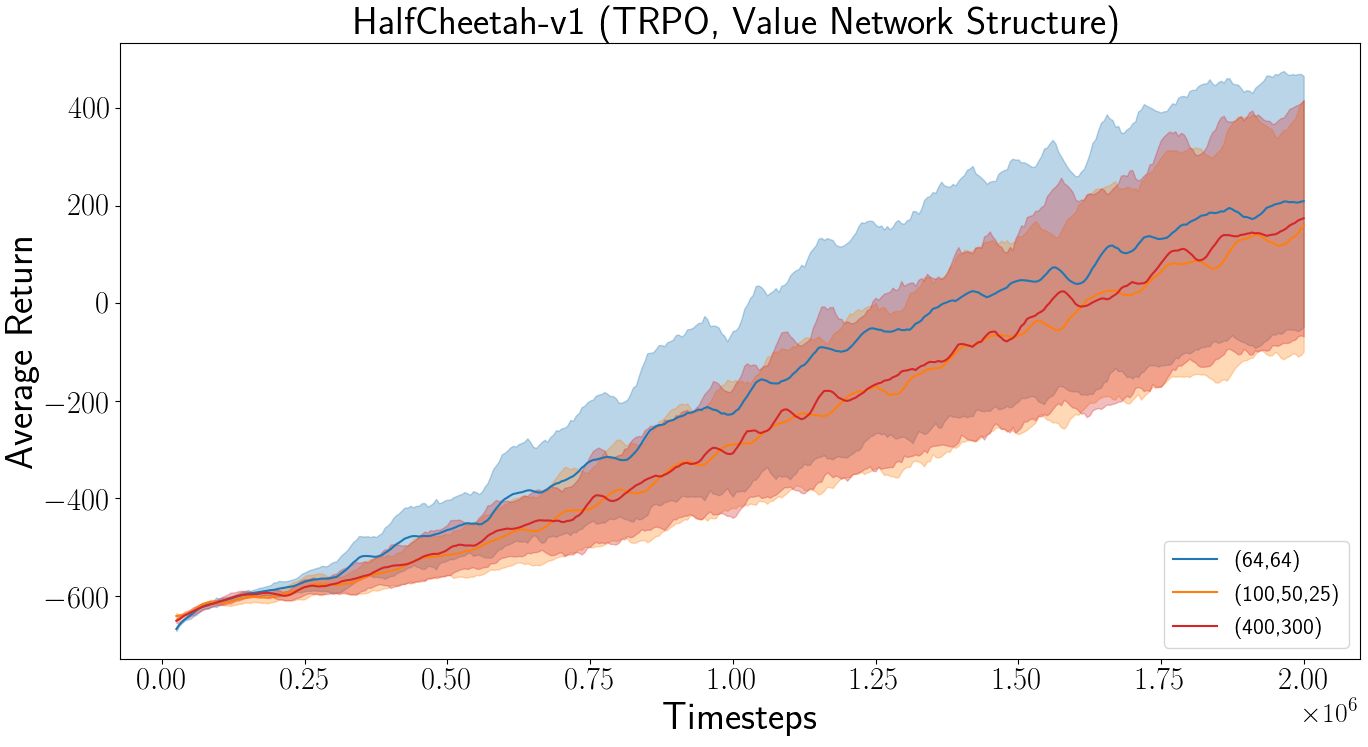}
    \includegraphics[width=.49\textwidth]{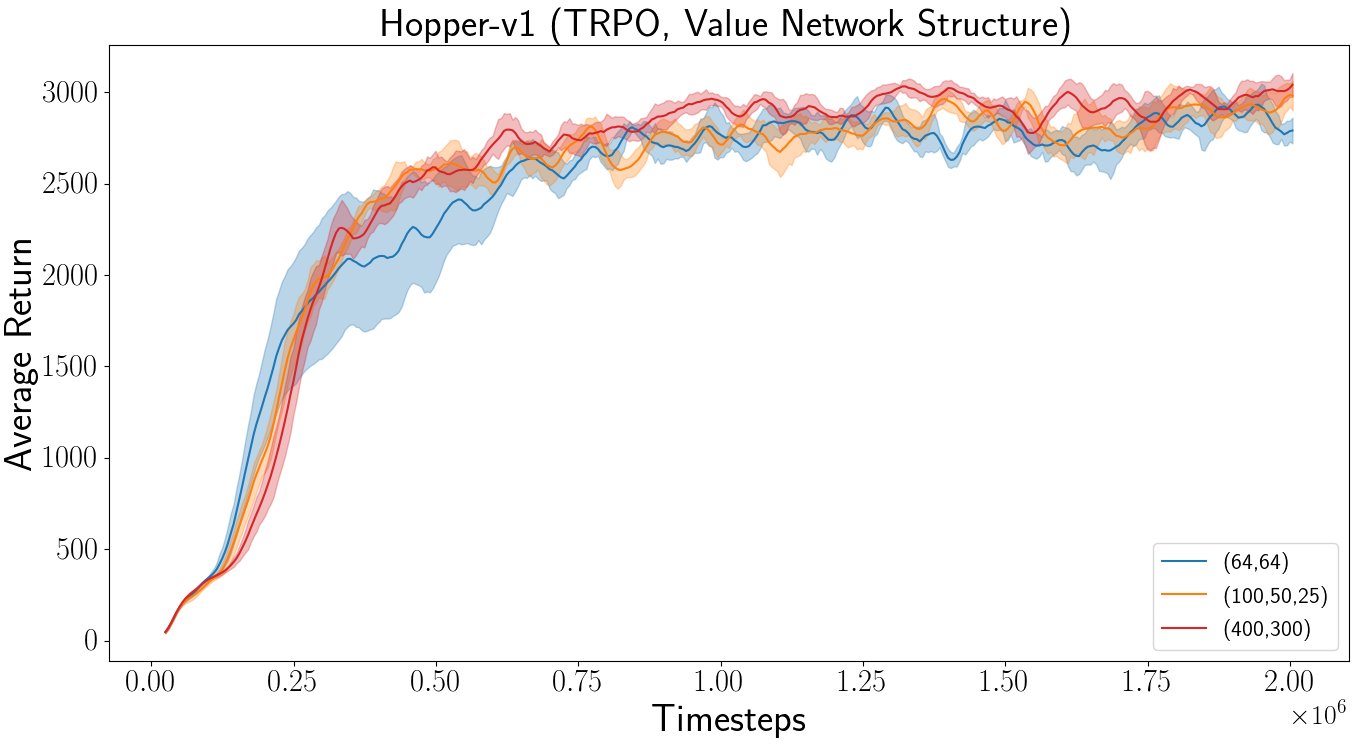}
    \caption{TRPO Value Network structure}
    \label{fig:trponetwork}
\end{figure}

\begin{figure}[H]
    \centering
    \includegraphics[width=.49\textwidth]{"images/HalfCheetah-v1__TRPO,_Policy_Network_Activation_"}
    \includegraphics[width=.49\textwidth]{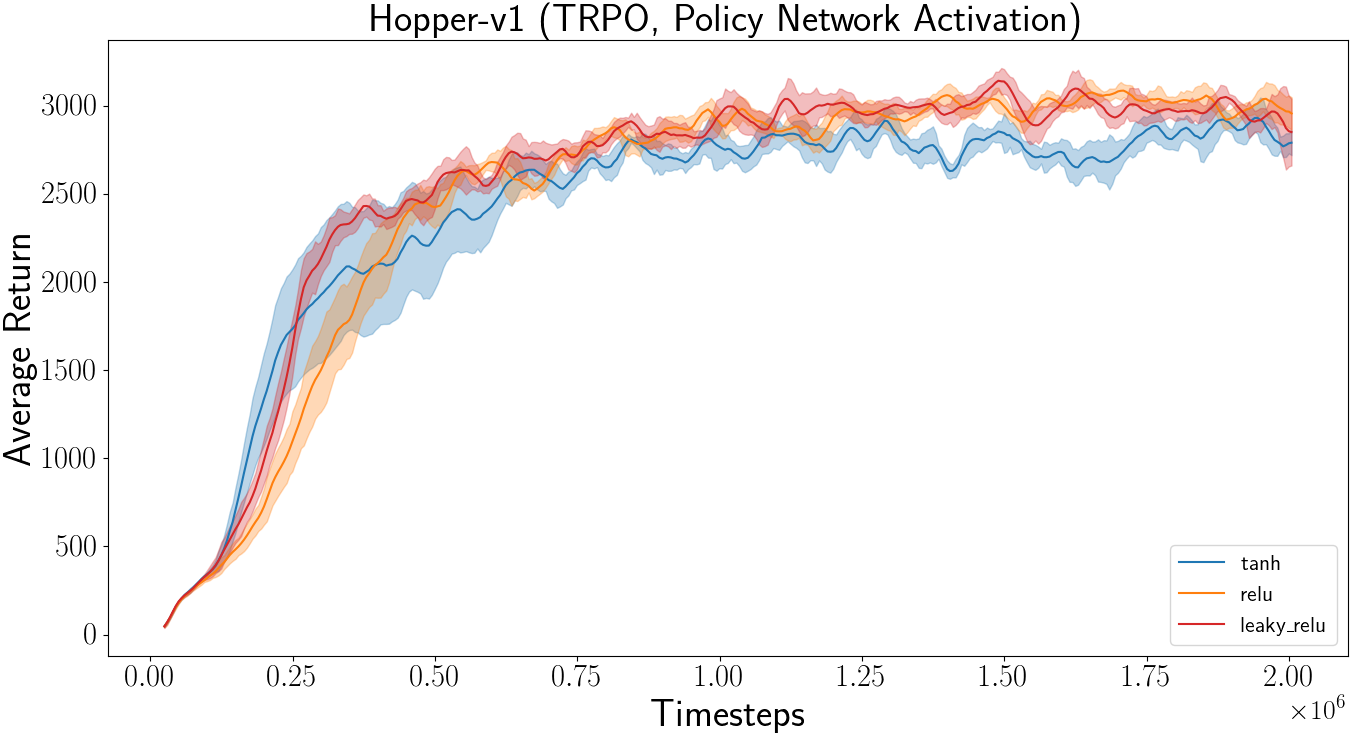}
    \caption{TRPO Policy and Value Network activation}
    \label{fig:trponetwork3}
  \end{figure}

\begin{figure}[H]
  \centering
    \includegraphics[width=.49\textwidth]{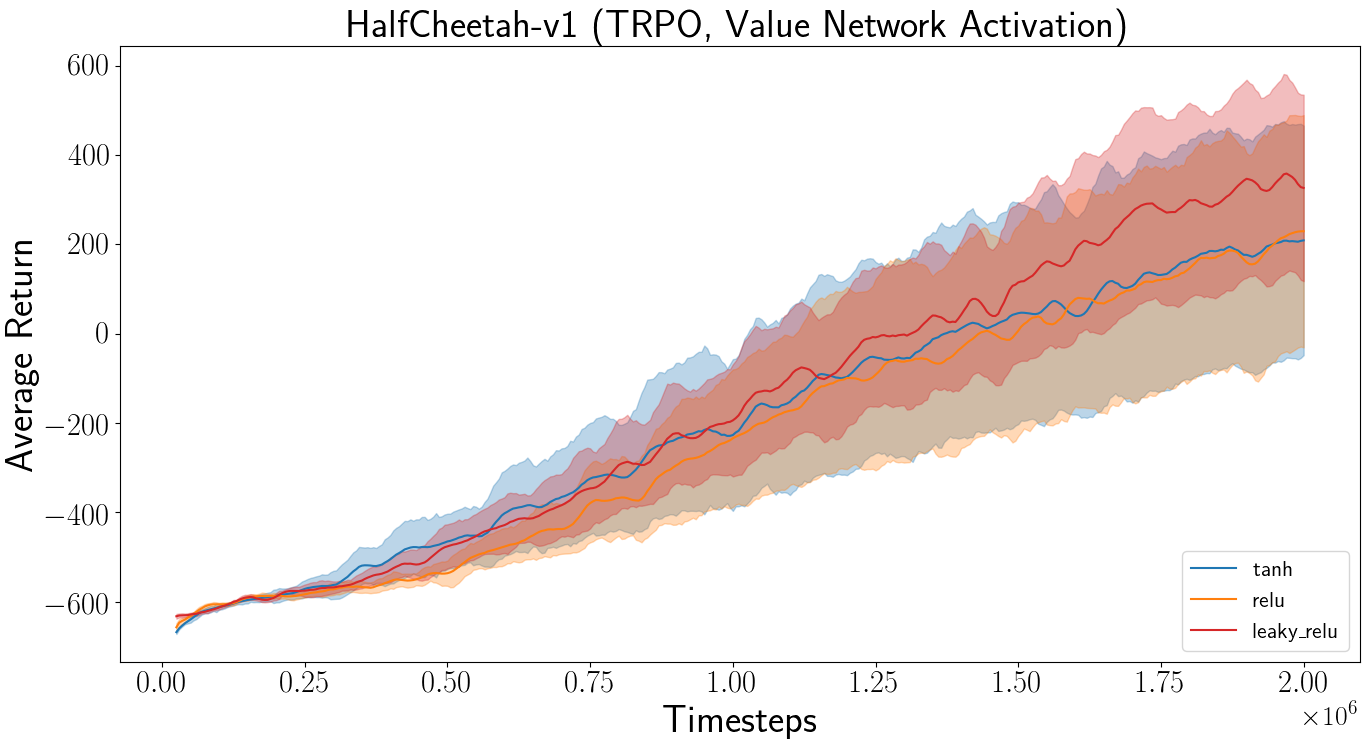}
    \includegraphics[width=.49\textwidth]{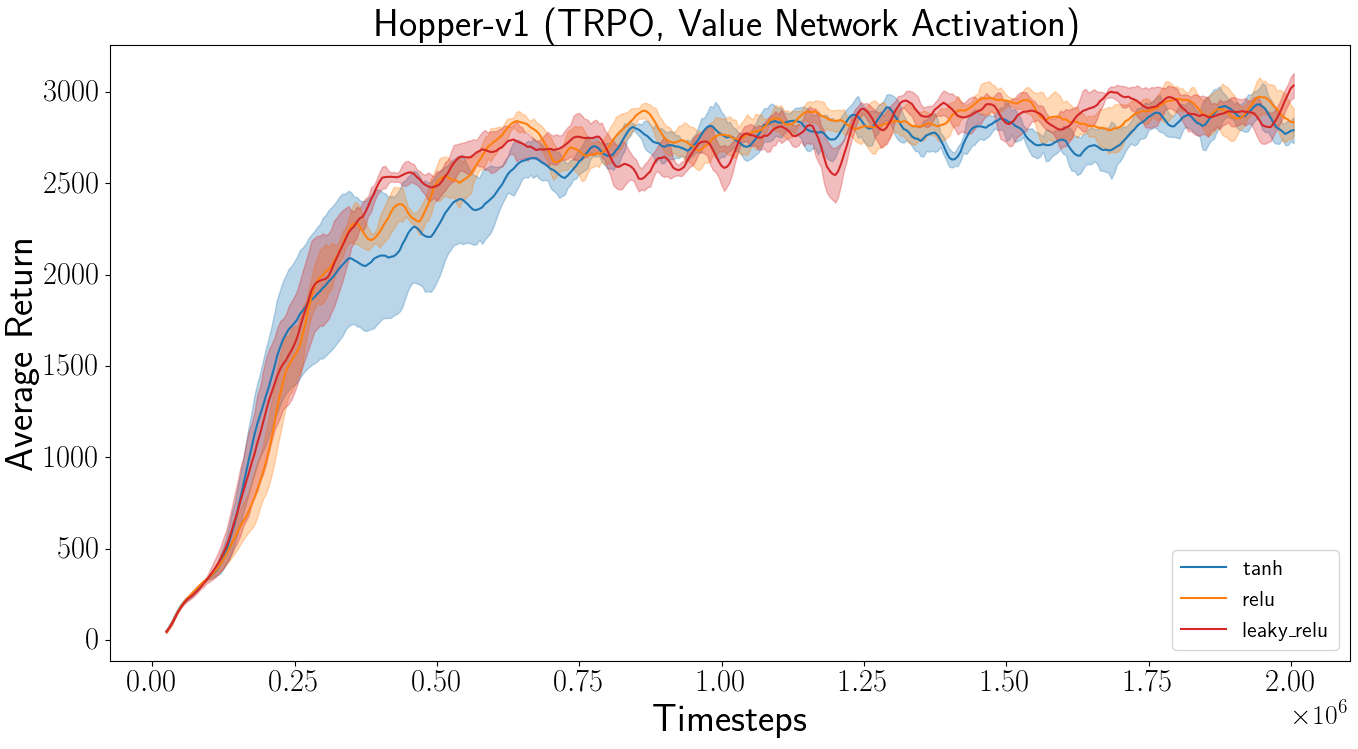}
    \caption{TRPO Policy and Value Network activation}
    \label{fig:trponetwork4}
\end{figure}

In Figures~\ref{fig:trponetwork2},~\ref{fig:trponetwork},~\ref{fig:trponetwork3}, and~\ref{fig:trponetwork4} we show the effects of network structure on the OpenAI baselines implementation of TRPO. In this case, only the policy architecture seems to have a large effect on the performance of the algorithm's ability to learn.

\subsection{Deep Deterministic Policy Gradient (DDPG)}

\begin{figure}[H]
    \centering
    \includegraphics[width=.49\textwidth]{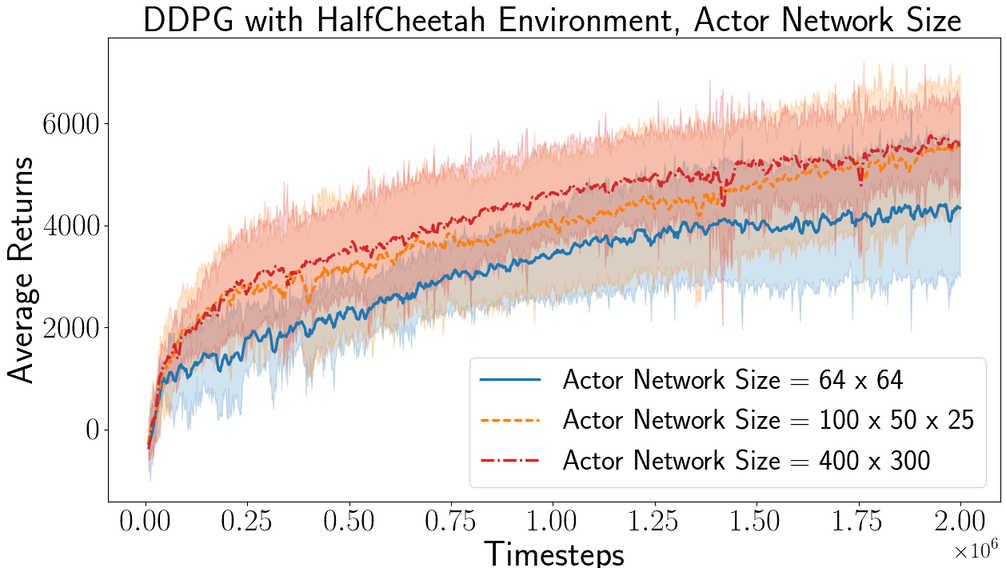}
        \includegraphics[width=.49\textwidth]{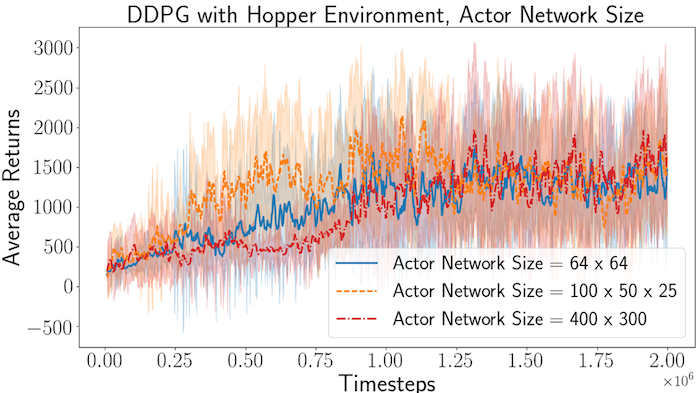}
    \includegraphics[width=.49\textwidth]{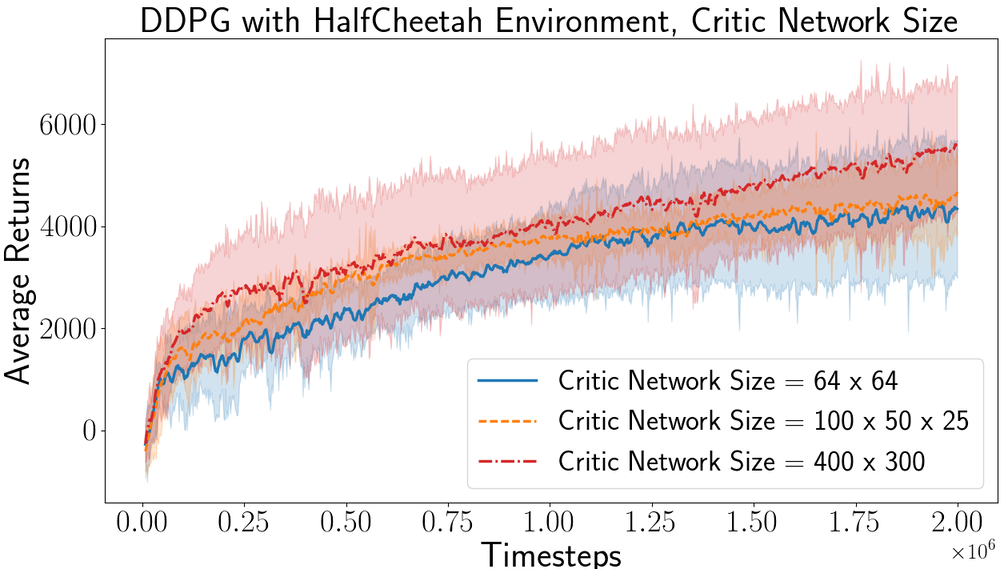}
        \includegraphics[width=.49\textwidth]{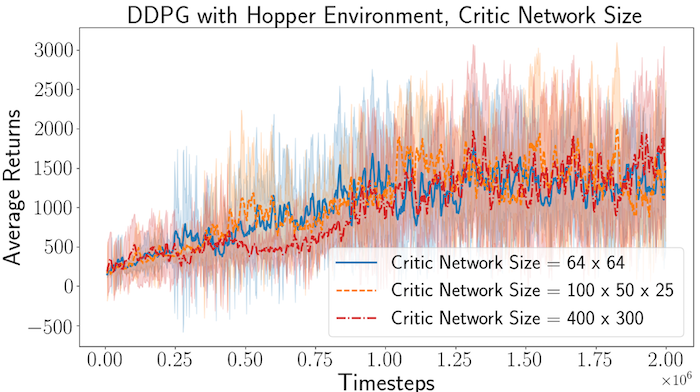}

    \caption{Policy or Actor Network Architecture experiments for DDPG on HalfCheetah and Hopper Environment}
    \label{fig:ddpgnetwork}
\end{figure}

We further analyze the actor and critic network configurations for use in DDPG. As in default configurations, we first use the ReLU activation function for policy networks, and examine the effect of different activations and network sizes for the critic networks. Similarly, keeping critic network configurations under default setting, we also examine the effect of actor network activation functions and network sizes.

\begin{figure}[H]
    \centering
    \includegraphics[width=.49\textwidth]{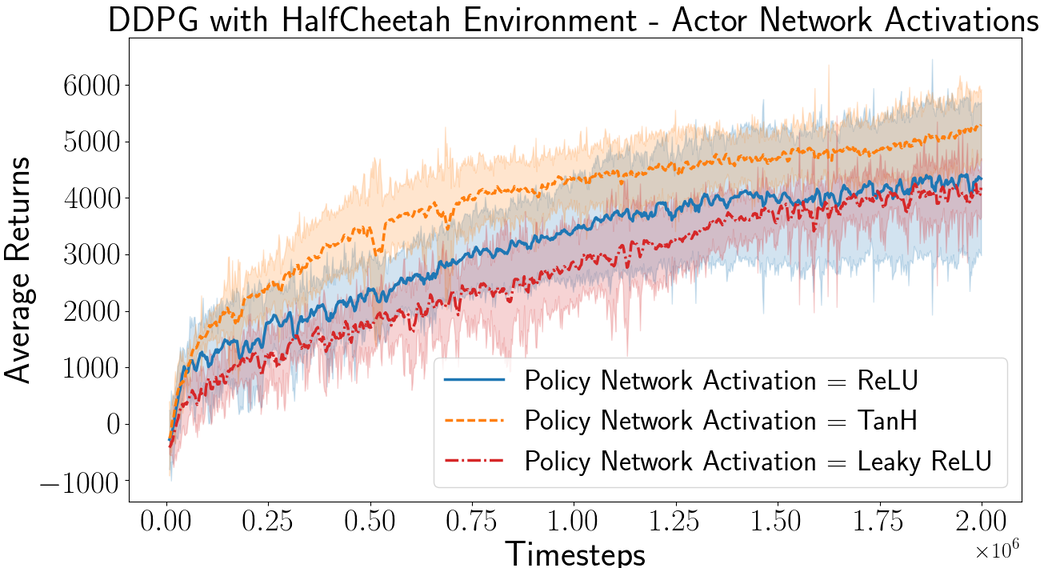}
    \includegraphics[width=.49\textwidth]{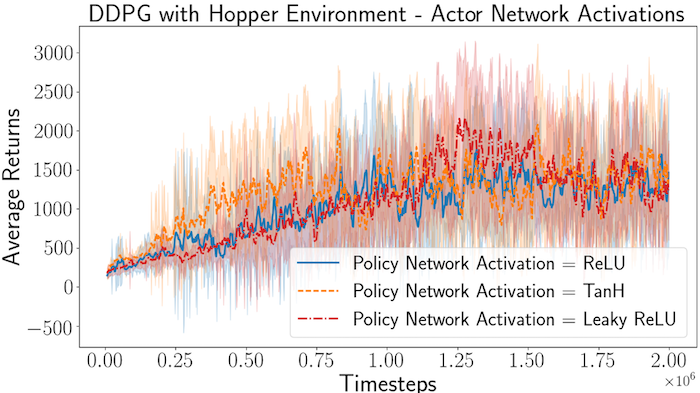}
        \label{fig:ddpgnetworkact}
\end{figure}

\begin{figure}[H]
    \includegraphics[width=.49\textwidth]{images/ddpg_halfcheetah_value_activations.png}
    \includegraphics[width=.49\textwidth]{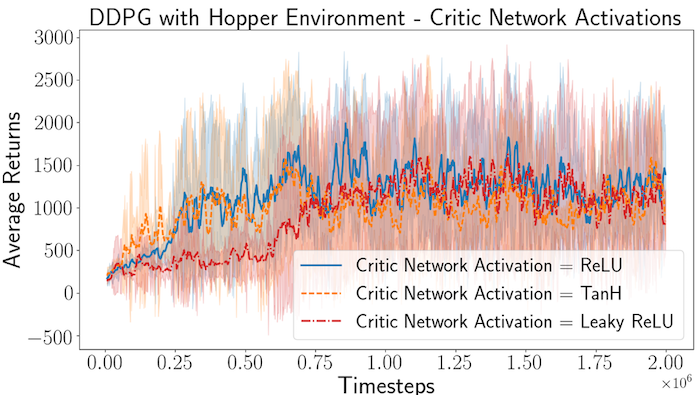}
    \caption{Significance of Value Function or Critic Network Activations for DDPG on HalfCheetah and Hopper Environment}
    \label{fig:ddpgnetworkact2}
\end{figure}

\section{Reward Scaling Parameter in DDPG}

\begin{figure}[H]
    \centering
    \includegraphics[width=.49\textwidth]{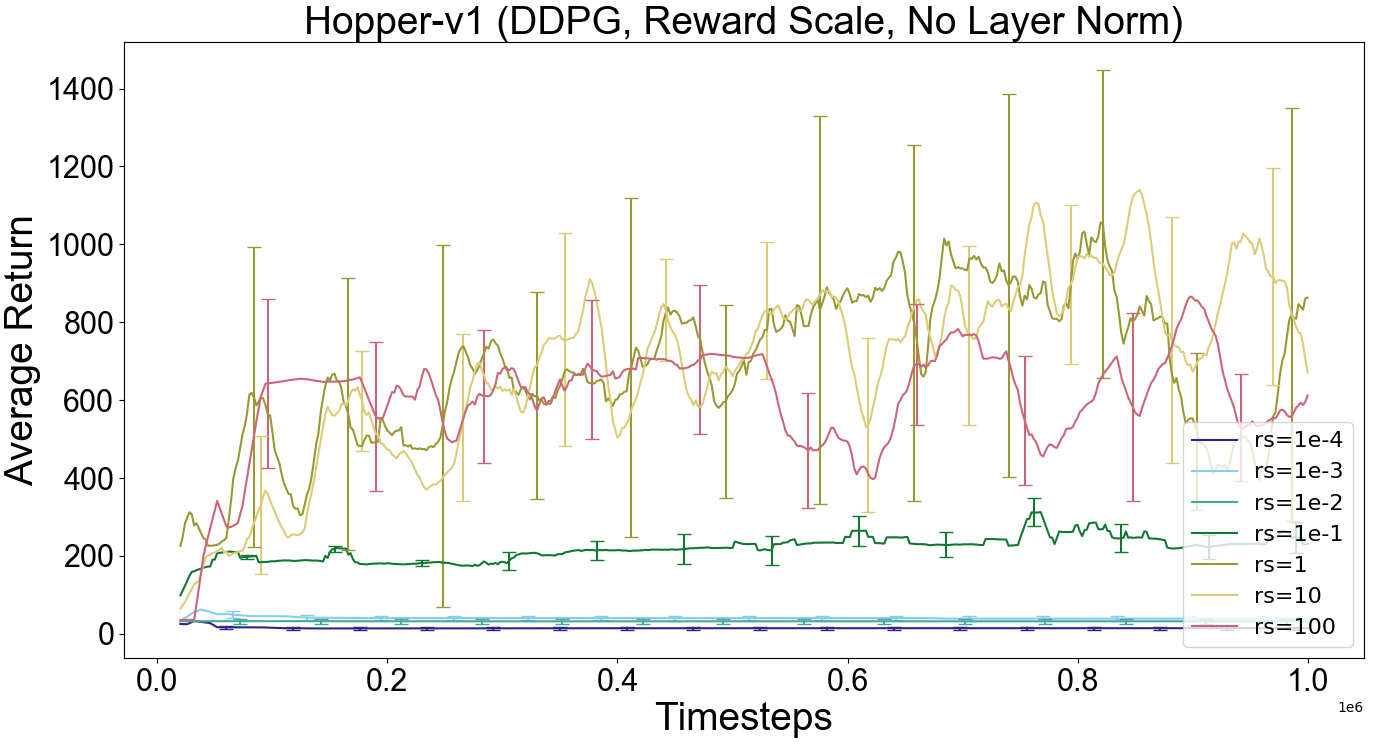}
    \includegraphics[width=.49\textwidth]{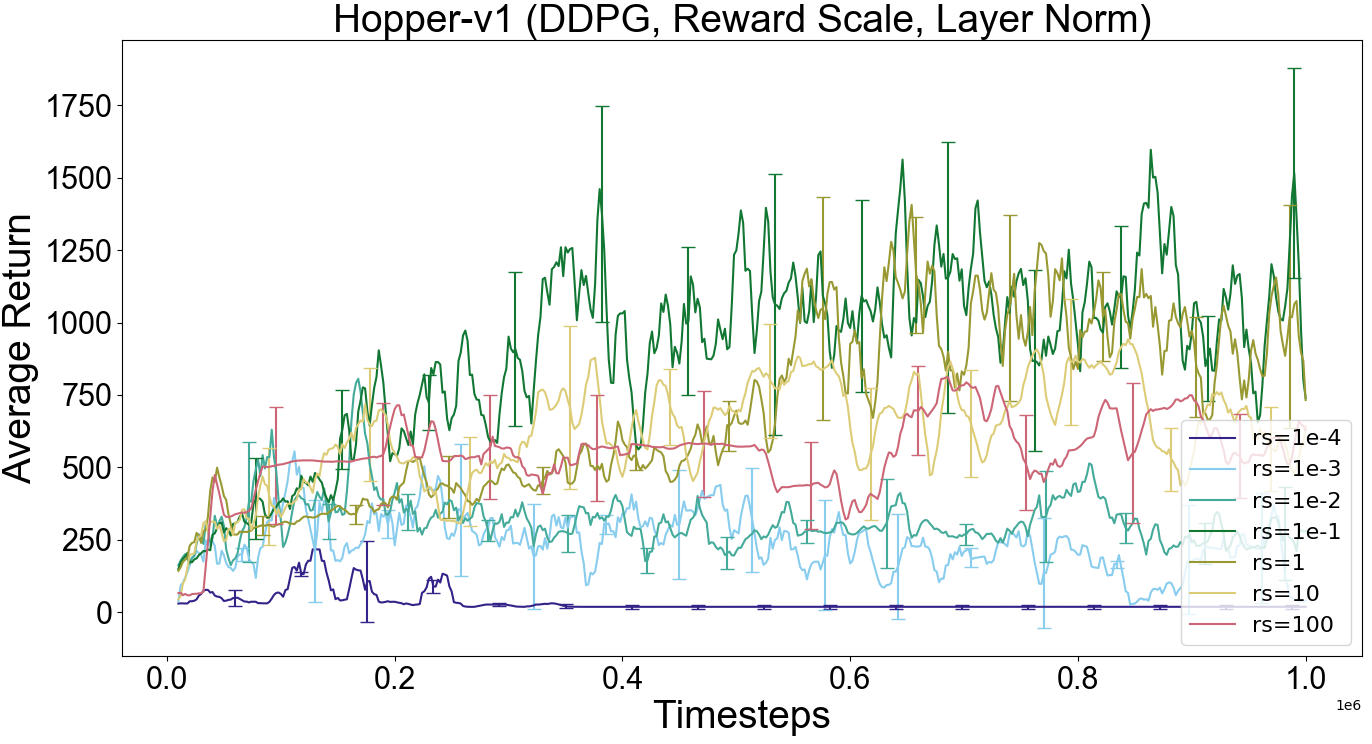}
    \caption{DDPG reward rescaling on Hopper-v1, with and without layer norm.}
    \label{fig:reward_scaling_ddpg2}
\end{figure}

\begin{figure}[H]
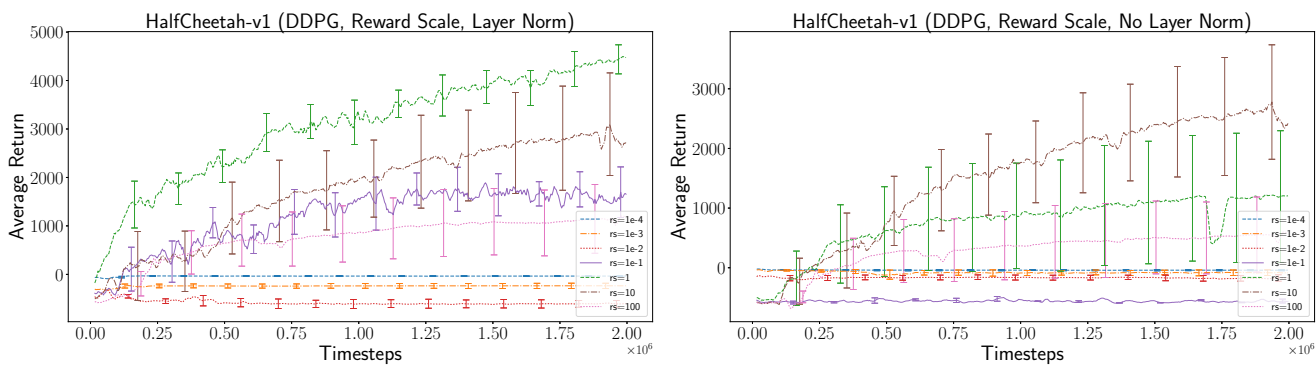

    \includegraphics[width=.49\textwidth]{"images/HalfCheetah-v1__DDPG,_Reward_Scale,_Layer_Norm_"}
    \includegraphics[width=.49\textwidth]{"images/HalfCheetah-v1__DDPG,_Reward_Scale,_No_Layer_Norm_"}
    \caption{DDPG reward rescaling on HalfCheetah-v1, with and without layer norm.}
    \label{fig:reward_scaling_ddpg}
\end{figure}

Several related work ~\cite{QPROP,IPG,rllab} have often reported  that for DDPG the reward scaling parameter often needs to be fine-tuned for stabilizing the performance of DDPG. It can make a significant impact in performance of DDPG based on the choice of environment. We examine several reward scaling parameters and demonstrate the effect this parameter can have on the stability and performance of DDPG, based on the HalfCheetah and Hopper environments. Our experiment results, as demonstrated in Figure~\ref{fig:reward_scaling_ddpg} and~\ref{fig:reward_scaling_ddpg2}, show that the reward scaling parameter indeed can have a significant impact on performance. Our results show that, very small or negligible reward scaling parameter can significantly detriment the performance of DDPG across all environments. Furthermore, a scaling parameter of $10$ or $1$ often performs good. Based on our analysis, we suggest that every time DDPG is reported as a baseline algorithm for comparison, the reward scaling parameter should be fine-tuned, specific to the algorithm.

\section{Batch Size in TRPO}

\begin{figure}[H]
    \centering
    \includegraphics[width=.49\textwidth]{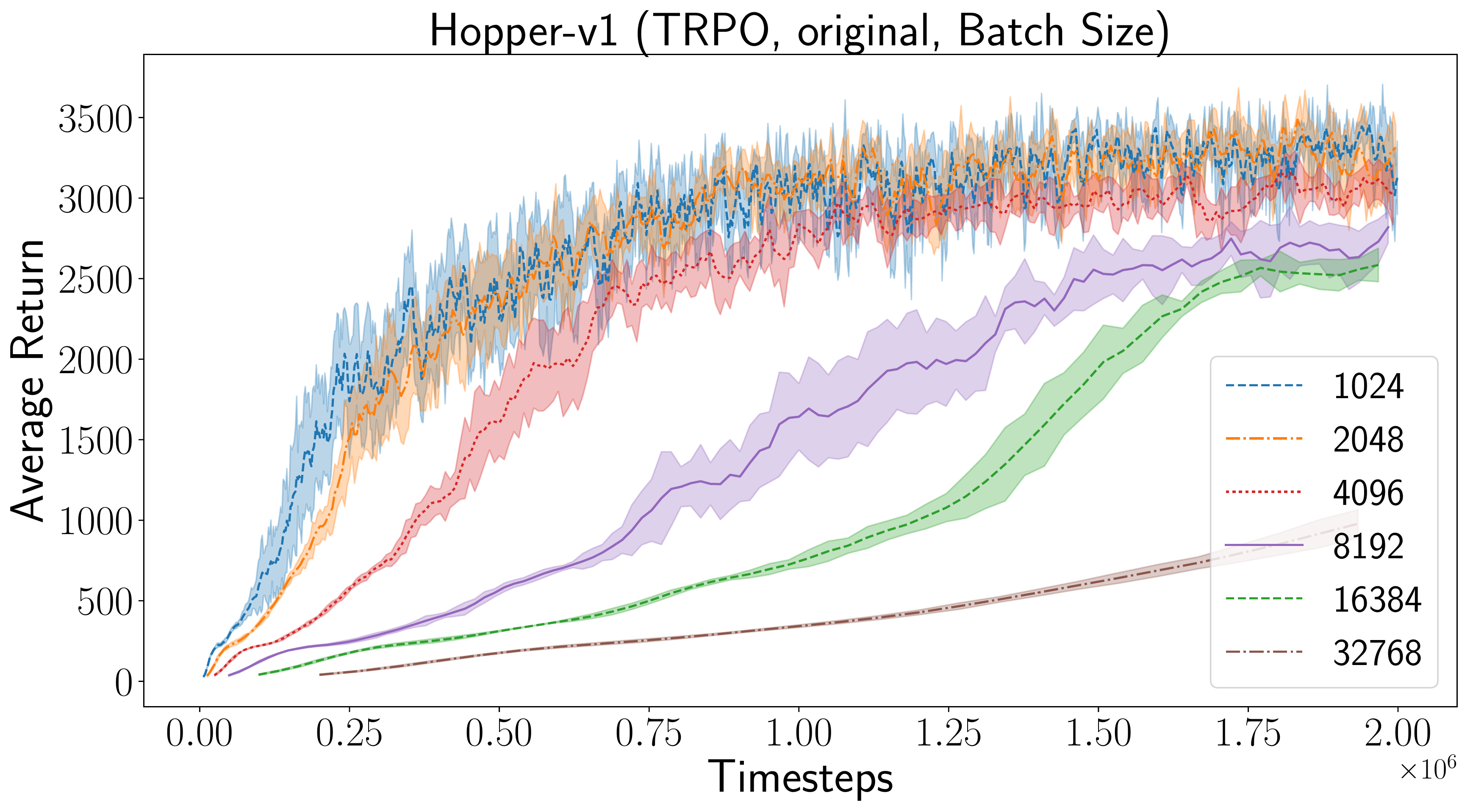}
    \includegraphics[width=.49\textwidth]{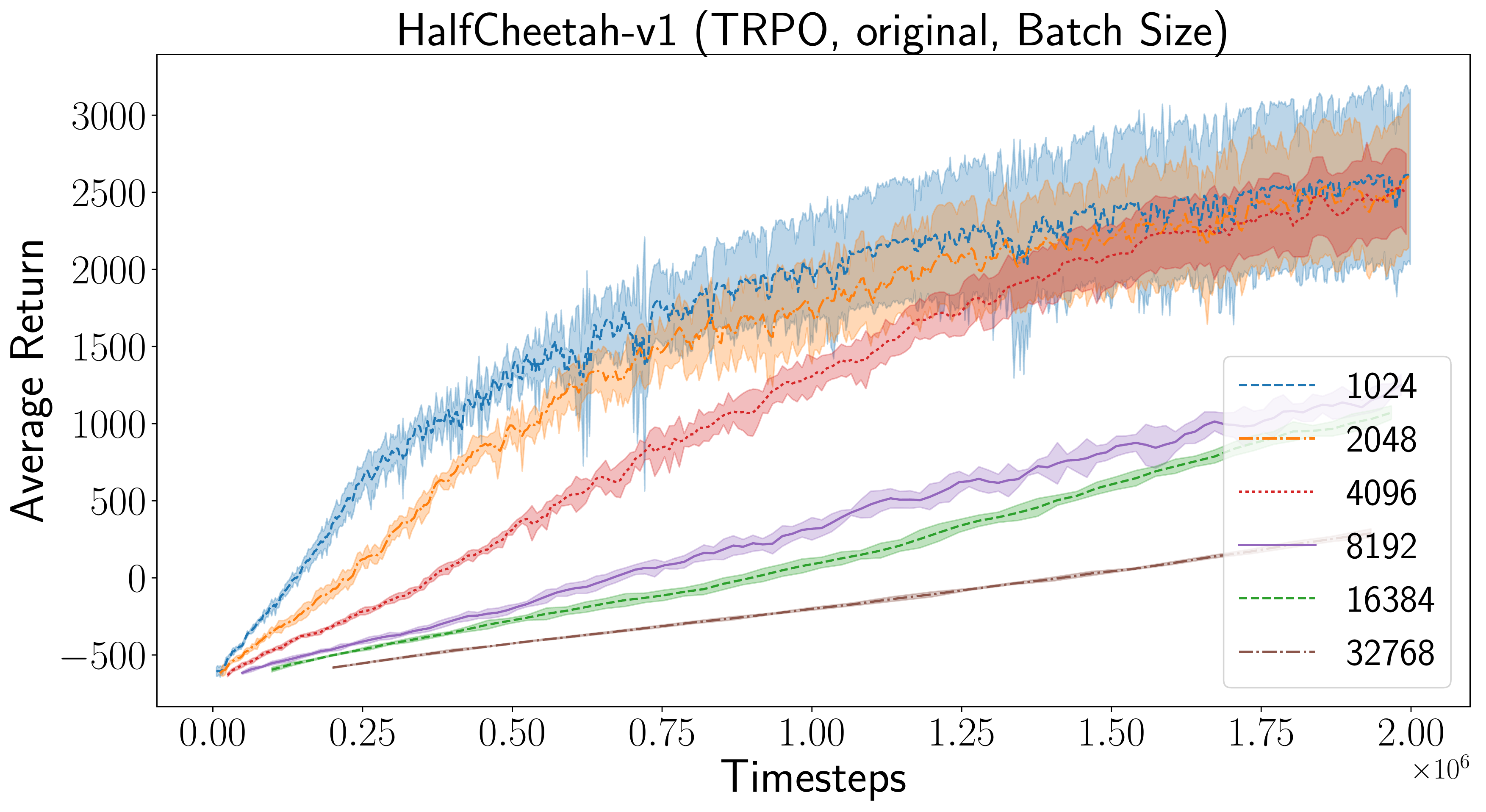}
    \caption{TRPO~\cite{TRPO} original code batch size experiments.}
    \label{fig:batch_size_original}
\end{figure}

\begin{figure}[H]
    \centering
    \includegraphics[width=.49\textwidth]{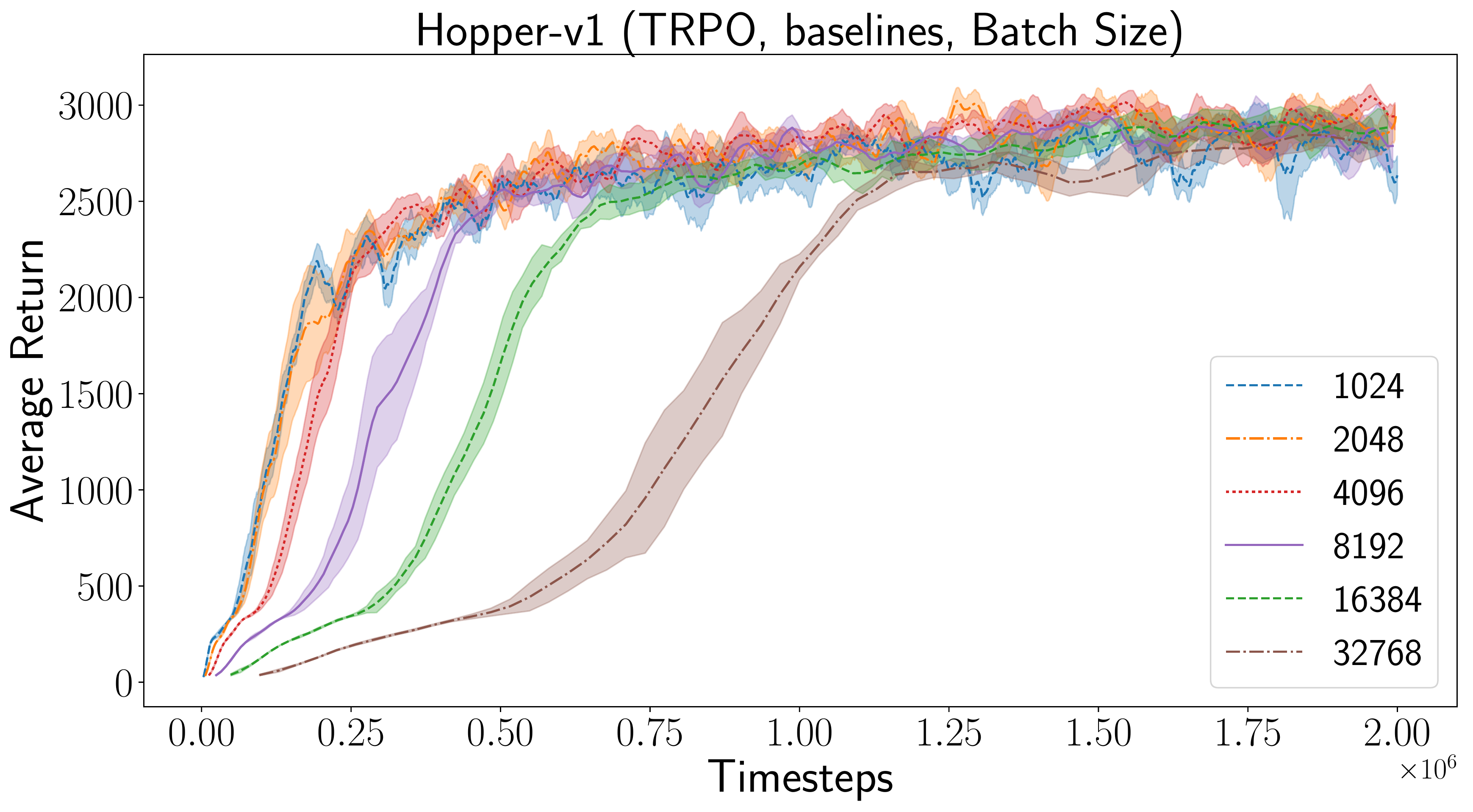}
    \includegraphics[width=.49\textwidth]{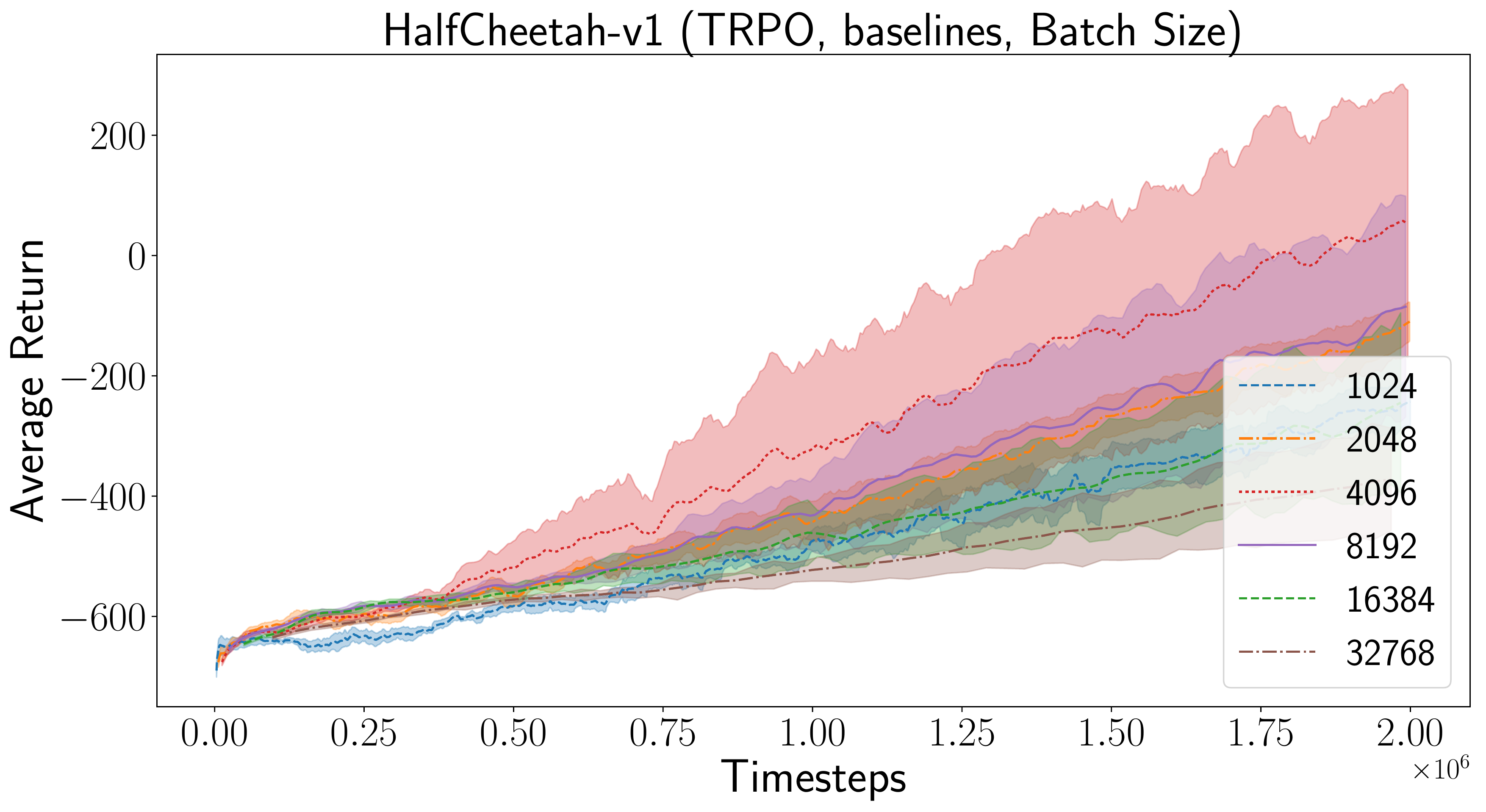}
    \includegraphics[width=.49\textwidth]{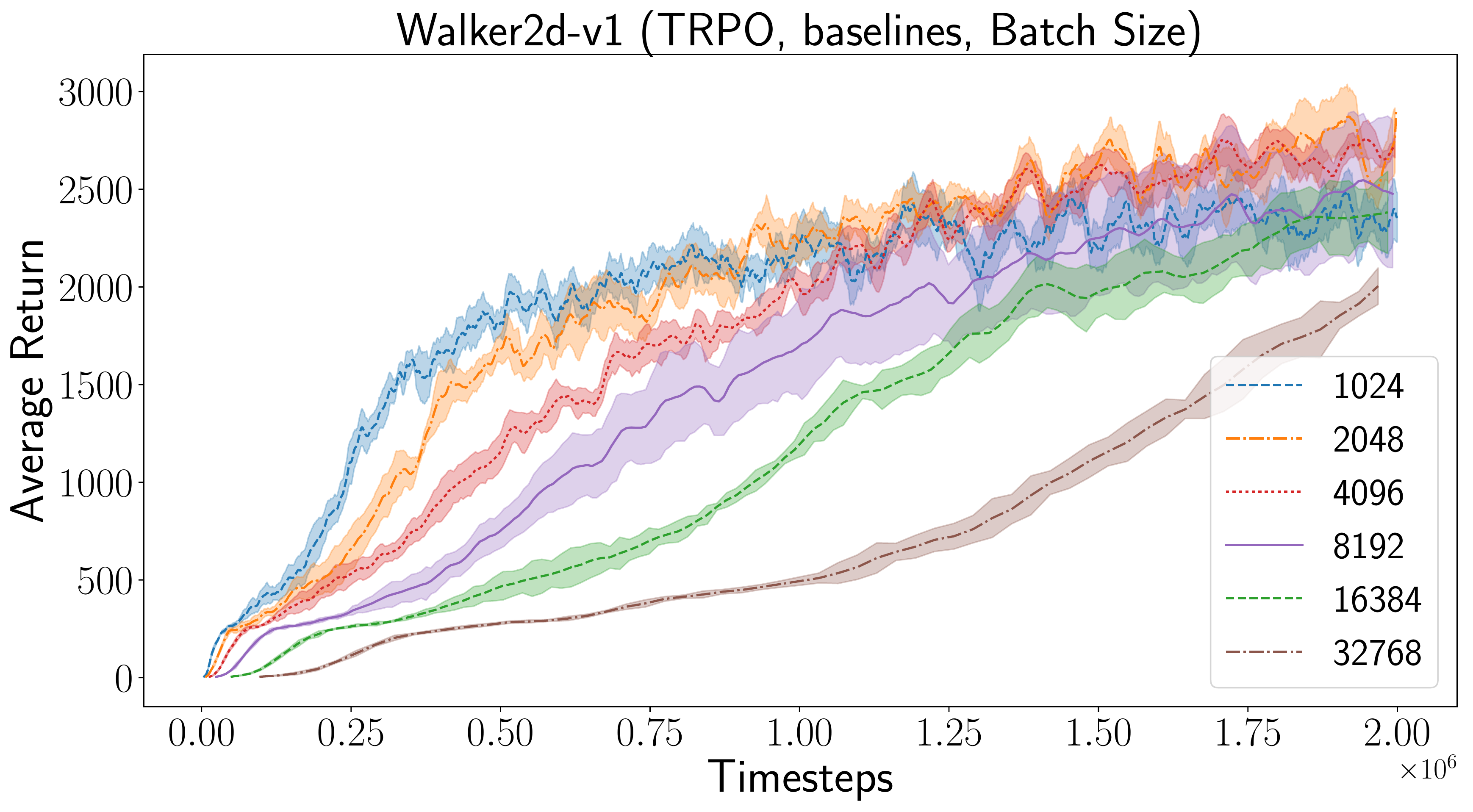}
    \includegraphics[width=.49\textwidth]{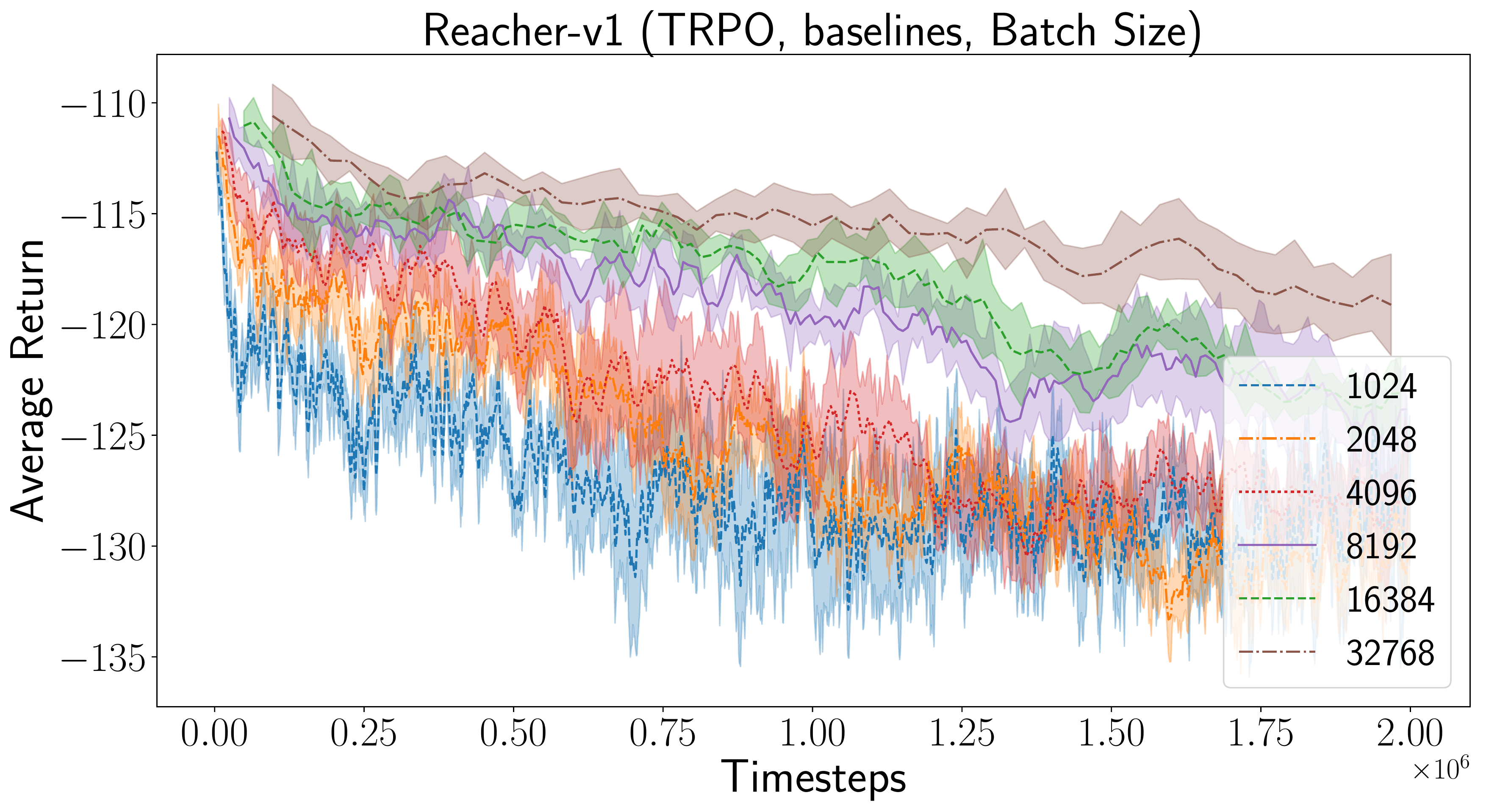}
    \caption{TRPO~\cite{PPO} baselines code batch size experiments.}
    \label{fig:batch_size_baselines}
\end{figure}

We run batch size experiments using the original TRPO code~\cite{TRPO} and the OpenAI baselines code~\cite{PPO}. These results can be found in Experiment results in Figure~\ref{fig:batch_size_original} and Figure~\ref{fig:batch_size_baselines}, show that for both HalfCheetah-v1 and Hopper-v1 environments, a batch size of $1024$ for TRPO performs best, while perform degrades consecutively as the batch size is increased.

\section{Random Seeds}

\begin{figure}[H]
    \centering
    \includegraphics[width=0.45\textwidth]{"images/HalfCheetah-v1__TRPO,_Different_Random_Seeds_"}
    \includegraphics[width=0.45\textwidth]{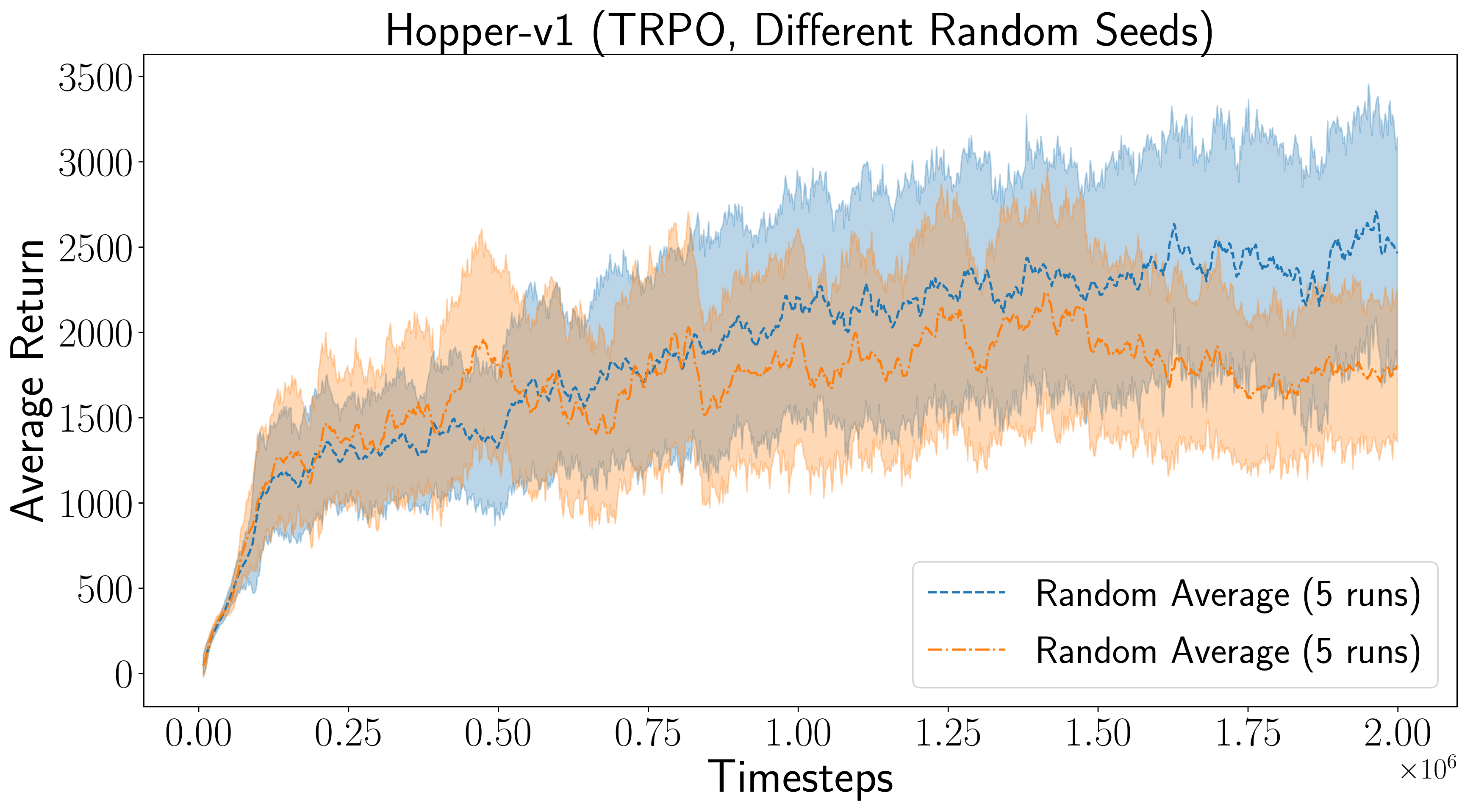}
    \caption{Two different TRPO experiment runs, with same hyperparameter configurations, averaged over two splits of 5 different random seeds. }
    \label{fig:randomseeds_trposup}
\end{figure}

\begin{figure}[H]
    \centering
    \includegraphics[width=0.45\textwidth]{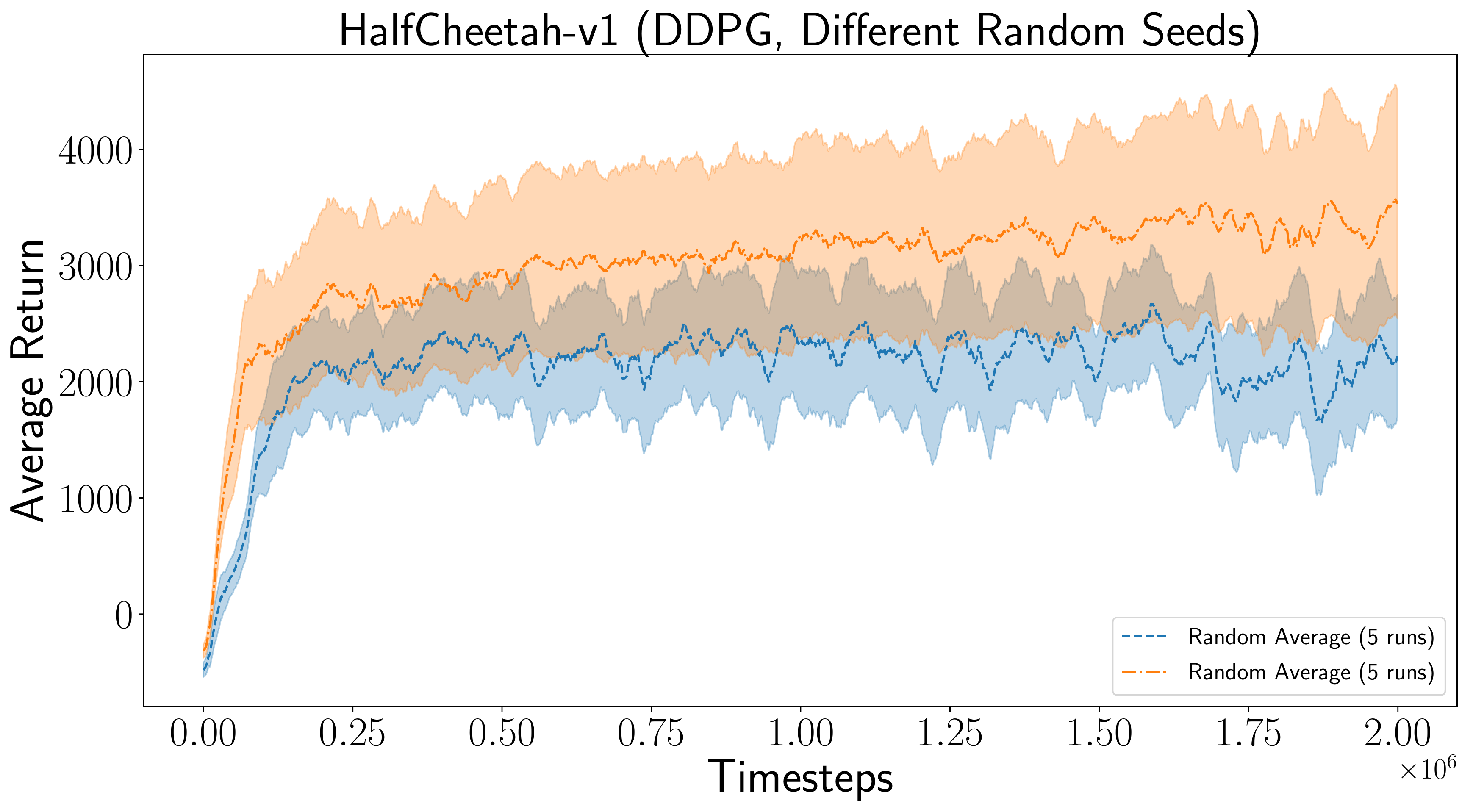}
    \includegraphics[width=0.45\textwidth]{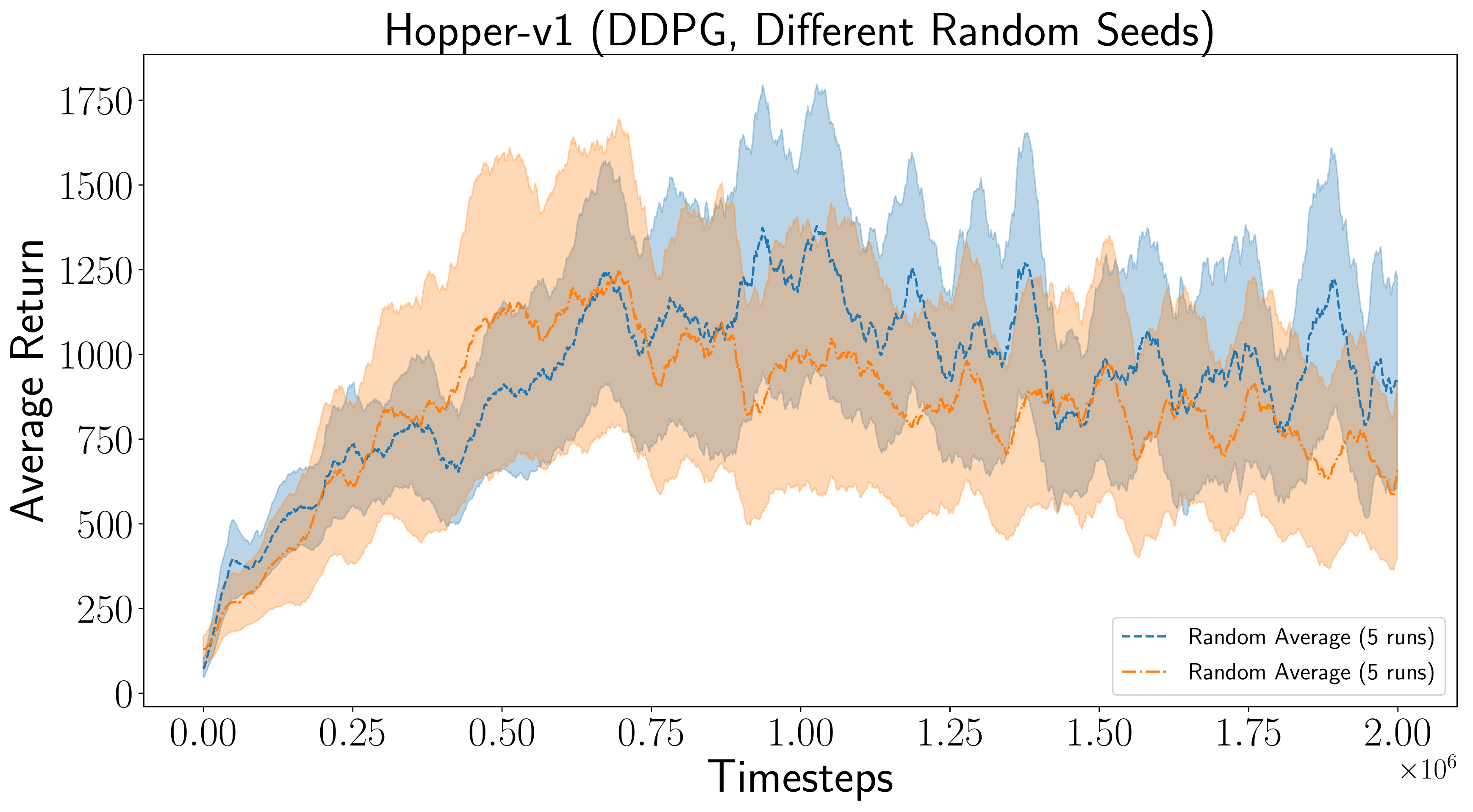}
    \caption{Two different DDPG experiment runs, with same hyperparameter configurations, averaged over two splits of 5 different random seeds. }
    \label{fig:randomseeds_ddpg_sup}
\end{figure}

To determine much random seeds can affect results, we run 10 trials total on two environments using the default previously described settings usign the~\cite{QPROP} implementation of DDPG and the~\cite{rllab} version of TRPO. We divide our trials random into 2 partitions and plot them in Figures~\ref{fig:randomseeds_trposup} and Fig~\ref{fig:randomseeds_ddpg_sup}. As can be seen, statistically different distributions can be attained just from the random seeds with the same exact hyperparameters. As we will discuss later, bootstrapping off of the sample can give an idea for how drastic this effect will be, though too small a bootstrap will still not give concrete enough results.

\section{Choice of Benchmark Continuous Control Environment}

We previously demonstrated that the performance of policy gradient algorithms can be highly biased based on the choice of the environment. In this section, we include further results examining the impact the choice of environment can have. We show that no single algorithm can perform consistenly better in all environments. This is often unlike the results we see with DQN networks in Atari domains, where results can often be demonstrated across a wide range of Atari games. Our results, for example, shows that while TRPO can perform significantly better than other algorithms on the Swimmer environment, it may perform quite poorly n the HalfCheetah environment, and marginally better on the Hopper environment compared to PPO. We demonstrate our results using the OpenAI MuJoCo Gym environments including Hopper, HalfCheetah, Swimmer and Walker environments. It is notable to see the varying performance these algorithms can have even in this small set of environment domains. The choice of reporting algorithm performance results can therefore often be biased based on the algorithm designer's experience with these environments.

\begin{figure}[H]
    \centering
    \includegraphics[width=.49\textwidth]{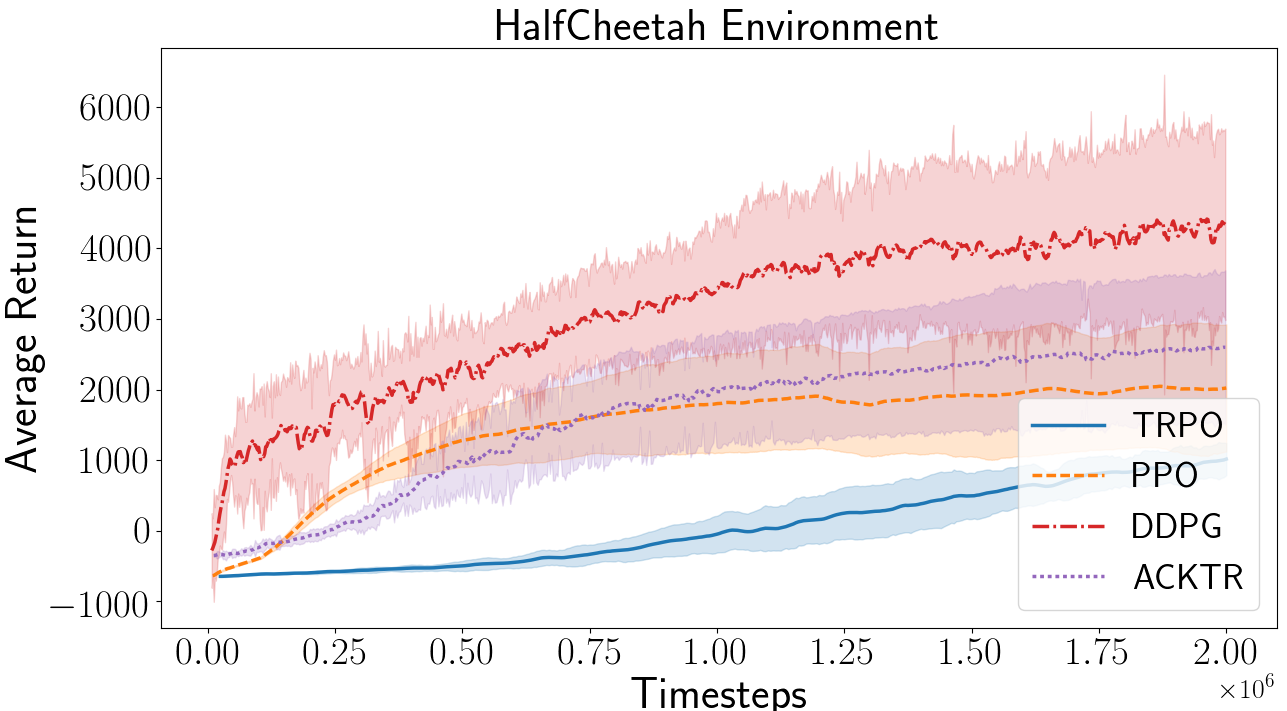}
    \includegraphics[width=.49\textwidth]{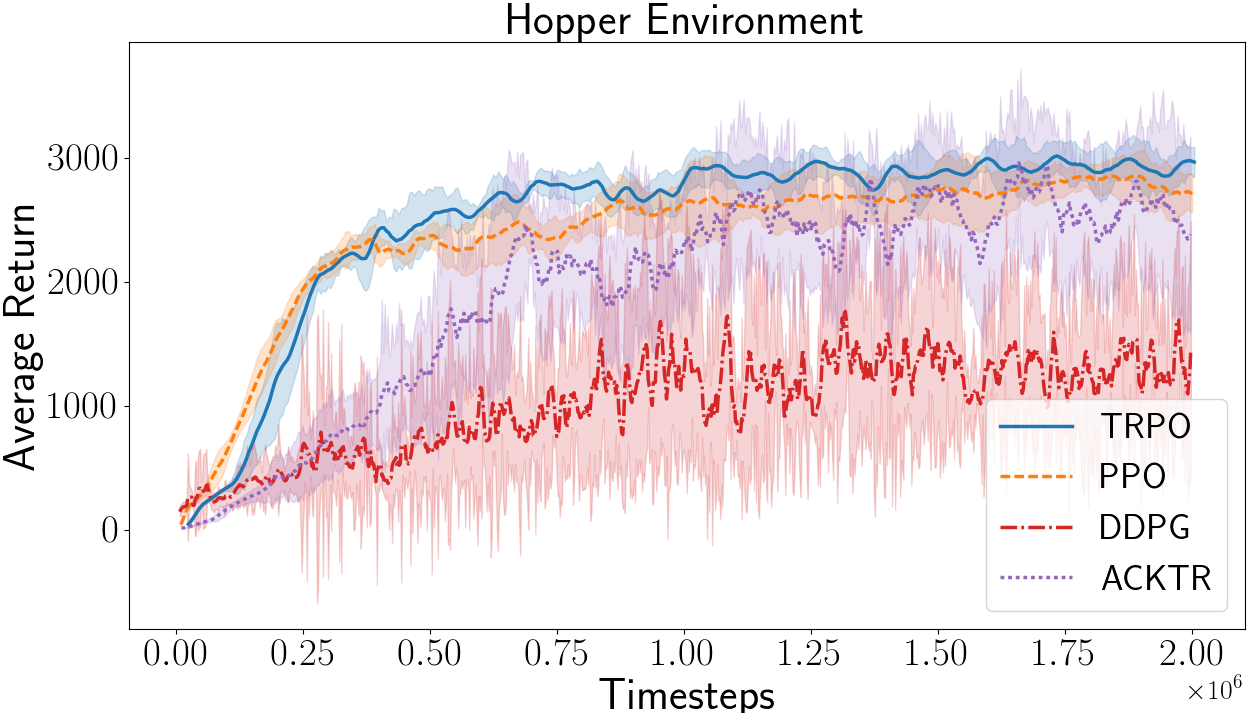}
\end{figure}

\begin{figure}[H]
    \includegraphics[width=.49\textwidth]{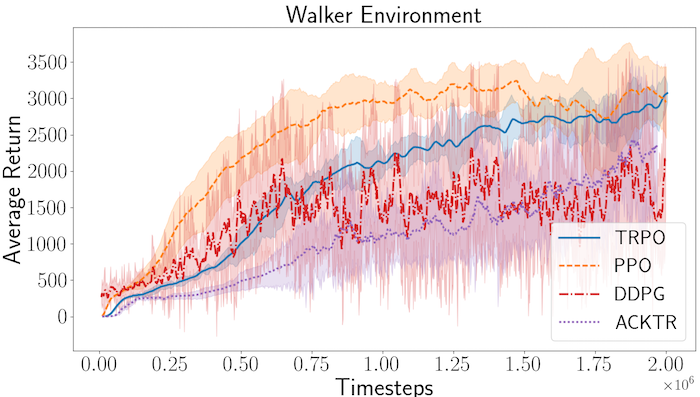}
    \includegraphics[width=.49\textwidth]{images/Swimmer_Env_Algos.png}
    \caption{Comparing Policy Gradients across various environments}
    \label{fig:halfcheetahenv}
\end{figure}

\section{Codebases}

We include a detailed analysis of performance comparison, with different network structures and activations, based on the choice of the algorithm implementation codebase.

\begin{figure}[H]
    \centering
    \includegraphics[width=.49\textwidth]{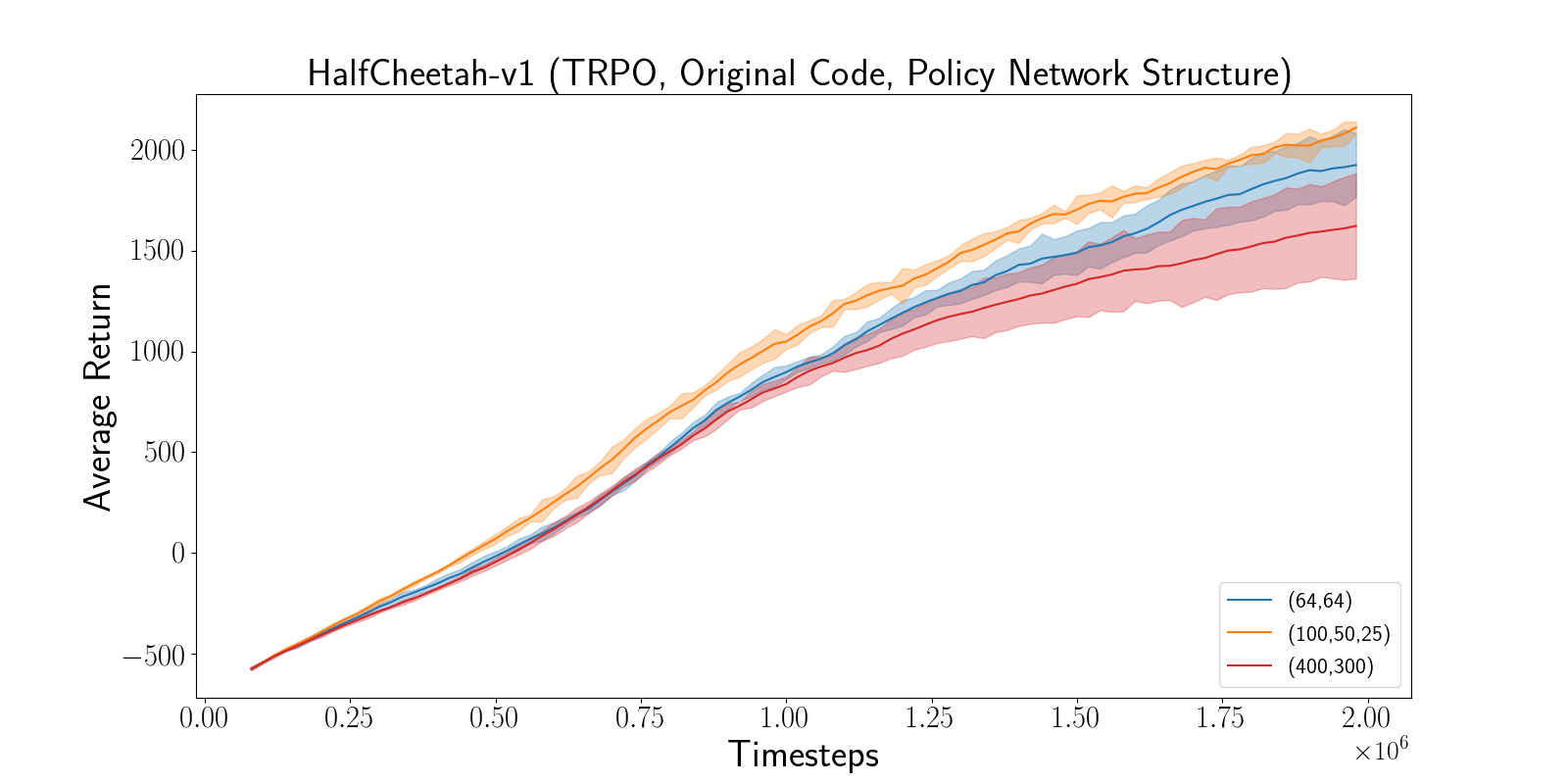}
    \includegraphics[width=.49\textwidth]{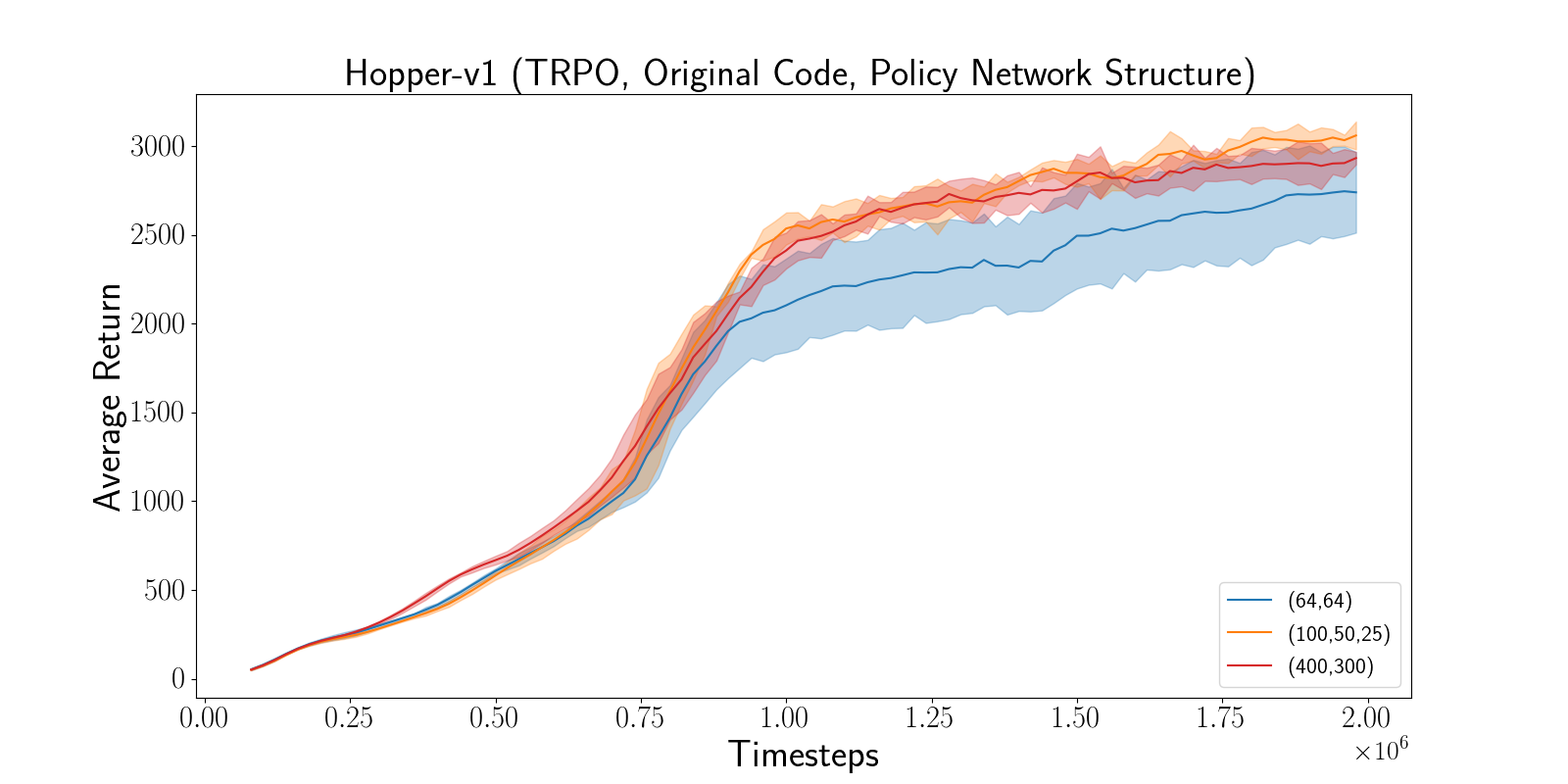}
    \includegraphics[width=.49\textwidth]{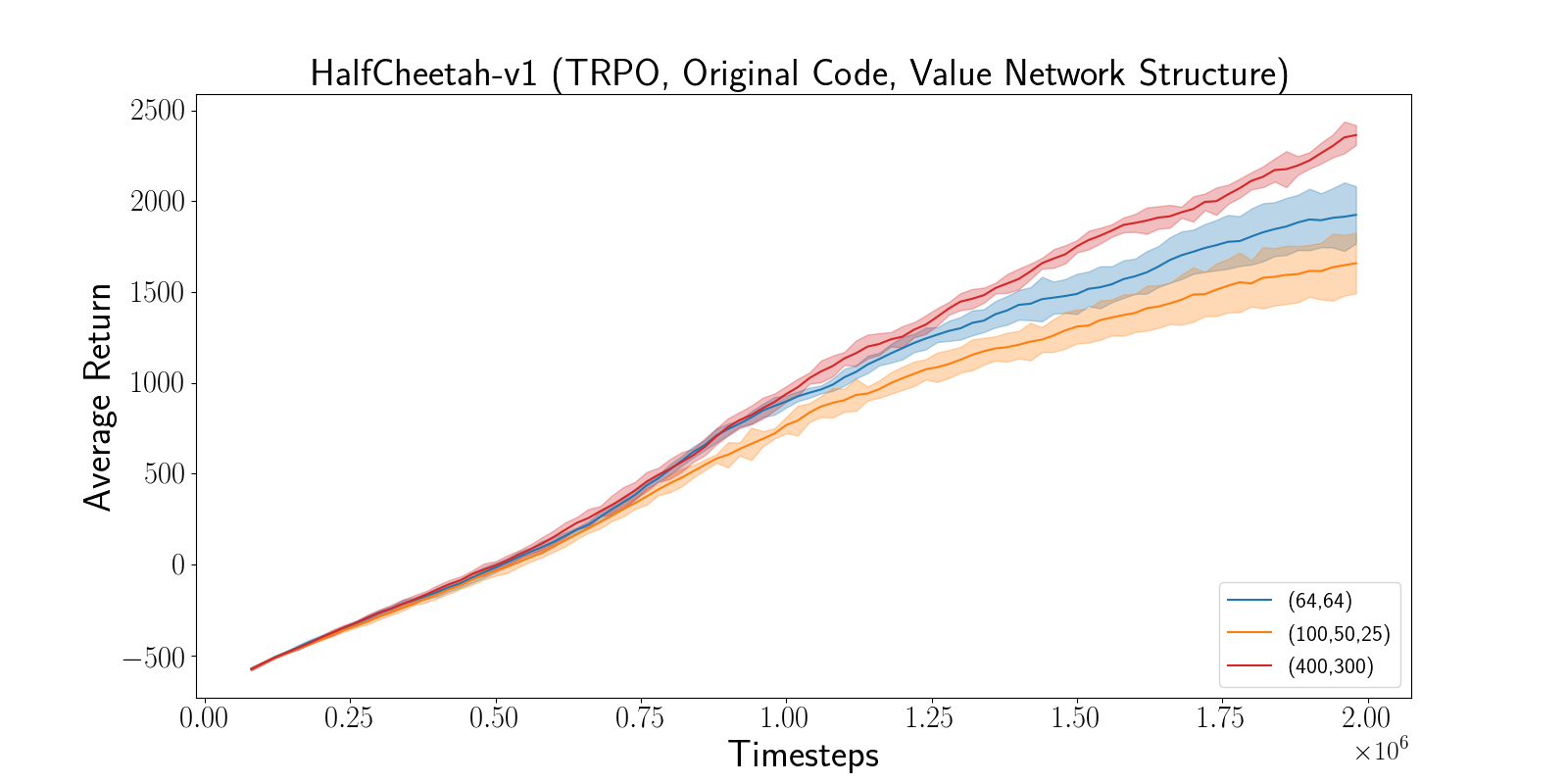}
    \includegraphics[width=.49\textwidth]{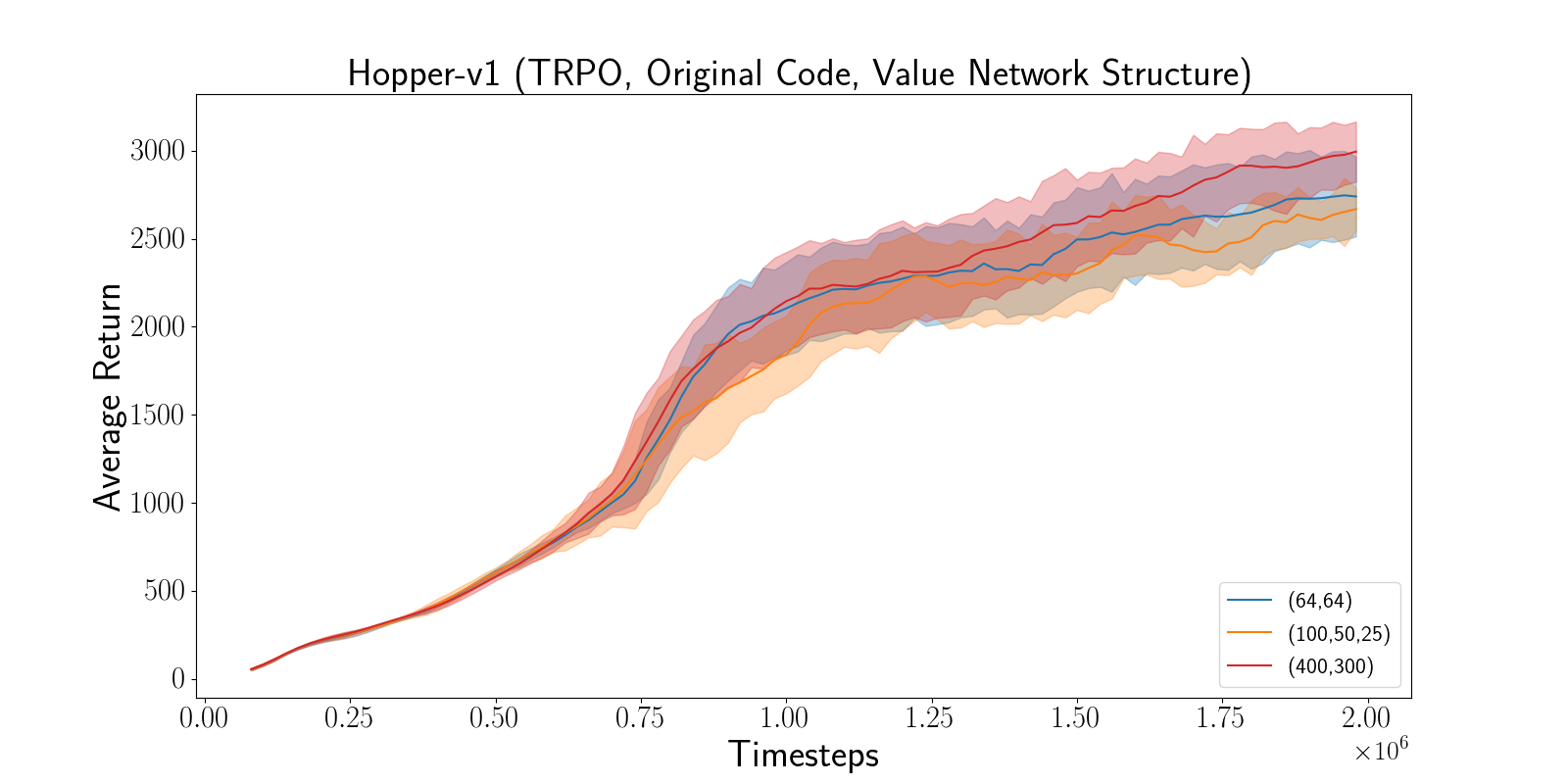}
    \caption{TRPO Policy and Value Network structure}
    \label{trpo_original}
\end{figure}

\begin{figure}[H]
    \centering
    \includegraphics[width=.49\textwidth]{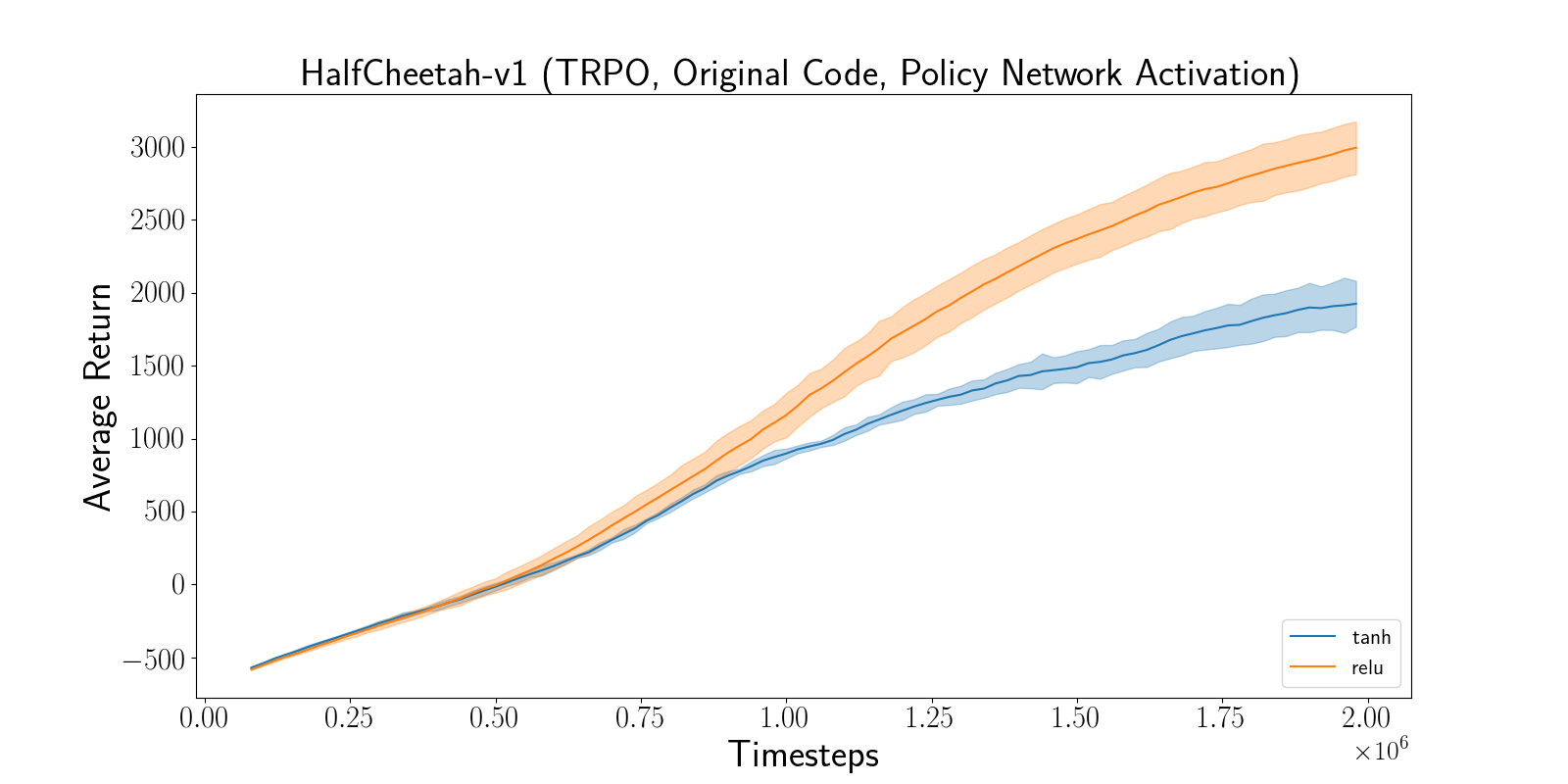}
    \includegraphics[width=.49\textwidth]{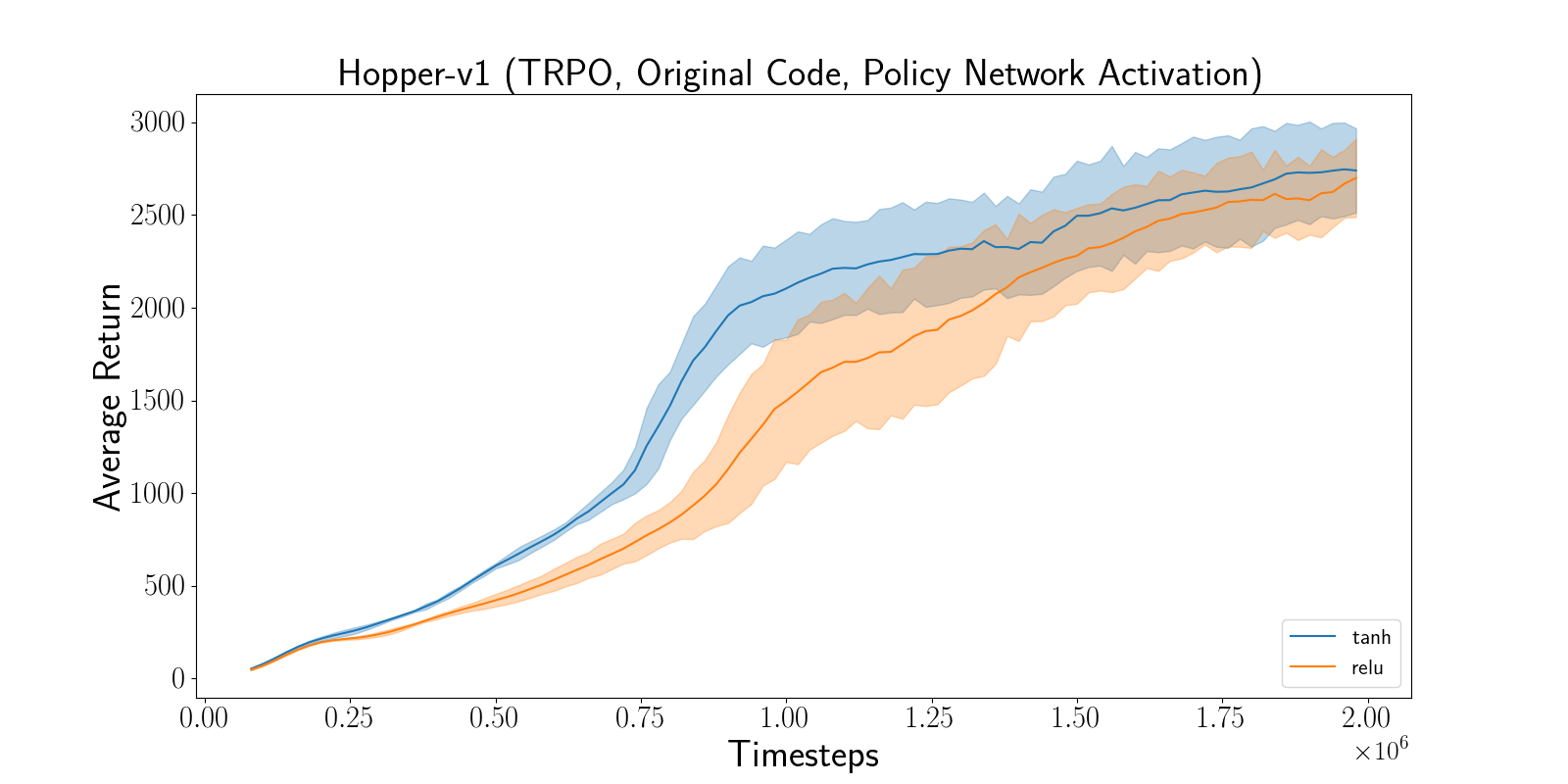}
    \includegraphics[width=.49\textwidth]{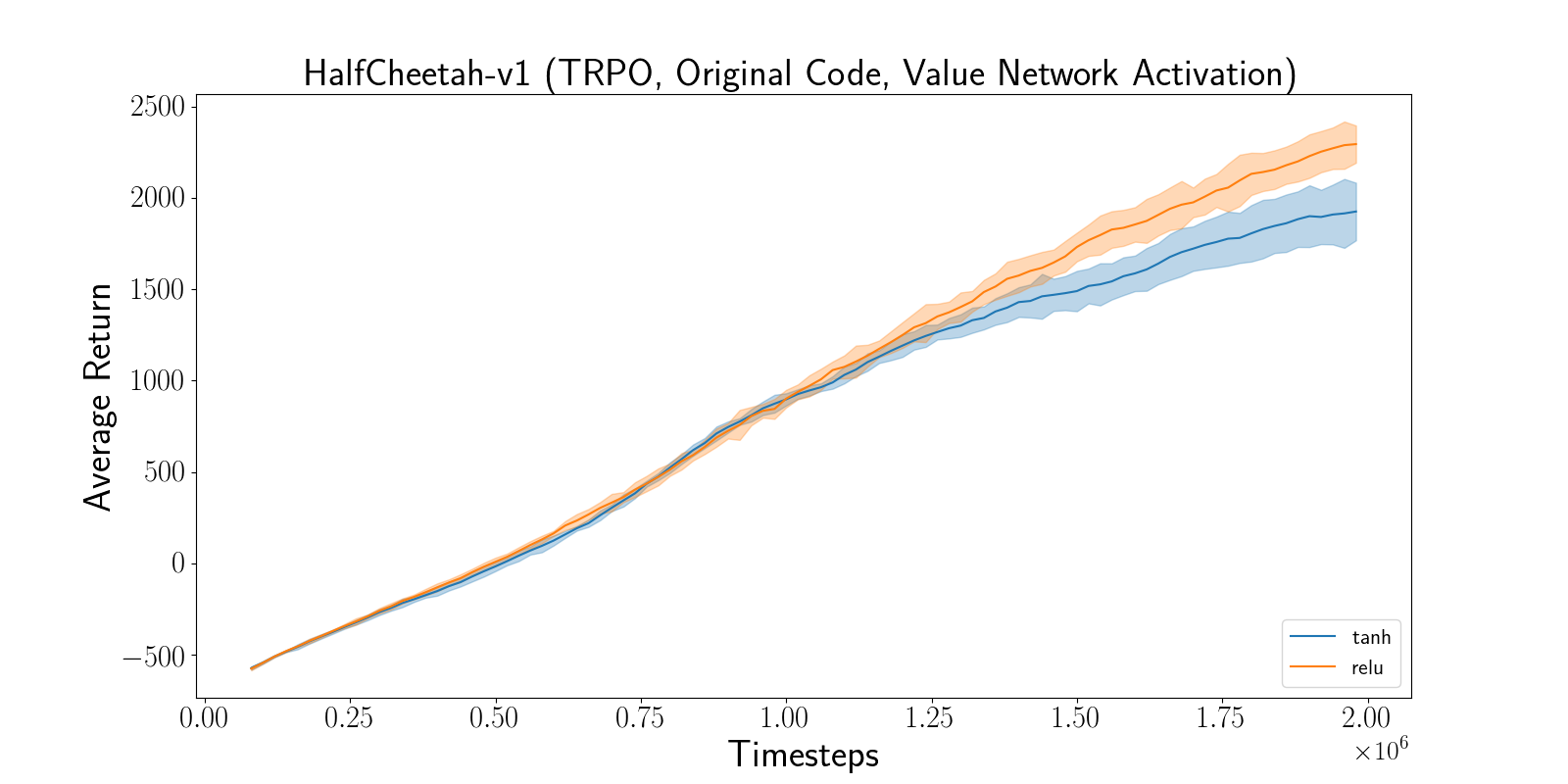}
    \includegraphics[width=.49\textwidth]{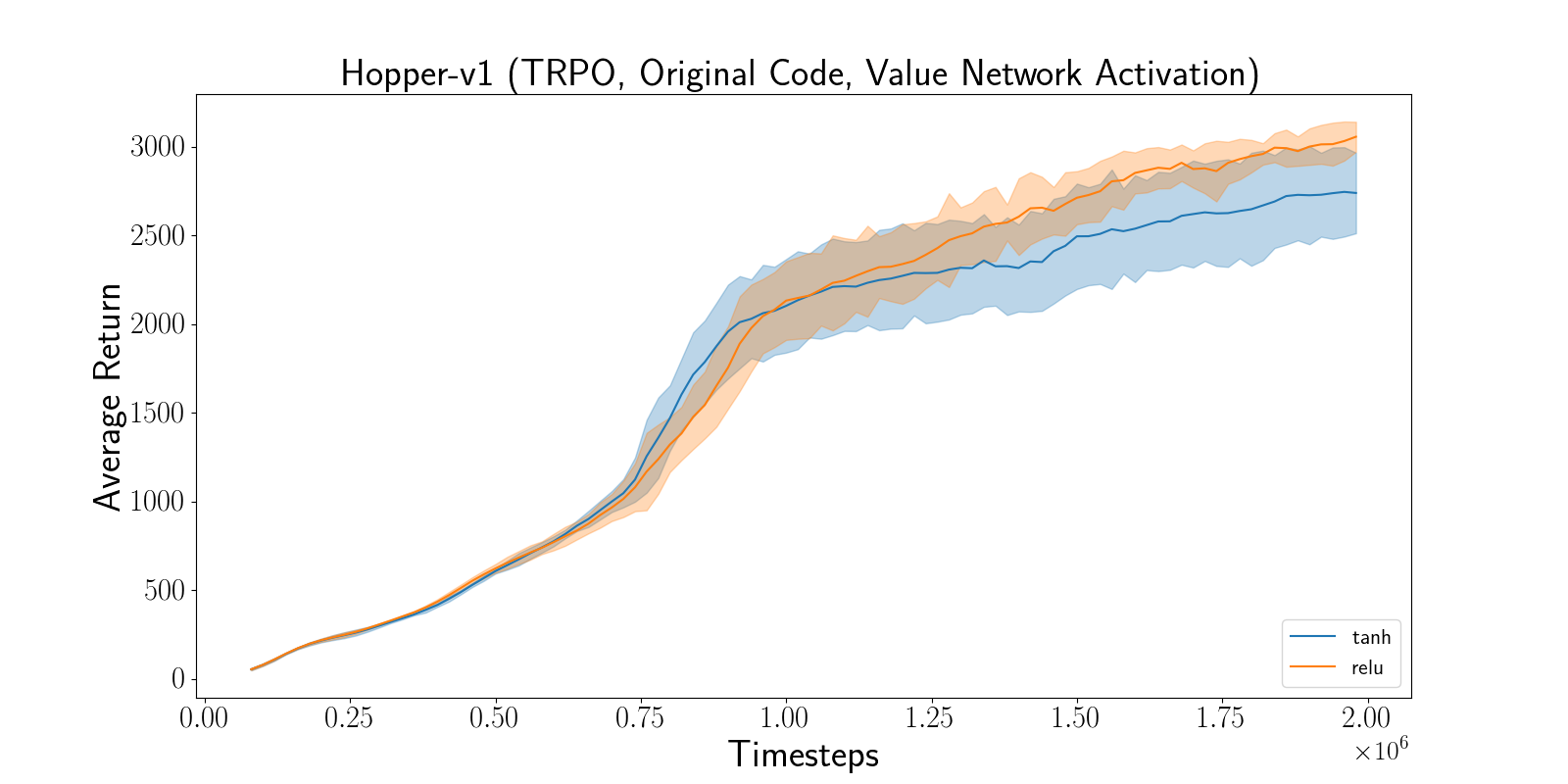}
    \caption{TRPO Policy and Value Network activations.}
    \label{trpo_original_2}
\end{figure}

\begin{figure}[H]
    \centering
    \includegraphics[width=.49\textwidth]{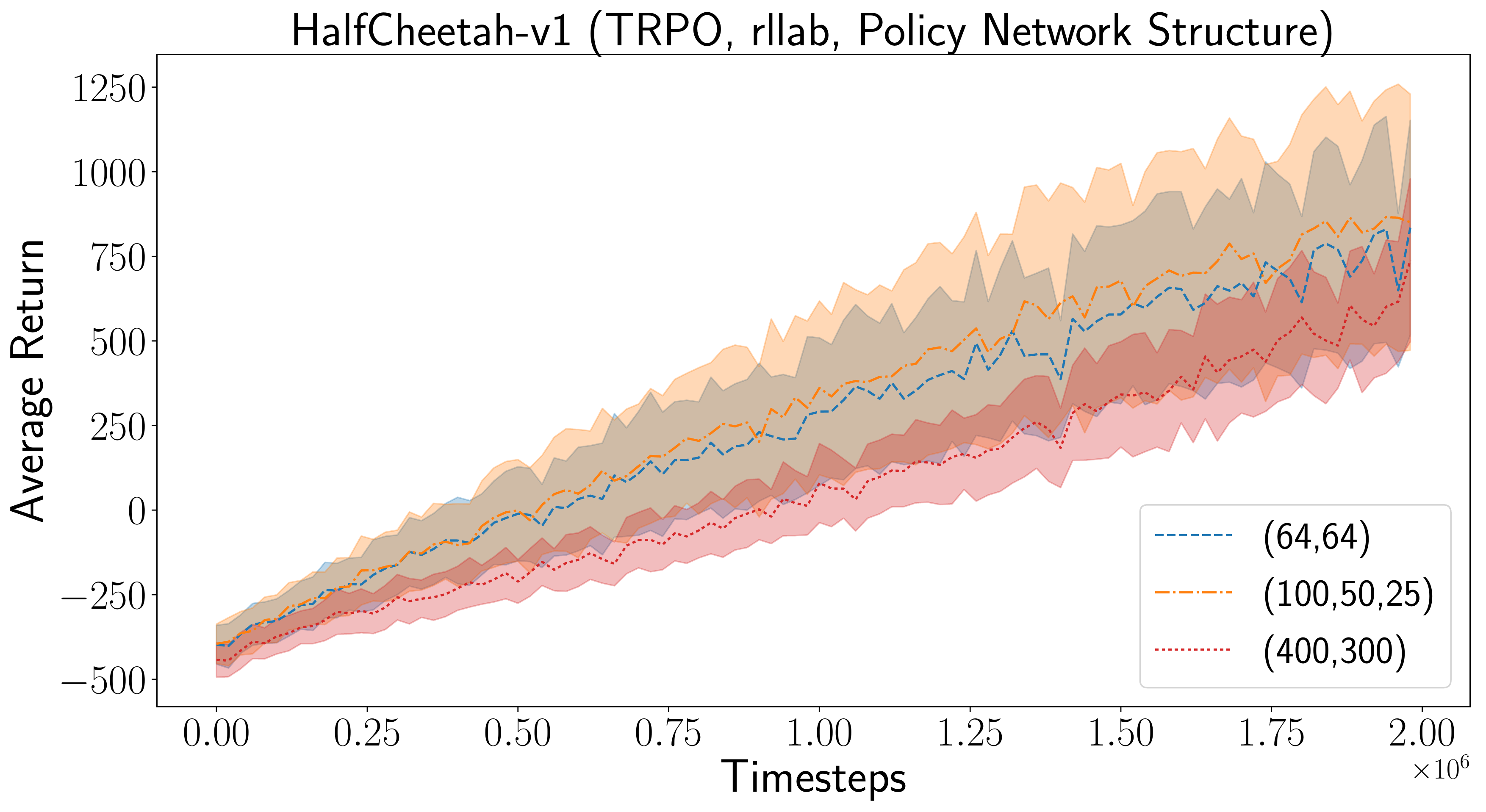}
    \includegraphics[width=.49\textwidth]{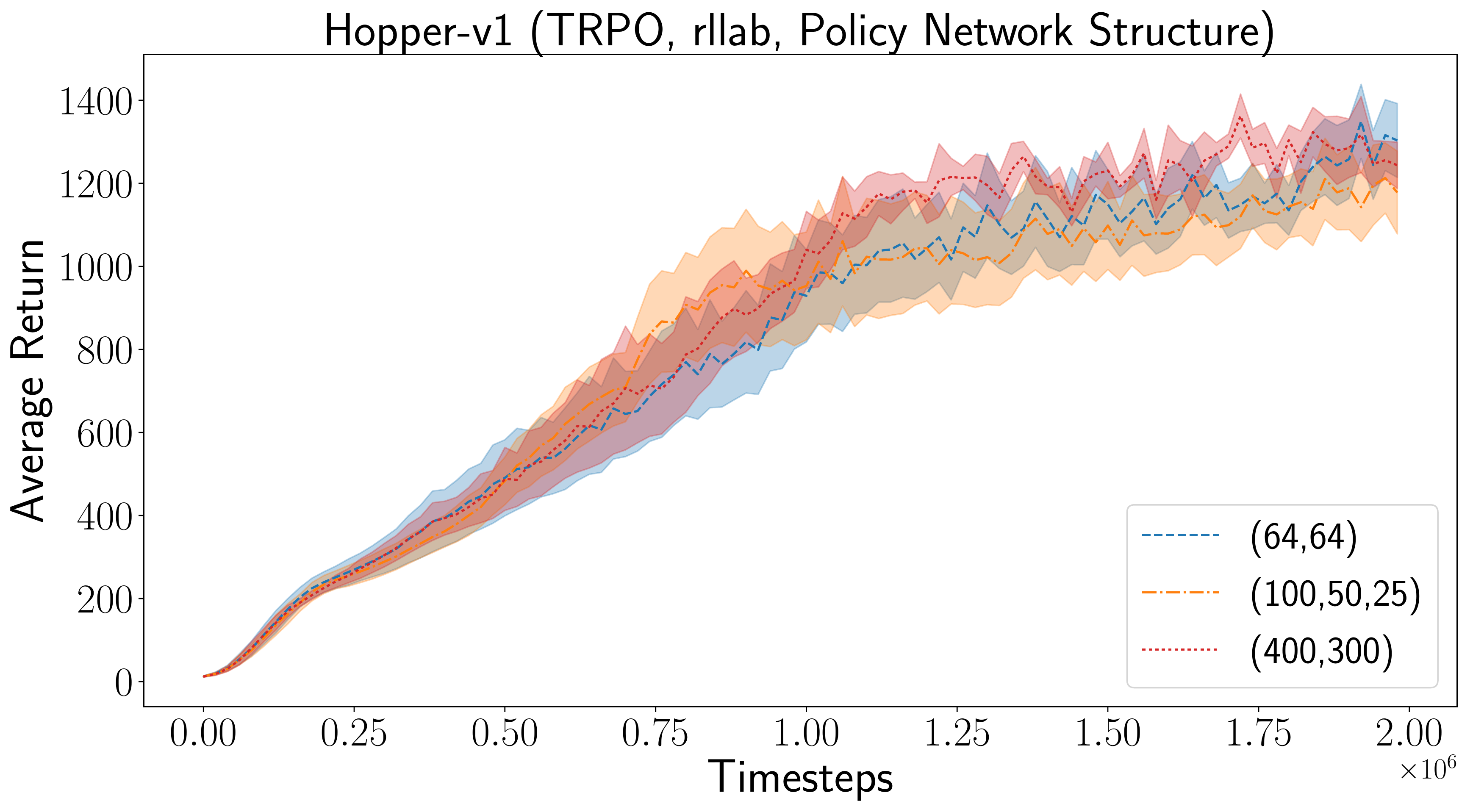}
    \includegraphics[width=.49\textwidth]{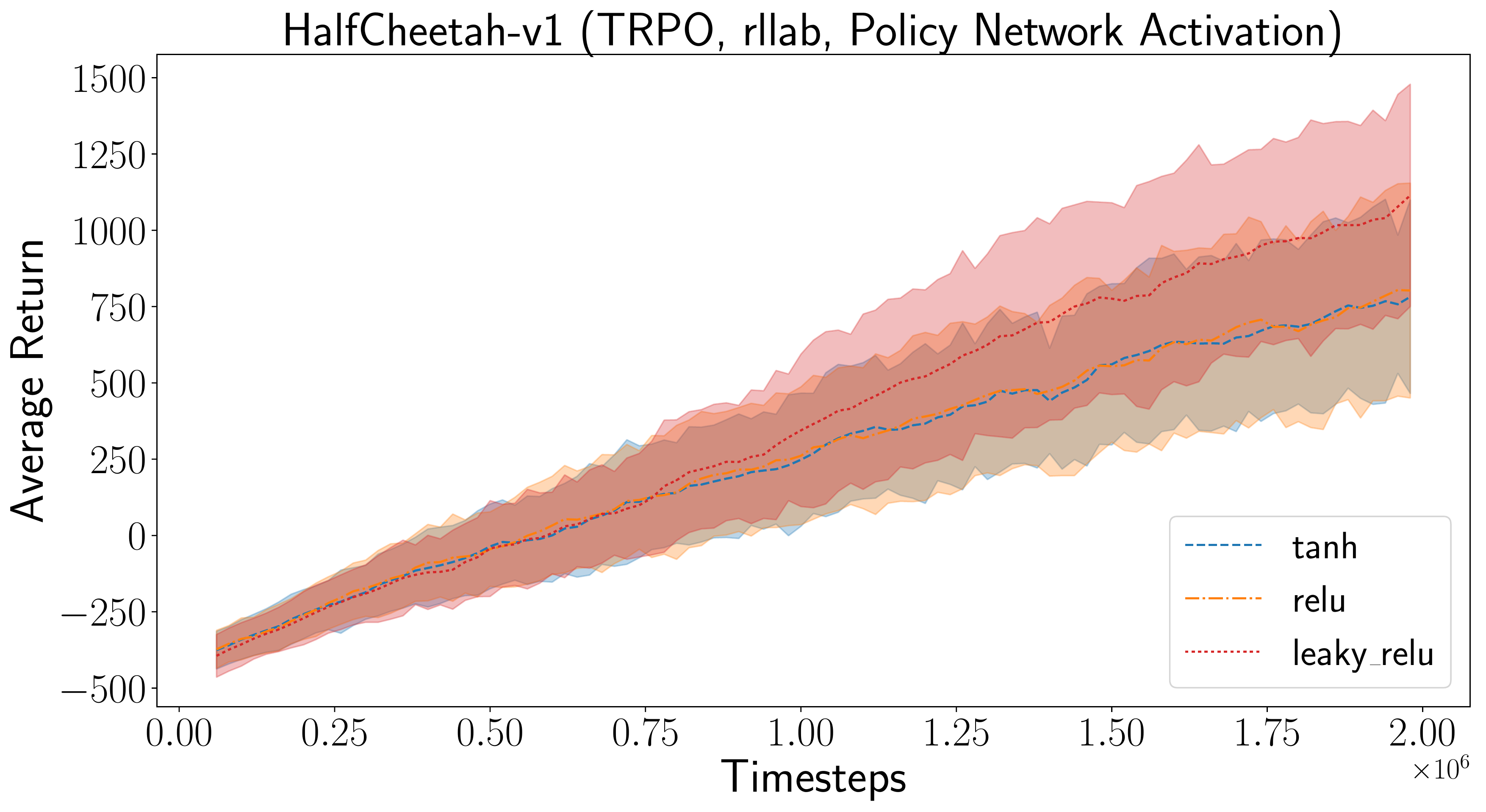}
    \includegraphics[width=.49\textwidth]{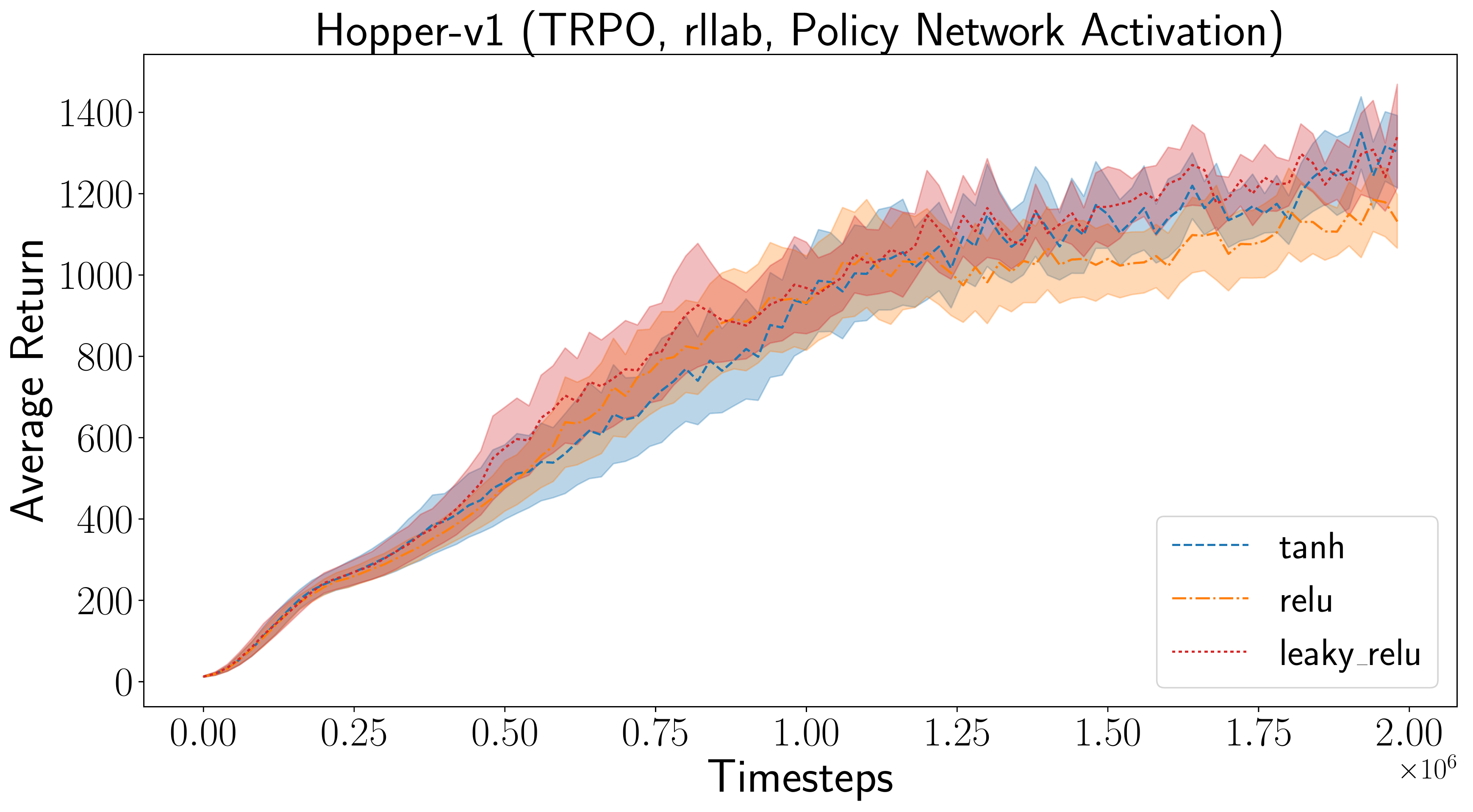}
    \caption{TRPO rllab Policy Structure and Activation}
    \label{fig:trpo_rllab}
\end{figure}

\begin{figure}[H]
    \centering
    \includegraphics[width=.49\textwidth]{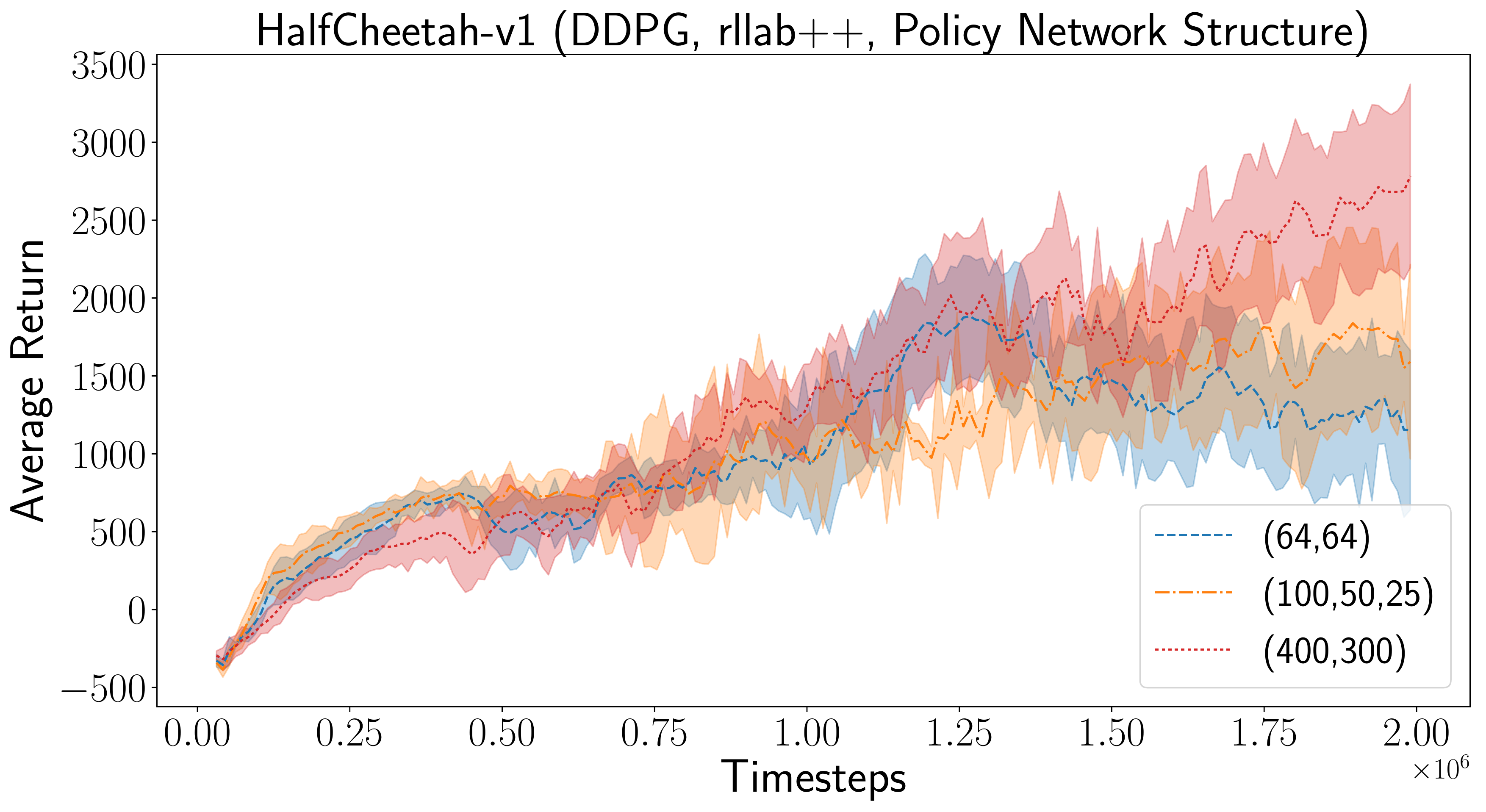}
    \includegraphics[width=.49\textwidth]{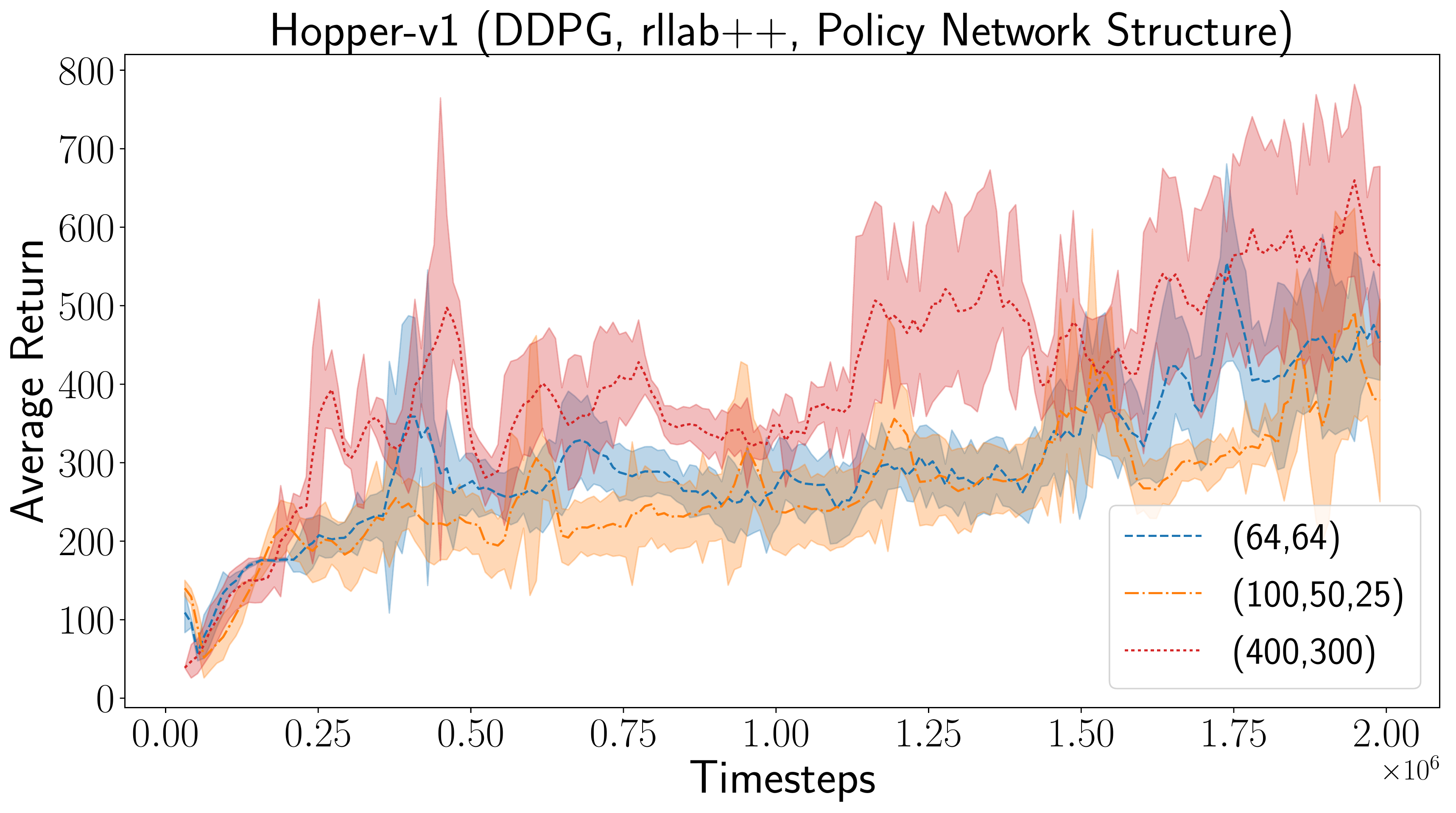}
    \includegraphics[width=.49\textwidth]{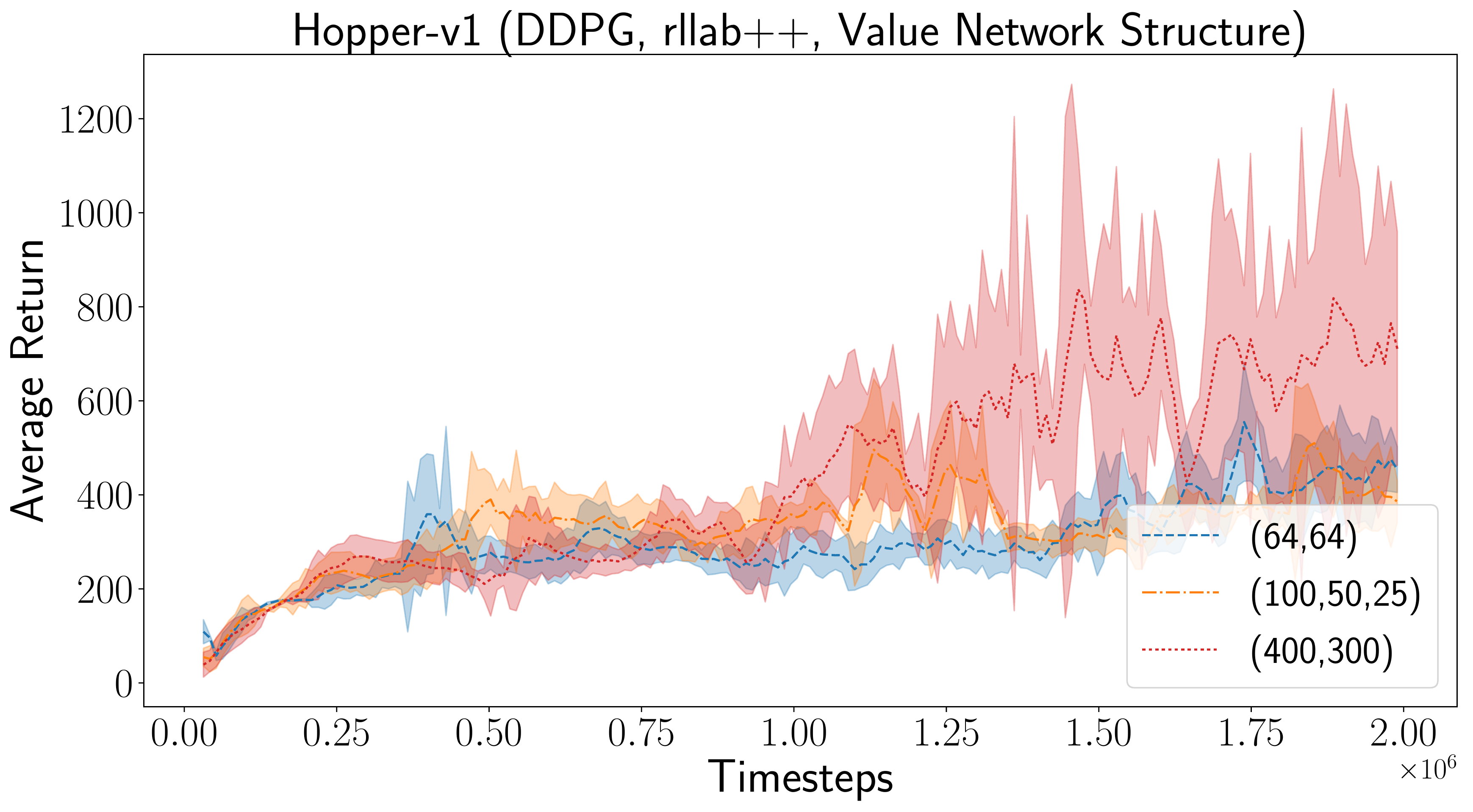}
    \includegraphics[width=.49\textwidth]{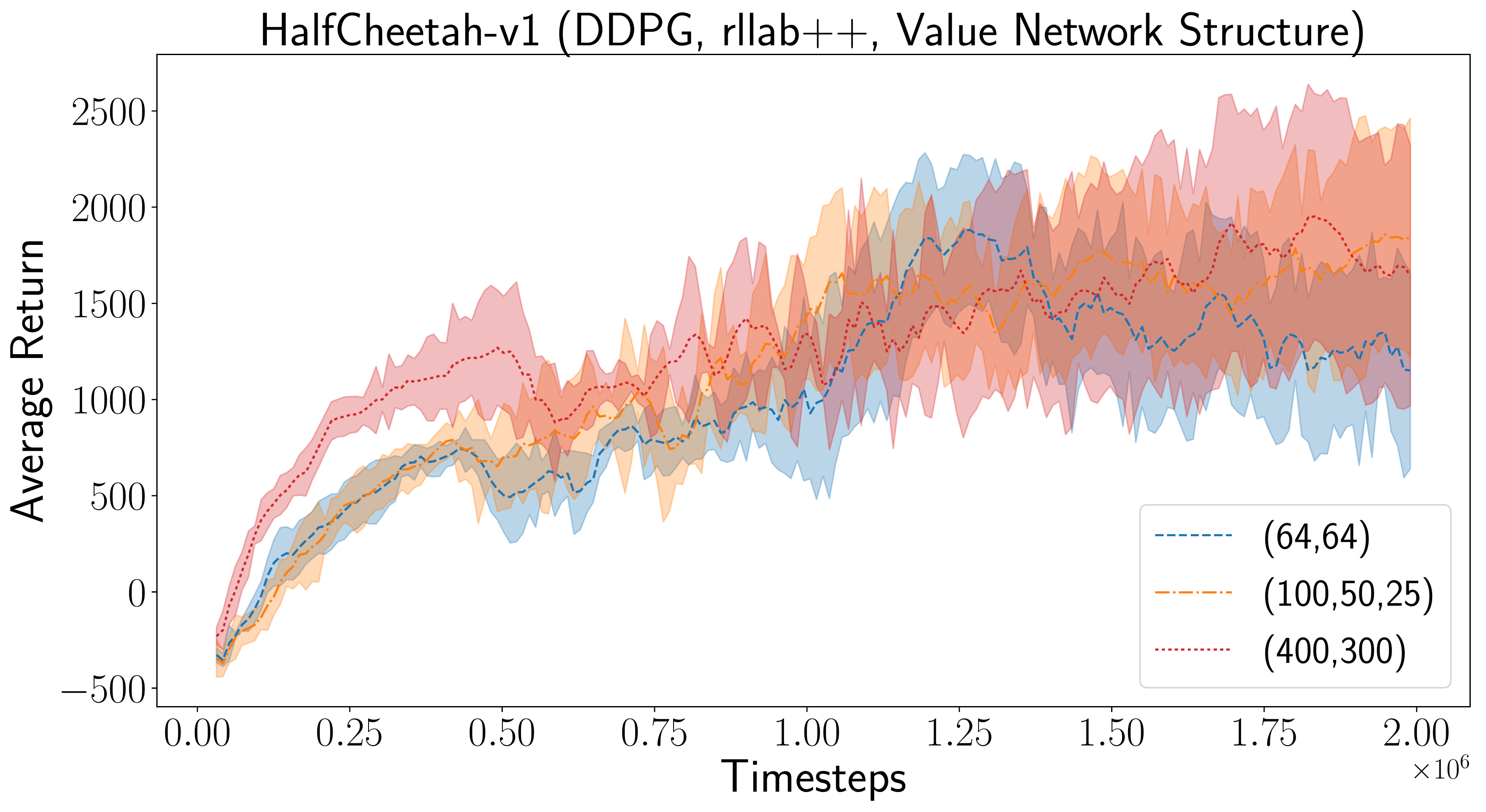}

    \caption{DDPG rllab++ Policy and Value Network structure}
    \label{ddpg_rllab_pp}
\end{figure}

\begin{figure}[H]
    \centering
    \includegraphics[width=.49\textwidth]{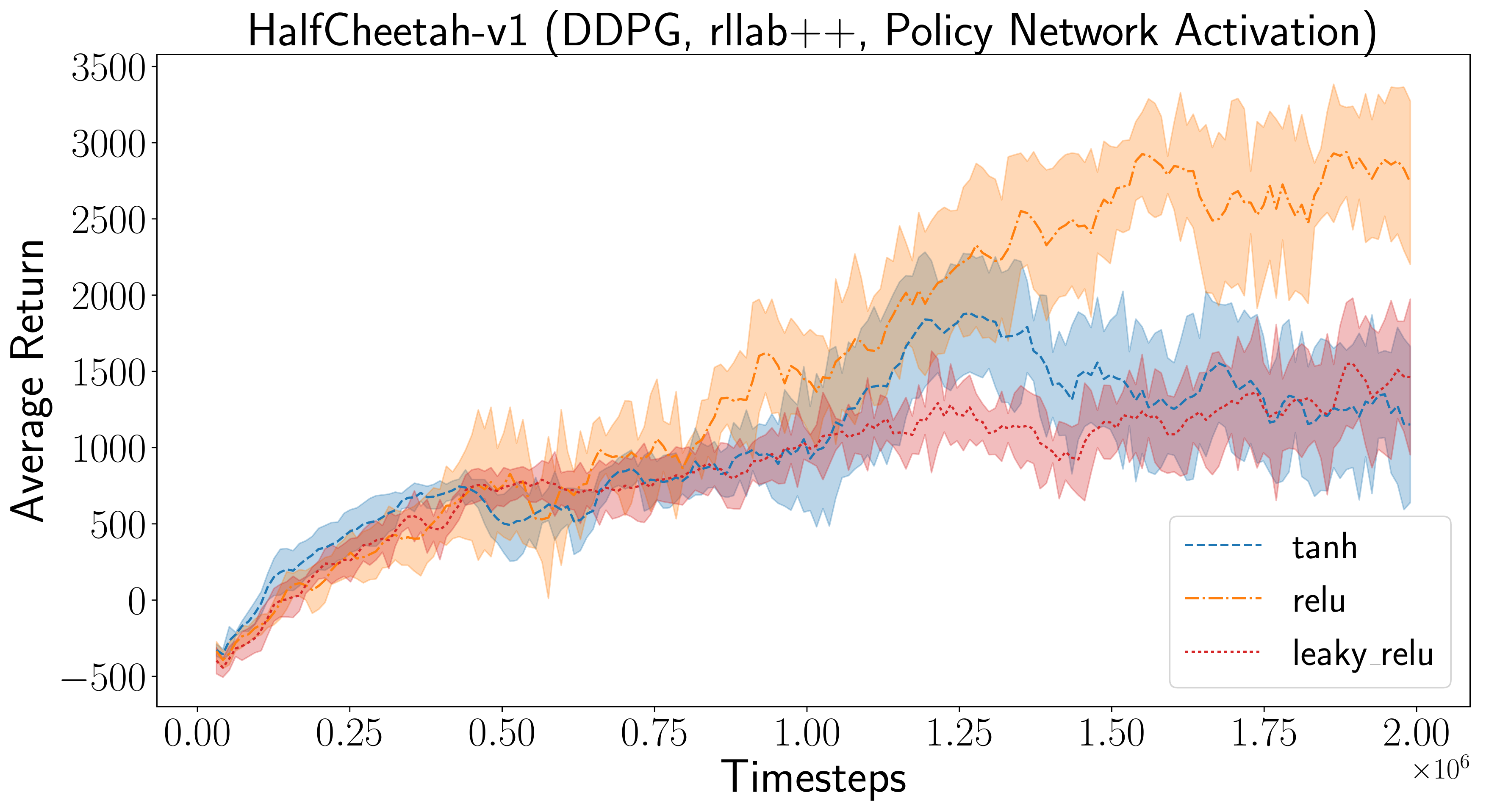}
    \includegraphics[width=.49\textwidth]{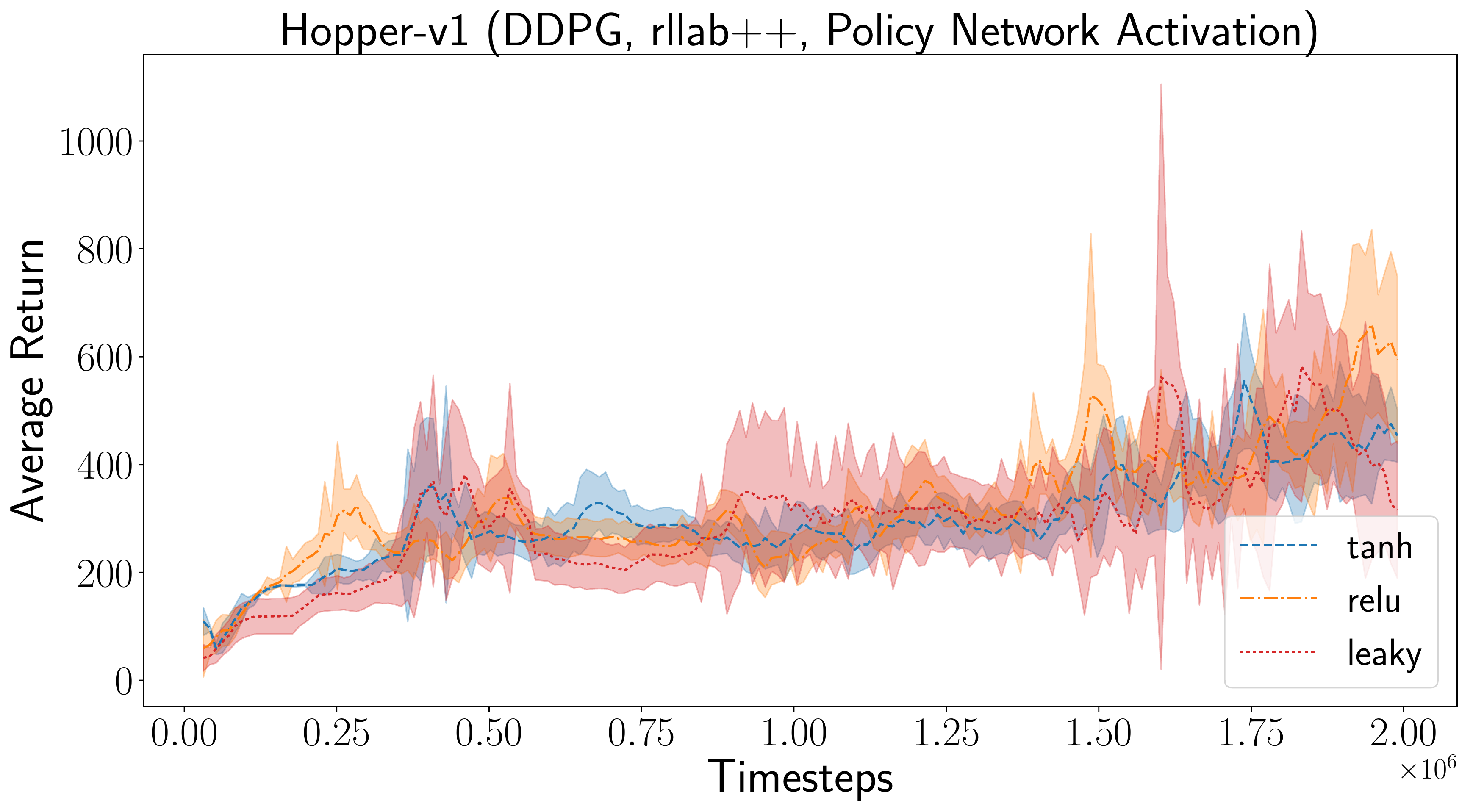}
    \includegraphics[width=.49\textwidth]{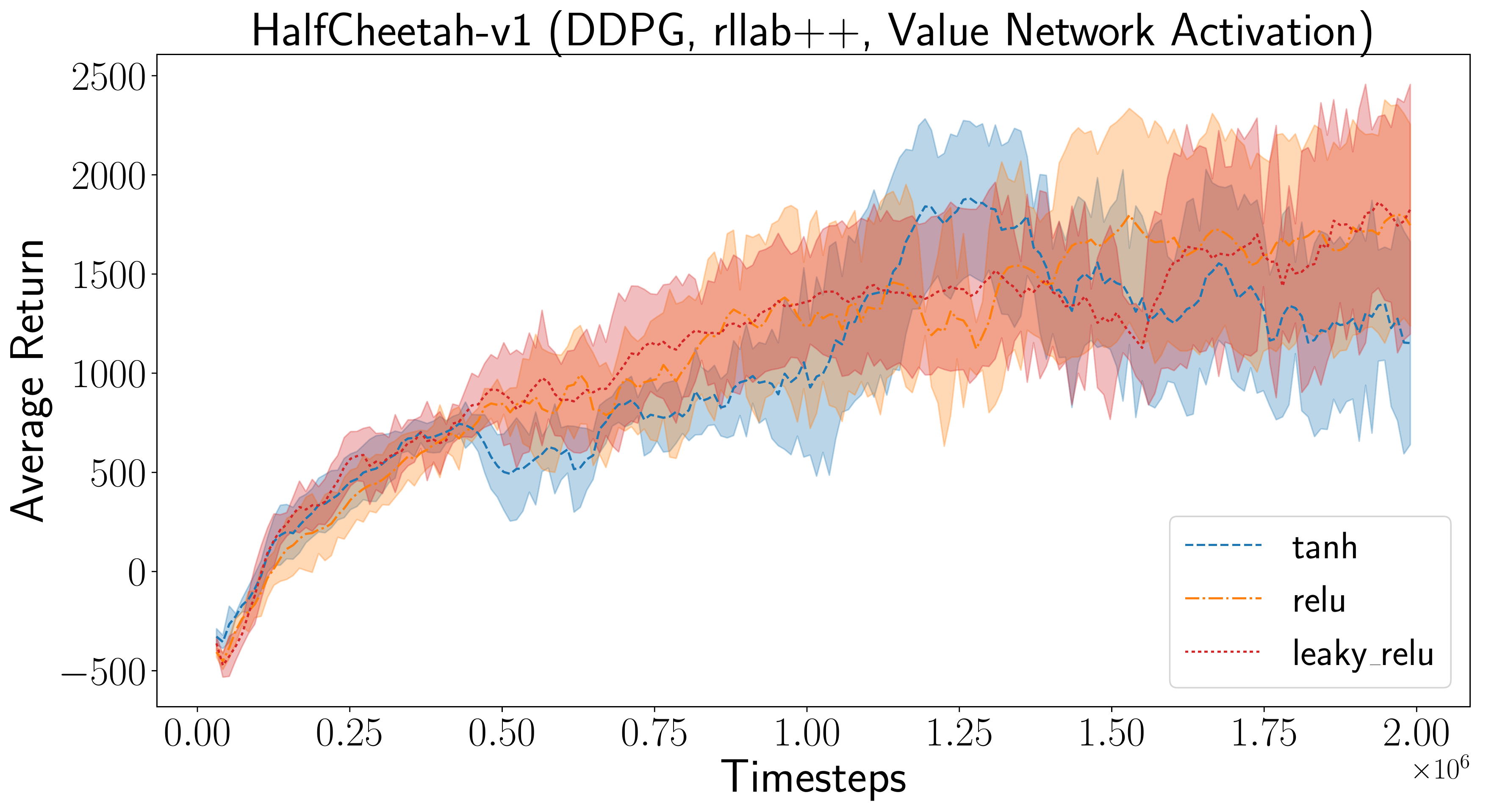}
    \includegraphics[width=.49\textwidth]{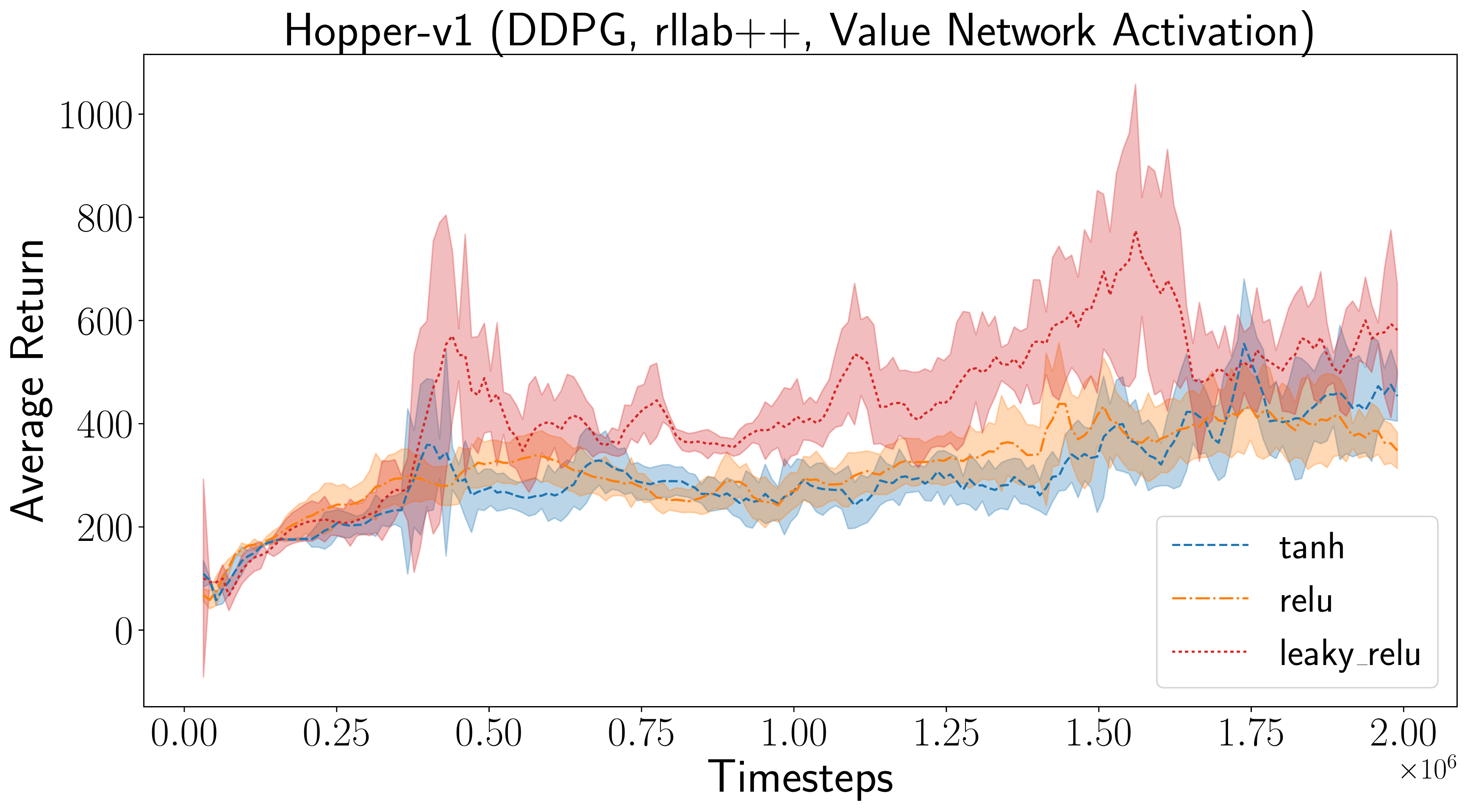}
    \caption{DDPG rllab++ Policy and Value Network activations.}
    \label{ddpg_rllab_pp_2}
\end{figure}

\begin{figure}[H]
    \centering
    \includegraphics[width=.49\textwidth]{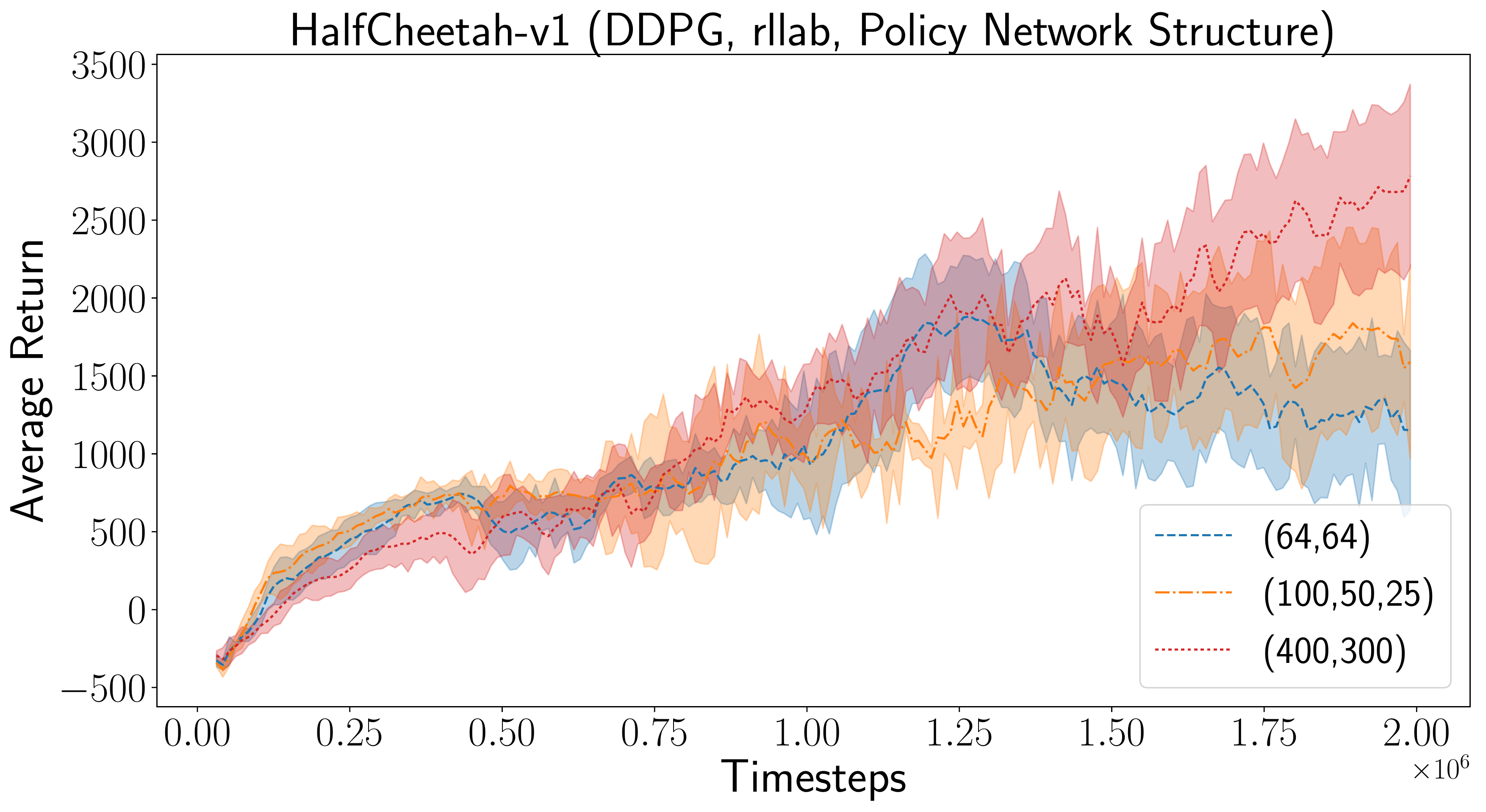}
    \includegraphics[width=.49\textwidth]{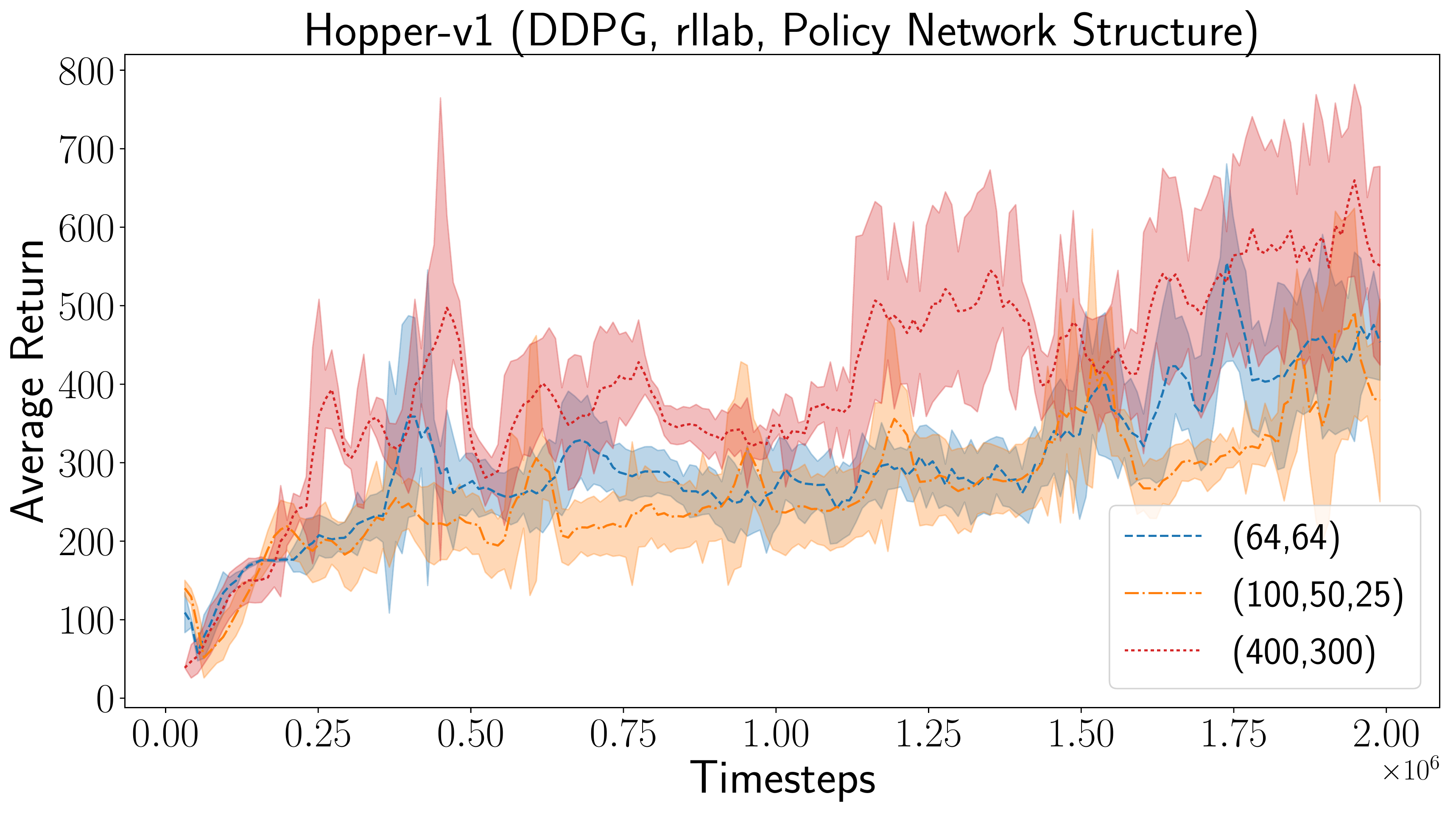}
    \includegraphics[width=.49\textwidth]{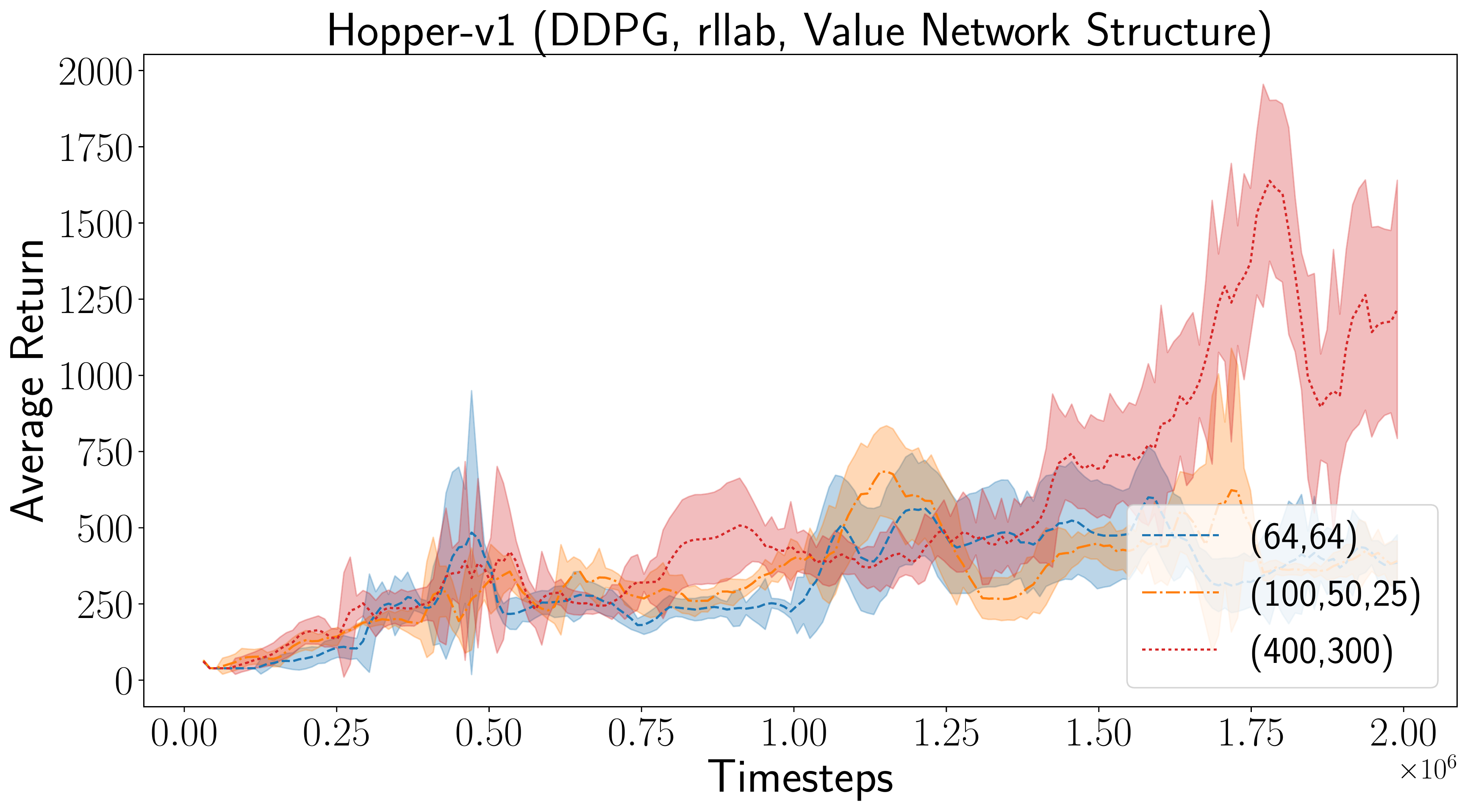}
    \includegraphics[width=.49\textwidth]{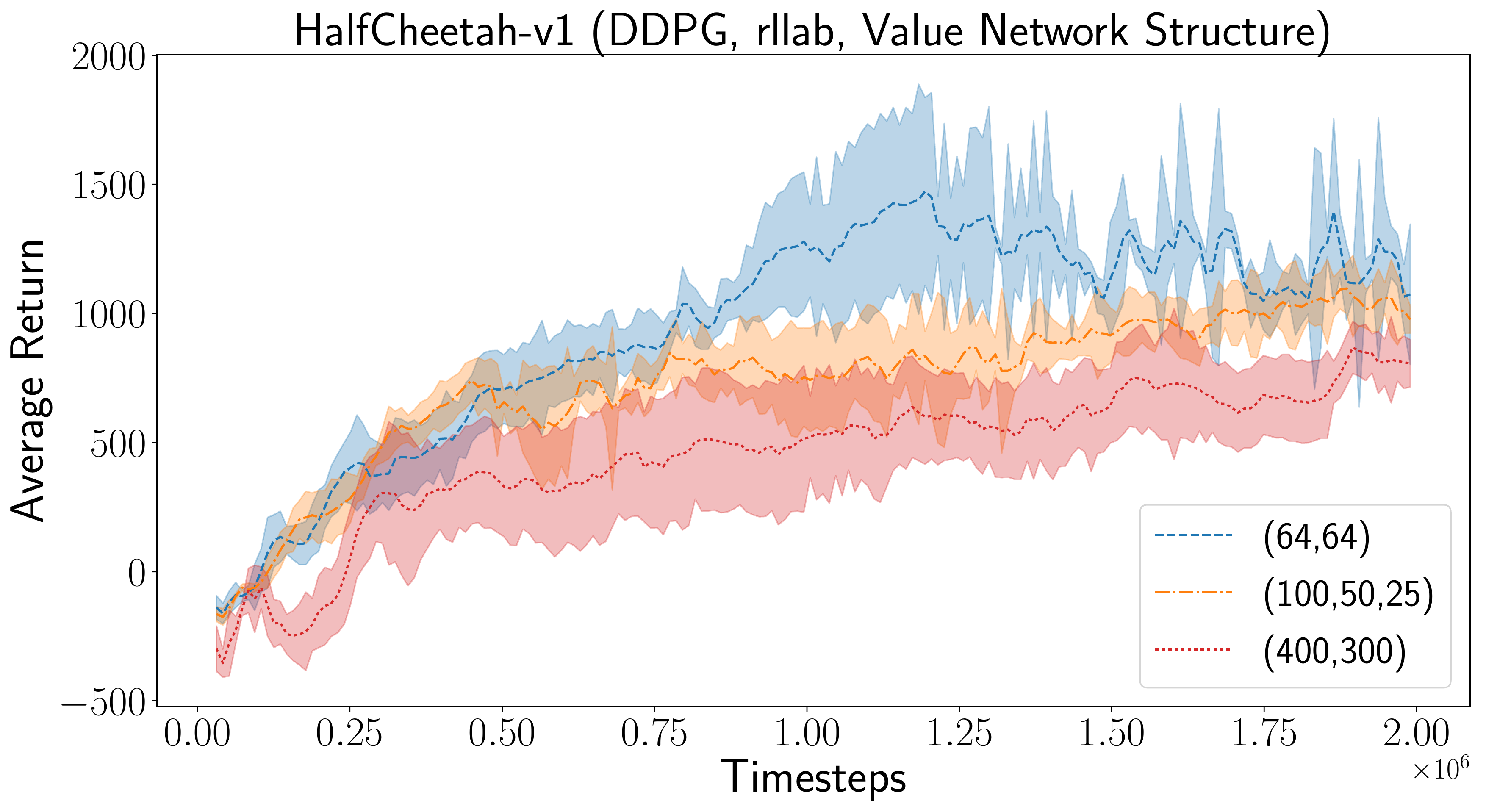}
    \caption{DDPG rllab Policy and Value Network structure}
    \label{ddpg_rllab}
\end{figure}

\begin{figure}[H]
    \centering
    \includegraphics[width=.49\textwidth]{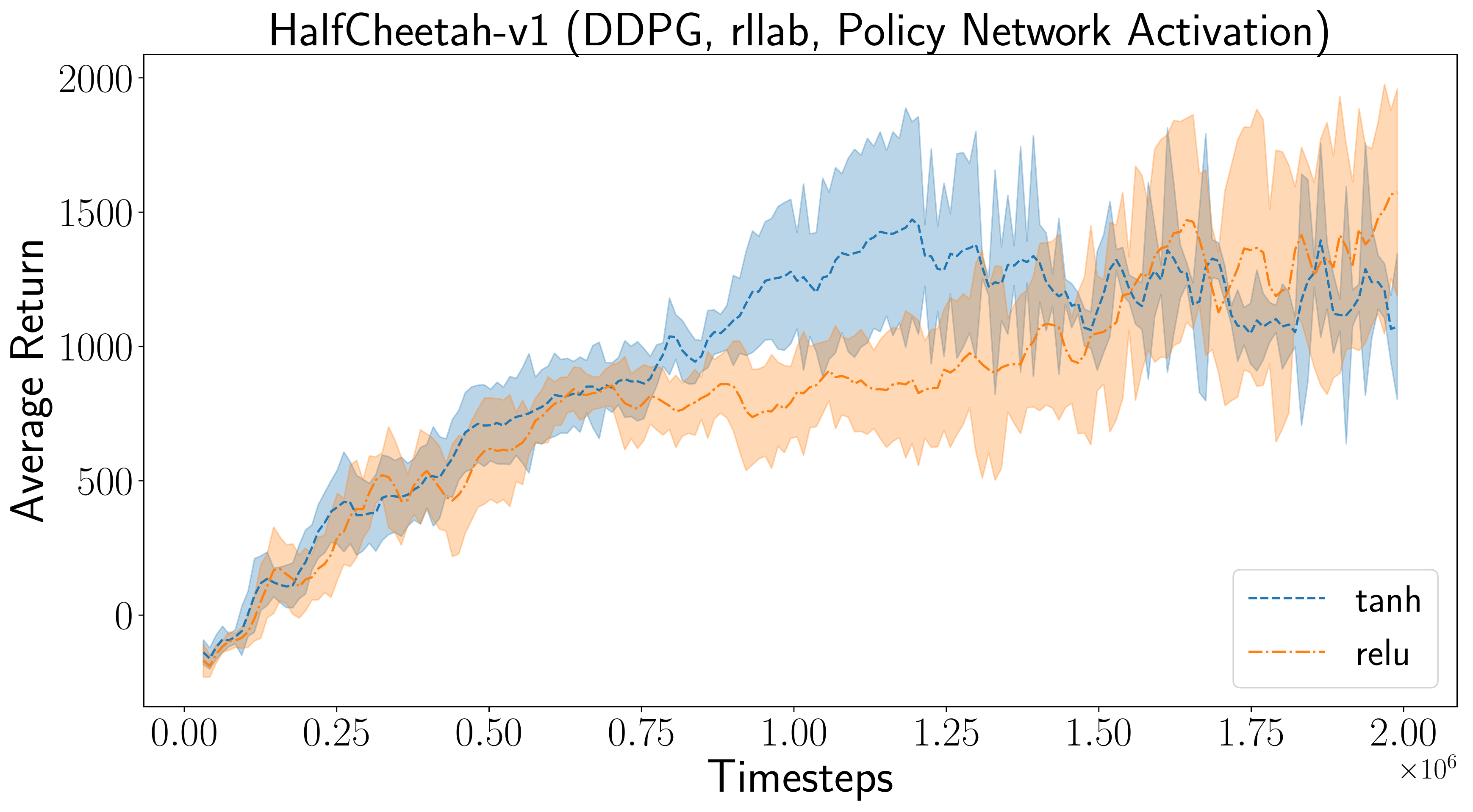}
    \includegraphics[width=.49\textwidth]{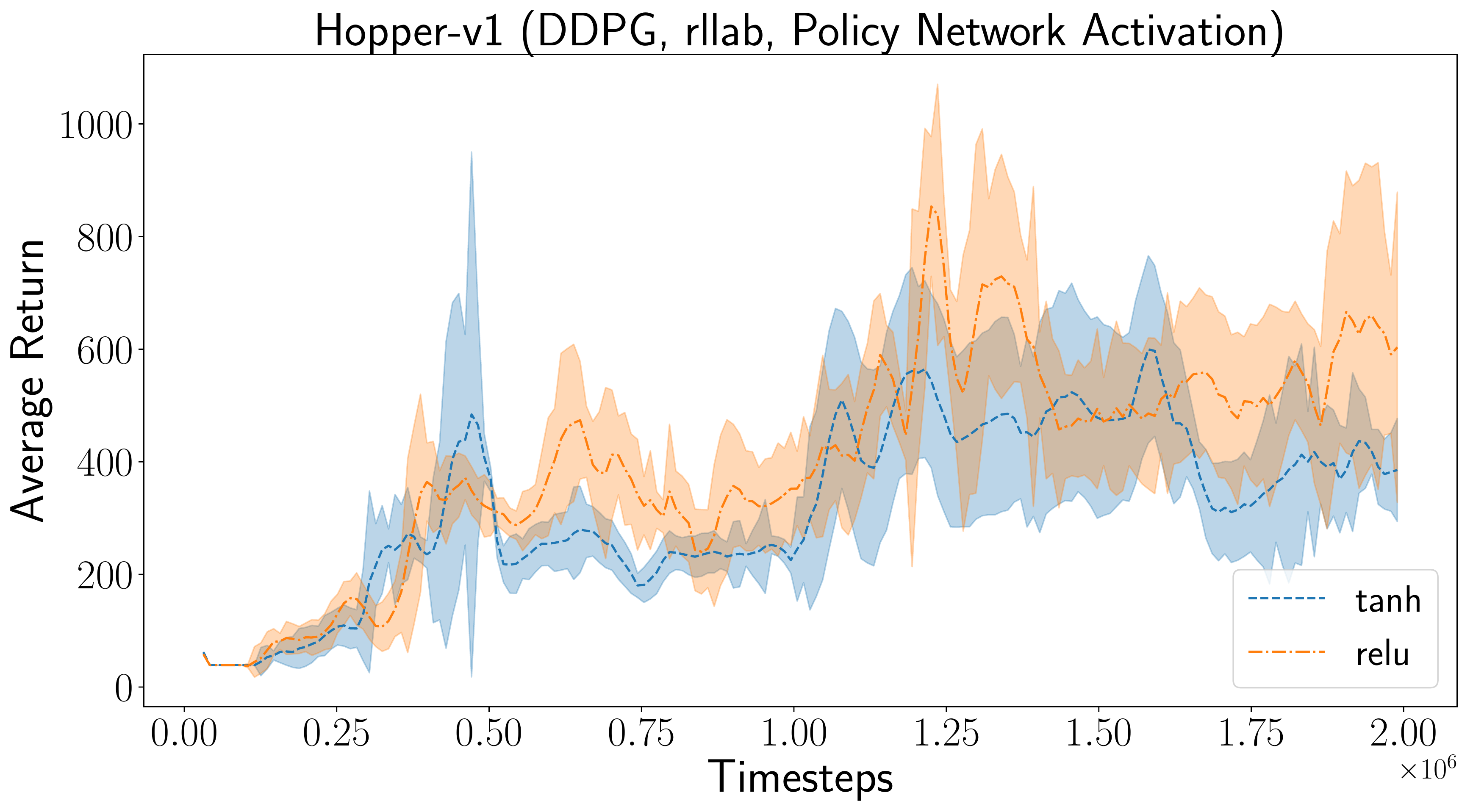}
    \includegraphics[width=.49\textwidth]{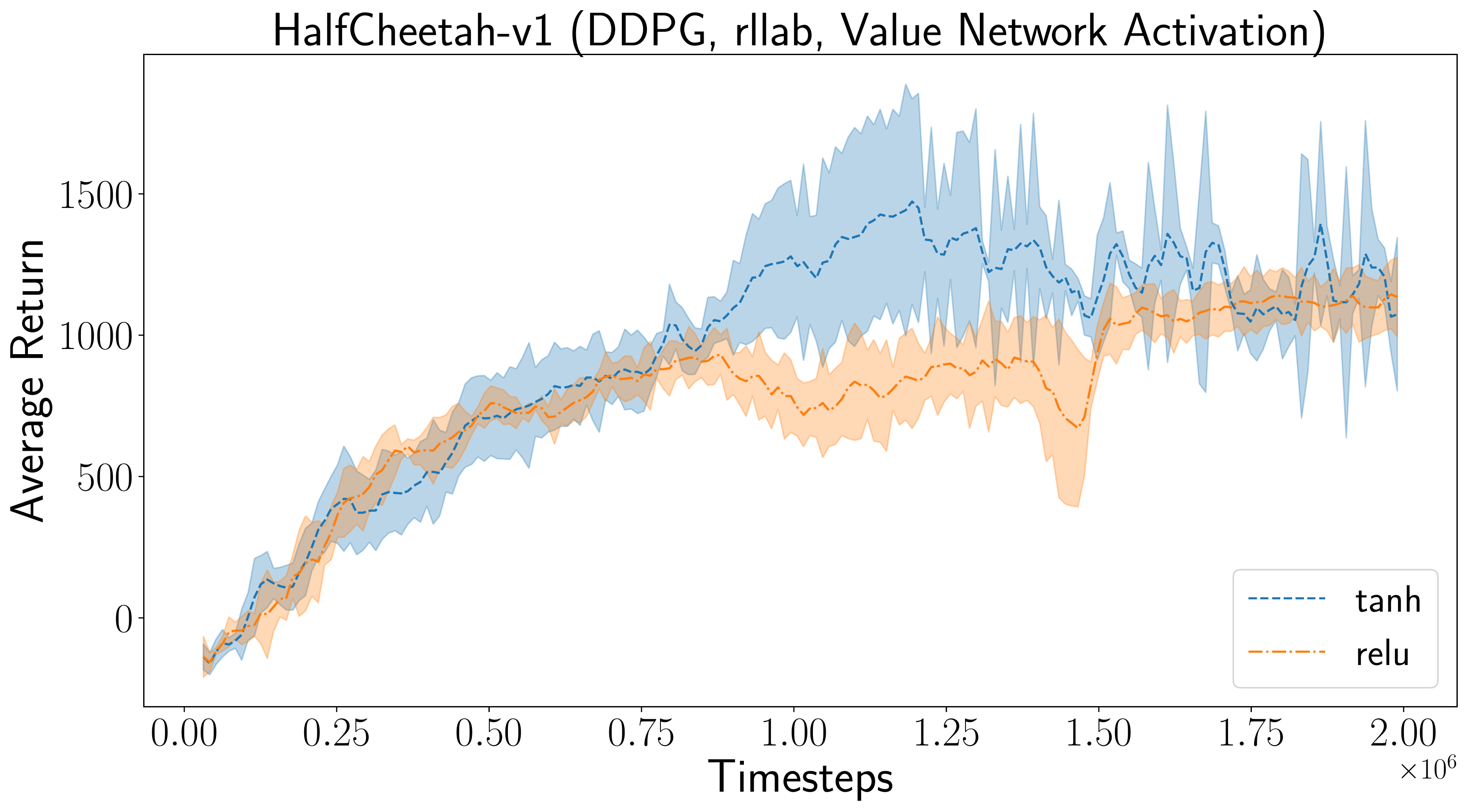}
    \includegraphics[width=.49\textwidth]{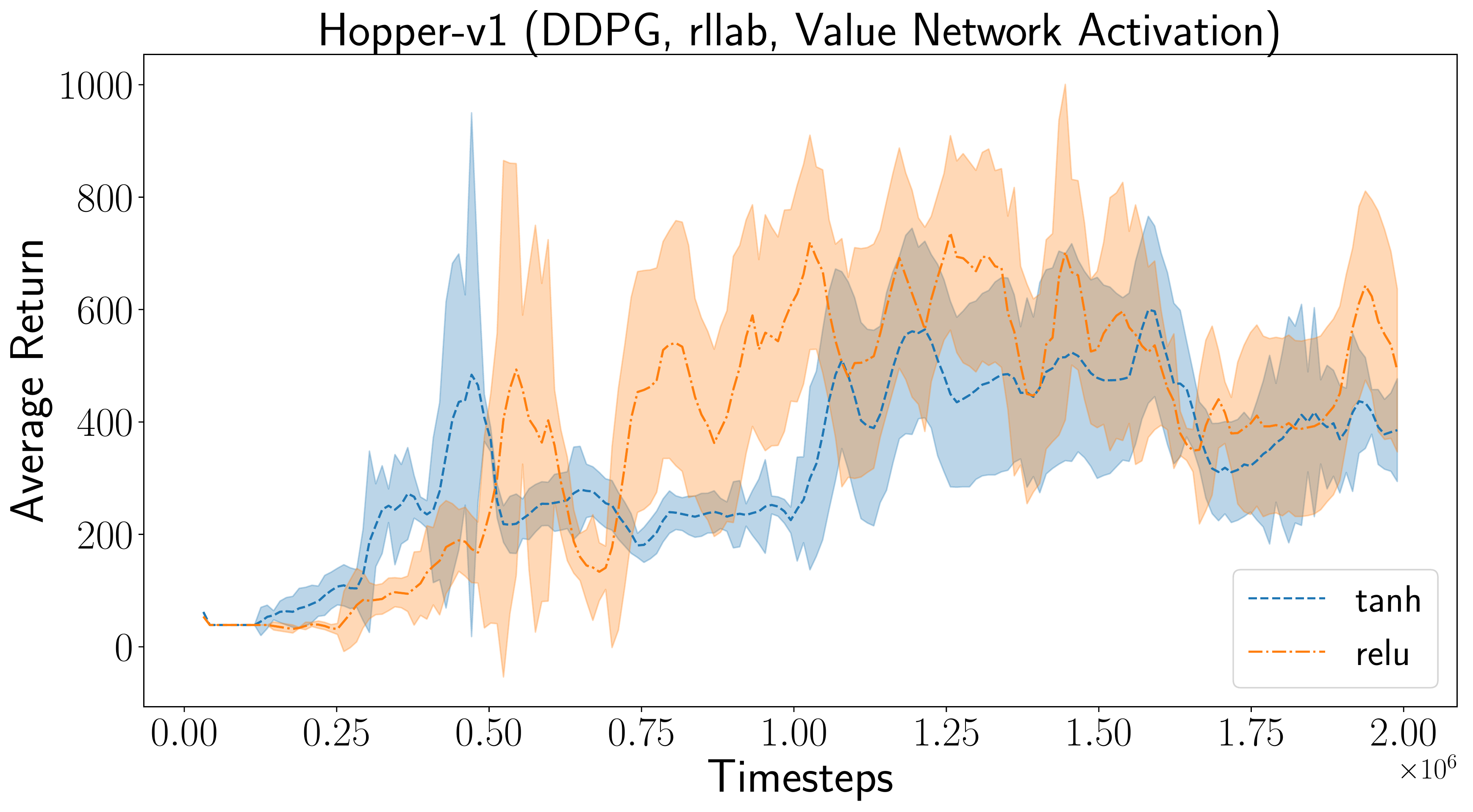}
    \caption{DDPG rllab Policy and Value Network activations.}
    \label{ddpg_rllab_2}
\end{figure}

Similarly, Figures~\ref{ddpg_rllab} and~\ref{ddpg_rllab_2} show the same network experiments for DDPG with the Theano implementation of rllab code~\cite{rllab}.

Often in related literature, there is different baseline codebase people use for implementation of algorithms. One such example is for the TRPO algorithm. It is a commonly used policy gradient method for continuous control tasks, and there exists several implementations from OpenAI Baselines \cite{plappert2017parameter}, OpenAI rllab \cite{rllab} and the original TRPO codebase \cite{TRPO}. In this section, we perform an analysis of the impact the choice of algorithm codebase can have on the performance. Figures \ref{trpo_original} and \ref{trpo_original_2} summarizes our results with TRPO policy network and value networks, using the original TRPO codebase from \cite{TRPO}. Figure~\ref{fig:trpo_rllab} shows the results using the rllab implementation of TRPO using the same hyperparameters as our default experiments aforementioned. Note, we use a linear function approximator rather than a neural network due to the fact that the Tensorflow implementation of OpenAI rllab doesn't provide anything else. We note that this is commonly used in other works~\cite{rllab,thirdperson}, but may cause differences in performance. Furthermore, we leave out our value function network experiments due to this.

\begin{figure}[H]
    \centering
    \includegraphics[width=.49\textwidth]{"images/HalfCheetah-v1__DDPG,_Codebase_Comparison_"}
    \includegraphics[width=.49\textwidth]{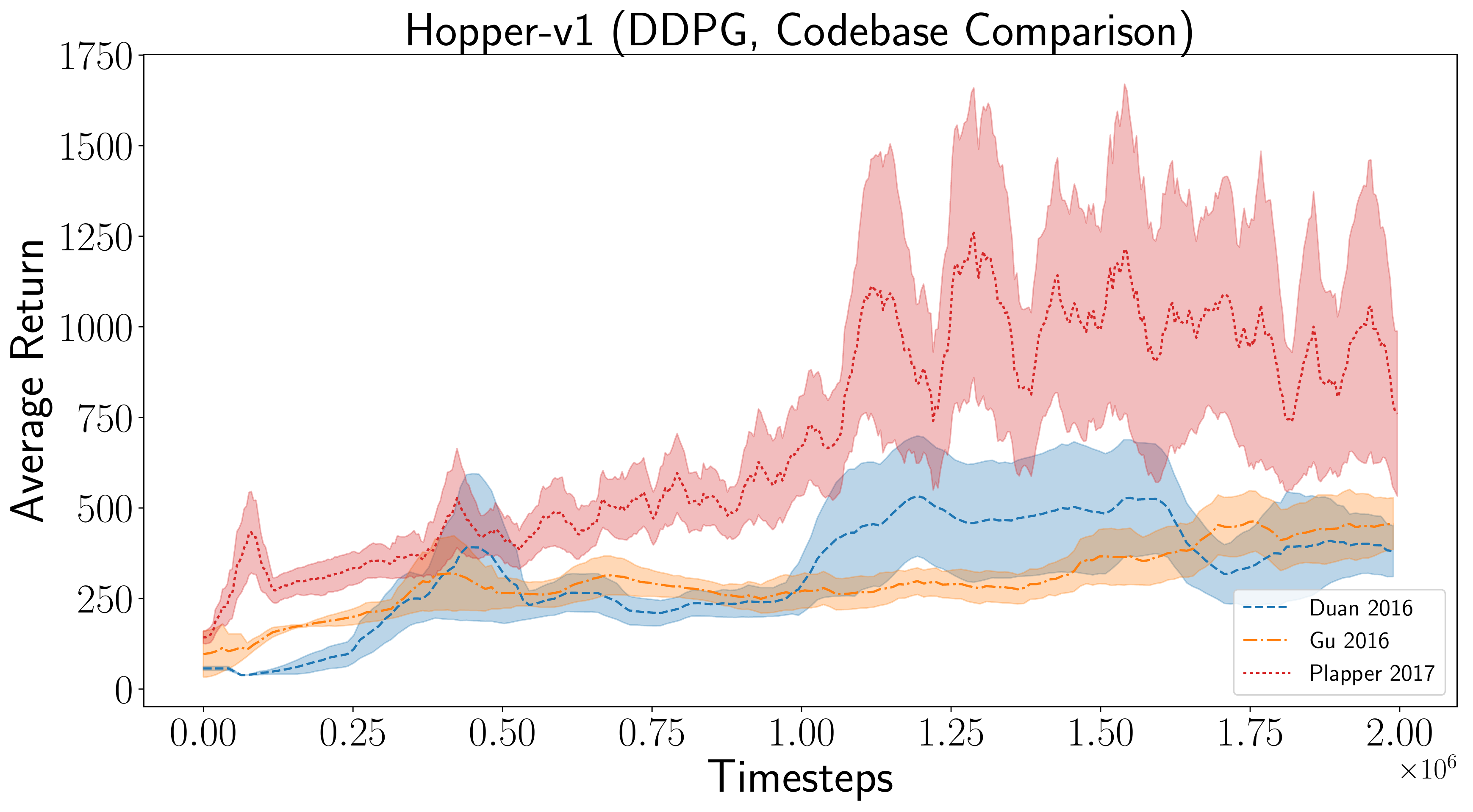}
    \caption{DDPG codebase comparison using our default set of hyperparameters (as used in other experiments).}
    \label{fig:ddpg_codebase}
\end{figure}

\begin{figure}[H]
    \centering
    \includegraphics[width=.49\textwidth]{"images/HalfCheetah-v1__TRPO,_Codebase_Comparison_"}
    \includegraphics[width=.49\textwidth]{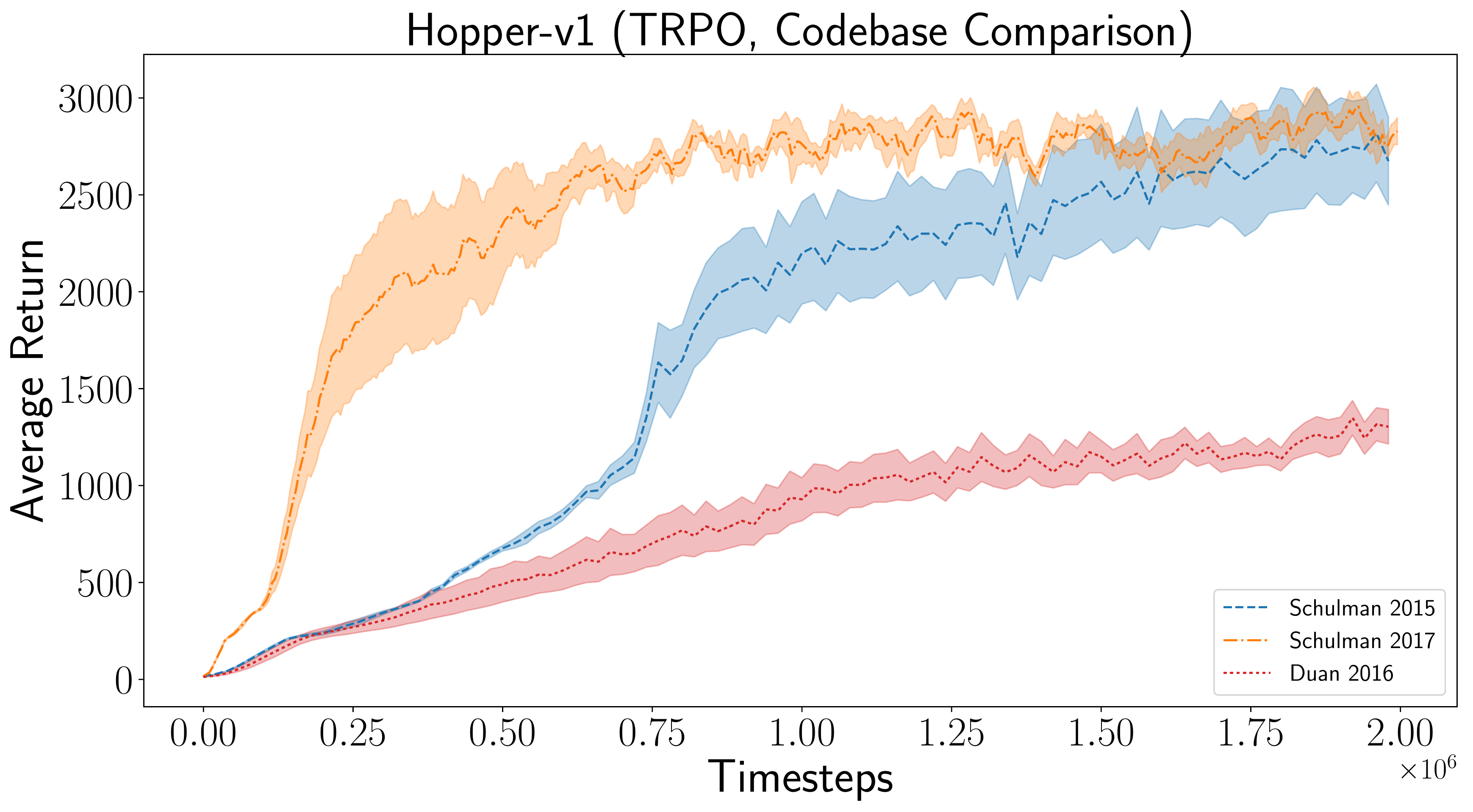}
    \caption{TRPO codebase comparison using our default set of hyperparameters (as used in other experiments).}
    \label{fig:trpo_codebase}
\end{figure}

Figure~\ref{fig:trpo_codebase} shows a comparison of the TRPO implementations using the default hyperparamters as specified earlier in the supplemental. Note, the exception is that we use a larger batch size for rllab and original TRPO code of 20k samples per batch, as optimized in a second set of experiments. Figure~\ref{ddpg_rllab_pp} and~\ref{ddpg_rllab_pp_2} show the same network experiments for DDPG with the rllab++ code~\cite{QPROP}. We can then compare the performance of the algorithm across 3 codebases (keeping all hyperparameters constant at the defaults), this can be seen in Figure~\ref{fig:ddpg_codebase}.

\section{Significance}

Our full results from significance testing with difference metrics can be found in Table~\ref{tab:significance_cheetah} and Table~\ref{tab:singificance_hopper}. Our bootstrap mean and confidence intervals can be found in Table~\ref{tab:bootstrap}. Bootstrap power analysis can be found in Table~\ref{tab:power}. To performance significance testing, we use our 5 sample trials to generate a bootstrap with 10k bootstraps. From this confidence intervals can be obtained. For the $t$-test and KS-test, the average returns from the 5 trials are sorted and compared using the normal 2-sample versions of these tests. Scipy ( \url{https://docs.scipy.org/doc/scipy-0.14.0/reference/generated/scipy.stats.ks_2samp.html}, \url{https://docs.scipy.org/doc/scipy/reference/generated/scipy.stats.ttest_ind.html}) and Facebook Boostrapped (\url{https://github.com/facebookincubator/bootstrapped}) are used for the KS test, $t$-test, and bootstrap analysis. For power analysis, we attempt to determine if a sample is enough to game the significance of a 25\% lift. This is commonly used in A/B testing~\cite{tuffery2011data}.

\begin{table}[H]
    \centering
    \scriptsize{\begin{tabular}{|c|c|c|c|c|}
    \hline
         - &DDPG&ACKTR&TRPO&PPO  \\
         \hline
         DDPG& - &\shortstack{$t=1.85,p=0.102$\\$KS=0.60,p=0.209$\\61.91 \% (-32.27 \%, 122.99 \%)}&\shortstack{$t=4.59,p=0.002$\\$KS=1.00,p=0.004$\\301.48 \% (150.50 \%, 431.67 \%)}&\shortstack{$t=2.67,p=0.029$\\$KS=0.80,p=0.036$\\106.91 \% (-37.62 \%, 185.26 \%)} \\
         \hline
         ACKTR&\shortstack{$t=-1.85,p=0.102$\\$KS=0.60,p=0.209$\\-38.24 \% (-75.42 \%, -15.19 \%)}&-&\shortstack{$t=2.78,p=0.024$\\$KS=0.80,p=0.036$\\147.96 \% (30.84 \%, 234.60 \%)}&\shortstack{$t=0.80,p=0.448$\\$KS=0.60,p=0.209$\\27.79 \% (-67.77 \%, 79.56 \%)}\\
         \hline
         TRPO&\shortstack{$t=-4.59,p=0.002$\\$KS=1.00,p=0.004$\\-75.09 \% (-86.44 \%, -68.36 \%)}&\shortstack{$t=-2.78,p=0.024$\\$KS=0.80,p=0.036$\\-59.67 \% (-81.70 \%, -46.84 \%) }&-&\shortstack{$t=-2.12,p=0.067$\\$KS=0.80,p=0.036$\\-48.46 \% (-81.23 \%, -32.05 \%) }\\
         \hline
         PPO&\shortstack{$t=-2.67,p=0.029$\\$KS=0.80,p=0.036$\\-51.67 \% (-80.69 \%, -31.94 \%) }&\shortstack{$t=-0.80,p=0.448$\\$KS=0.60,p=0.209$\\-21.75 \% (-75.99 \%, 11.68 \%) }&\shortstack{$t=2.12,p=0.067$\\$KS=0.80,p=0.036$\\94.04 \% (2.73 \%, 169.06 \%) }&-\\
         \hline
    \end{tabular}}
    \caption{HalfCheetah Significance values and metrics for different algorithms. Rows in cells are: sorted 2-sample $t$-test, Kolmogorov-Smirnov test, bootstrap A/B comparison \% difference with 95\% confidence bounds.}
    \label{tab:significance_cheetah}
\end{table}

\begin{table}[H]
    \centering
    \scriptsize{\begin{tabular}{|c|c|c|c|c|}
    \hline
         - &DDPG&ACKTR&TRPO&PPO  \\
         \hline
         DDPG&-&\shortstack{$t=-1.41,p=0.196$\\$KS=0.60,p=0.209$\\-35.92 \% (-85.62 \%, -5.38 \%) }&\shortstack{$t=-2.58,p=0.033$\\$KS=0.80,p=0.036$\\-44.96 \% (-78.82 \%, -20.29 \%) }&\shortstack{$t=-2.09,p=0.070$\\$KS=0.80,p=0.036$\\-39.90 \% (-77.12 \%, -12.95 \%) }\\
         \hline
         ACKTR&\shortstack{$t=1.41,p=0.196$\\$KS=0.60,p=0.209$\\56.05 \% (-87.98 \%, 123.15 \%) }&-&\shortstack{$t=-1.05,p=0.326$\\$KS=0.60,p=0.209$\\-14.11 \% (-37.17 \%, 9.11 \%) }&\shortstack{$t=-0.42,p=0.686$\\$KS=0.40,p=0.697$\\-6.22 \% (-31.58 \%, 18.98 \%) }\\
         \hline
         TRPO&\shortstack{$t=2.58,p=0.033$\\$KS=0.80,p=0.036$\\81.68 \% (-67.76 \%, 151.64 \%) }&\shortstack{$t=1.05,p=0.326$\\$KS=0.60,p=0.209$\\16.43 \% (-27.92 \%, 41.17 \%) }&-&\shortstack{$t=2.57,p=0.033$\\$KS=0.60,p=0.209$\\9.19 \% (2.37 \%, 15.58 \%) }\\
         \hline
         PPO&\shortstack{$t=2.09,p=0.070$\\$KS=0.80,p=0.036$\\66.39 \% (-67.80 \%, 130.16 \%) }&\shortstack{$t=0.42,p=0.686$\\$KS=0.40,p=0.697$\\6.63 \% (-33.54 \%, 29.59 \%) }&\shortstack{$t=-2.57,p=0.033$\\$KS=0.60,p=0.209$\\-8.42 \% (-14.08 \%, -2.97 \%) }&-\\
         \hline
    \end{tabular}}
    \caption{Hopper Significance values and metrics for different algorithms. Rows in cells are: sorted 2-sample $t$-test, Kolmogorov-Smirnov test, bootstrap A/B comparison \% difference with 95\% confidence bounds.}
    \label{tab:singificance_hopper}
\end{table}

\begin{table}[H]
    \centering
    \scriptsize{\begin{tabular}{|c|c|c|c|c|}
    \hline
         - &DDPG&ACKTR&TRPO&PPO  \\
         \hline
         DDPG&-&\shortstack{$t=-1.03,p=0.334$\\$KS=0.40,p=0.697$\\-30.78 \% (-91.35 \%, 1.06 \%) }&\shortstack{$t=-4.04,p=0.004$\\$KS=1.00,p=0.004$\\-48.52 \% (-70.33 \%, -28.62 \%) }&\shortstack{$t=-3.07,p=0.015$\\$KS=0.80,p=0.036$\\-45.95 \% (-70.85 \%, -24.65 \%) }\\
         \hline
         ACKTR&\shortstack{$t=1.03,p=0.334$\\$KS=0.40,p=0.697$\\44.47 \% (-80.62 \%, 111.72 \%) }&-&\shortstack{$t=-1.35,p=0.214$\\$KS=0.60,p=0.209$\\-25.63 \% (-61.28 \%, 5.54 \%) }&\shortstack{$t=-1.02,p=0.338$\\$KS=0.60,p=0.209$\\-21.91 \% (-61.53 \%, 11.02 \%) }\\
         \hline
         TRPO&\shortstack{$t=4.04,p=0.004$\\$KS=1.00,p=0.004$\\94.24 \% (-22.59 \%, 152.61 \%) }&\shortstack{$t=1.35,p=0.214$\\$KS=0.60,p=0.209$\\34.46 \% (-60.47 \%, 77.32 \%) }&-&\\
         \hline
         PPO&\shortstack{$t=3.07,p=0.015$\\$KS=0.80,p=0.036$\\85.01 \% (-31.02 \%, 144.35 \%) }&\shortstack{$t=1.02,p=0.338$\\$KS=0.60,p=0.209$\\28.07 \% (-65.67 \%, 71.71 \%) }&\shortstack{$t=-0.57,p=0.582$\\$KS=0.40,p=0.697$\\-4.75 \% (-19.06 \%, 10.02 \%) }&-\\
         \hline
    \end{tabular}}
    \caption{Walker2d Significance values and metrics for different algorithms. Rows in cells are: sorted 2-sample $t$-test, Kolmogorov-Smirnov test, bootstrap A/B comparison \% difference with 95\% confidence bounds.}
    \label{tab:singificance_walker}
\end{table}

\begin{table}[H]
    \centering
    \scriptsize{\begin{tabular}{|c|c|c|c|c|}
    \hline
         - &DDPG&ACKTR&TRPO&PPO  \\
         \hline
         DDPG&-&\shortstack{$t=-2.18,p=0.061$\\$KS=0.80,p=0.036$\\-36.44 \% (-61.04 \%, -6.94 \%) }&\shortstack{$t=-4.06,p=0.004$\\$KS=1.00,p=0.004$\\-85.13 \% (-97.17 \%, -77.95 \%) }&\shortstack{$t=-8.33,p=0.000$\\$KS=1.00,p=0.004$\\-70.41 \% (-80.86 \%, -56.52 \%) }\\
         \hline
         ACKTR&\shortstack{$t=2.18,p=0.061$\\$KS=0.80,p=0.036$\\57.34 \% (-80.96 \%, 101.11 \%) }&-&\shortstack{$t=-3.69,p=0.006$\\$KS=1.00,p=0.004$\\-76.61 \% (-90.68 \%, -70.06 \%) }&\shortstack{$t=-8.85,p=0.000$\\$KS=1.00,p=0.004$\\-53.45 \% (-62.22 \%, -47.30 \%) }\\
         \hline
         TRPO&\shortstack{$t=4.06,p=0.004$\\$KS=1.00,p=0.004$\\572.61 \% (-73.29 \%, 869.24 \%) }&\shortstack{$t=3.69,p=0.006$\\$KS=1.00,p=0.004$\\327.48 \% (165.47 \%, 488.66 \%) }&-&\shortstack{$t=2.39,p=0.044$\\$KS=0.60,p=0.209$\\99.01 \% (28.44 \%, 171.85 \%) }\\
         \hline
         PPO&\shortstack{$t=8.33,p=0.000$\\$KS=1.00,p=0.004$\\237.97 \% (-59.74 \%, 326.85 \%) }&\shortstack{$t=8.85,p=0.000$\\$KS=1.00,p=0.004$\\114.80 \% (81.85 \%, 147.33 \%) }&\shortstack{$t=-2.39,p=0.044$\\$KS=0.60,p=0.209$\\-49.75 \% (-78.58 \%, -36.43 \%) }&-\\
         \hline
    \end{tabular}}
    \caption{Swimmer Significance values and metrics for different algorithms. Rows in cells are: sorted 2-sample $t$-test, Kolmogorov-Smirnov test, bootstrap A/B comparison \% difference with 95\% confidence bounds.}
    \label{tab:singificance_swimmer}
\end{table}

\begin{table}[H]
    \centering
    \scriptsize{\begin{tabular}{|c|c|c|c|c|}
    \hline
         Environment&DDPG&ACKTR&TRPO&PPO  \\
         \hline
         HalfCheetah-v1&5037.26 (3664.11, 6574.01)&3888.85 (2288.13, 5131.96)&1254.55 (999.52, 1464.86)& 3043.1 (1920.4, 4165.86)\\
         \hline
         Hopper-v1&1632.13 (607.98, 2370.21)&2546.89 (1875.79, 3217.98)&2965.33 (2854.66, 3076.00)&2715.72 (2589.06, 2847.93)\\
         Walker2d-v1&1582.04 (901.66, 2174.66)&2285.49 (1246.00, 3235.96)&3072.97 (2957.94, 3183.10)&2926.92 (2514.83, 3361.43)\\
         \hline
         Swimmer-v1&31.92 (21.68, 46.23)&50.22 (42.47, 55.37)&214.69 (141.52, 287.92)&107.88 (101.13, 118.56)\\
         \hline
    \end{tabular}}
    \caption{Envs bootstrap mean and 95\% confidence bounds}
    \label{tab:bootstrap}
\end{table}

\begin{table}[H]
    \centering
    \begin{tabular}{|c|c|c|c|c|}
    \hline
         Environment&DDPG&ACKTR&TRPO&PPO  \\
         \hline
         HalfCheetah-v1&\shortstack{100.00 \% \\ 0.00 \% \\ 0.00 \%}&\shortstack{79.03 \%\\11.53 \% \\ 9.43 \%}&\shortstack{79.47 \% \\ 20.53 \% \\ 0.00 \%}&\shortstack{61.07 \% \\ 10.50 \% \\ 28.43 \%}\\
         \hline
         Hopper-v1&\shortstack{60.90 \%\\10.00 \% \\ 29.10 \%}&\shortstack{79.60 \% \\ 11.00 \% \\ 9.40 \%}&\shortstack{0.00 \% \\ 100.00 \% \\ 0.00 \%}&\shortstack{0.00 \% \\ 100.00 \% \\ 0.00 \%}\\
         \hline
         Walker2d-v1&\shortstack{89.50 \%\\0.00 \% \\ 10.50 \%}&\shortstack{60.33 \% \\ 9.73 \% \\ 29.93 \%}&\shortstack{0.00 \% \\ 100.00 \% \\ 0.00 \%}&\shortstack{59.80 \% \\ 31.27 \% \\ 8.93 \%}\\
         \hline
         Swimmer-v1&\shortstack{89.97 \%\\0.00 \% \\ 10.03 \%}&\shortstack{59.90 \% \\ 40.10 \% \\ 0.00 \%}&\shortstack{89.47 \% \\ 0.00 \% \\ 10.53 \%}&\shortstack{40.27 \% \\ 59.73 \% \\ 0.00 \%}\\
         \hline
    \end{tabular}
    \caption{Power Analysis for predicted significance of 25\% lift. Rows in cells are: \% insignificant simulations,\% positive significant, \% negative significant.}
    \label{tab:power}
\end{table}

% % REFERENCES
% % \section{References}
% \bibliographystyle{aaai}
% \bibliography{bibliography}